\documentclass{article}

\usepackage{amssymb}  
\usepackage{amsmath}
\usepackage{subcaption} 
\usepackage{csquotes} 

\usepackage[toc,page]{appendix}  
\usepackage{breakcites}  
\usepackage{float}
\usepackage{enumitem}
\usepackage{hyphenat}
\usepackage{svg}
\usepackage{natbib}
\usepackage{hyperref}
\usepackage{authblk}

\DeclareMathOperator*{\argmin}{arg\,min}
\DeclareMathOperator*{\argmax}{arg\,max}

\begin{document}

\title{Evaluating Feature Attribution Methods in the Image Domain}

\author[1]{Arne Gevaert\thanks{arne.gevaert@ugent.be}}
\author[2]{Axel-Jan Rousseau}
\author[2]{Thijs Becker}
\author[2]{Dirk Valkenborg}
\author[3]{Tijl De Bie}
\author[1]{Yvan Saeys}

\affil[1]{Department of Applied Mathematics, Computer Science and Statistics: Data mining and Modeling for Biomedicine (DaMBi)\\Ghent University}
\affil[2]{Center for Statistics (CENSTAT)\\Hasselt University}
\affil[3]{IDLab, Department of Electronics and Information Systems\\Ghent University}

\maketitle

\begin{abstract}%
Feature attribution maps are a popular approach to highlight the most important pixels in an image for a given prediction of a model. Despite a recent growth in popularity and available methods, the objective evaluation of such attribution maps remains an open problem. Building on previous work in this domain, we investigate existing quality metrics and propose new variants of metrics for the evaluation of attribution maps. We confirm a recent finding that different quality metrics seem to measure different underlying properties of attribution maps, and extend this finding to a larger selection of attribution methods, quality metrics, and datasets. We also find that metric results on one dataset do not necessarily generalize to other datasets, and methods with desirable theoretical properties do not necessarily outperform computationally cheaper alternatives in practice. Based on these findings, we propose a general benchmarking approach to help guide the selection of attribution methods for a given use case. Implementations of attribution metrics and our experiments are available online\footnote{\url{https://github.com/arnegevaert/benchmark-general-imaging}}.
\end{abstract}

\section{Introduction}
Deep neural networks have for some years been the state of the art for a number of predictive tasks, such as image classification \citep{krizhevsky2012, simonyan2015, he2015}, language modeling \citep{vaswani2017, brown2020} and reinforcement learning \citep{mnih2015, silver2017, lillicrap2019}. This has led to their widespread adoption in many areas of machine learning. Such models are however notorious for their black box nature: due to the large numbers of parameters and complex neural architectures, their predictions become very difficult or even impossible to understand. Interpretability of predictive models is a very useful property for many different reasons: it allows us to extract understandable knowledge from large datasets, potentially leading to new knowledge about the data itself, debug models when they fail, and explain predictions to end users to build trust in the system \citep{Doshi-Velez2017}. In some cases, the ability to explain predictions is crucial for model deployment.

For this reason, a number of different techniques have been proposed to try to make neural networks more explainable. Proposed approaches include extracting interpretable rules \citep{ribeiro2018anchors}, counterfactual explanations \citep{wachter2017counterfactual, dandl2020multi}, model distillation \citep{liu2018}, and feature attribution \citep{Ribeiro2016, Selvaraju2017, Sundararajan2017,Springenberg2014,Simonyan2014,erion2020,Smilkov2017}. In this work, we focus on the latter type of explanation. Feature attribution explanations are among the most popular techniques for explaining image classification models, because they can easily be visualized as a heatmap showing which pixels in an image are important (in the case of color images, the attribution value of a pixel can be defined to be the average or maximum absolute value of its three color components). Feature attribution techniques can also be used to measure the importance of hidden neurons or layers \citep{Shrikumar2017, Selvaraju2017}, although the focus in this work is on pixel attribution in the image domain.

The exact task of feature attribution can be interpreted in different ways, leading to some discussion about which properties a feature attribution method should satisfy \citep{chen2020}. Feature attribution methods can roughly be divided in four categories using two properties: local vs. global, and model- vs. data-centric. The first property concerns the scope of the explanation: local feature attribution maps the importance of features \textit{in a given sample}, whereas global feature attribution maps the importance of features for \textit{all samples} in a dataset (also called feature \textit{importance}). The second property concerns the target of the explanation: model-centric feature attribution concerns the importance of features \textit{for a specific model}, whereas data-centric feature attribution measures the \textit{informativeness} of features in the data, independently of any specific model. This can be estimated using classical statistical or information-theoretical techniques \citep{chandrashekar2014survey}. Model- and data-centric feature attributions are not necessarily the same, as a model can often make predictions using only a subset of the informative features, or even using features that are generally non-informative (in which case the model is overfitting). In this work, we specifically evaluate \textbf{local, model-centric} feature attributions.

Because of the desire for model explanations and the widespread popularity of deep neural networks in the domain of image classification, many feature attribution methods have been proposed in recent years. These methods can roughly be divided into gradient-based, perturbation-based and CAM-based methods \citep{Ancona2017,Selvaraju2017}. Each attribution method creates a different explanation for the same prediction. This has naturally led to the question of explanation quality: which methods work best? This turns out to be a very difficult question, since there is no ground truth available in the form of ``perfect'' feature attribution scores.

Attempts at evaluating feature attribution explanations can roughly be categorized in three types of approaches. The first is human evaluation. This includes simply looking at an explanation and seeing if it ``makes sense'', or performing a user study to see how helpful explanations are for predicting model behaviour \citep{schmidt2019quantifying}. The disadvantage of these approaches is that such user studies are difficult to set up, and their results are inherently subjective. It has been shown that, just because an explanation makes sense to humans, does not mean that it is true to the underlying workings of the model \citep{Adebayo2018a}. Also, a user study is generally infeasible to perform for each use case, and it is unclear whether results from user studies can be generalized to different datasets.

A second approach is to define a set of desirable properties, or \textit{axioms}, that a method should have \citep{Lundberg2017, Sundararajan2017}. Examples of such axioms include local accuracy, missingness and consistency \citep{Lundberg2017}. Such approaches are more objective in nature, but recent work has shown that methods that conform to these axioms are still not necessarily accurate \citep{Adebayo2018a}. Some of these axioms can also be implemented in different ways, leading to a number of methods that all conform to certain axioms, but still provide different explanations for the same prediction \citep{Sundararajan2019}.

Finally, we can define quantitative metrics that try to indicate the quality of an explanation by measuring the behaviour of the model or explanation after applying specific perturbations \citep{Ancona2017, Yeh2019}. A simple example of this kind of measure is Deletion \citep{Samek2015}. Here, we iteratively mask the top $n$ most important features, as indicated by the explanation. If the features that were marked as important are truly important, we would expect the output of the model to drop rapidly with increasing $n$. Another example is the so-called \textit{sanity check} proposed by \citet{Adebayo2018a}. This sanity check works by randomizing the parameters of the model, and comparing the original attribution map to the attribution map computed for the randomized model. If these two maps are similar, then the attribution method is independent of the model parameters, and is therefore viewed as \textit{failing} the sanity check.

In this work, we implement several existing and newly proposed quality metrics for evaluating feature attribution methods. These metrics are evaluated on a large number of attribution methods, and we investigate the results on 8 different datasets of varying dimensionality. Our contributions are as follows:
\begin{itemize}
	\item We expand on the work done in \citet{Tomsett}, showing that different quality metrics measure different underlying properties of attribution maps. We extend this finding to a significantly larger set of quality metrics, attribution methods, and datasets.
	\item We demonstrate that the results of quality metrics for attribution maps, including the sanity check from \citet{Adebayo2018a}, vary significantly across different datasets. From this observation, we conclude that quality metrics should be computed separately for each given use case, rather than assuming that the results for one dataset or model will generalize to another setting.
	\item We propose three new metrics: Minimal Subset Deletion, Minimal Subset Insertion and Seg-Sensitivity-n (based on Sensitivity-n \citep{Ancona2017}). We empirically show that Seg-Sensitivity-n provides results with a higher signal-to-noise ratio on high-dimensional datasets.
	\item We find that the performance of some methods is complementary to that of other methods, suggesting that a combination of these attribution methods may be valuable.
	\item We find that, depending on the dataset, methods with strong theoretical foundations such as DeepSHAP \citep{Lundberg2017} do not necessarily outperform their computationally cheaper counterparts such as DeepLIFT \citep{Shrikumar2017}. This suggests that a benchmarking approach can be useful to check if a computationally intensive method is truly more valuable than a simpler one for a given use case.
	\item Finally, we provide general benchmarking guidelines to help guide the search for an appropriate attribution method or set of attribution methods for a given use case.
\end{itemize}

\section{Related Work}
Although the systematic evaluation of feature attribution methods is a fairly recent topic, a number of attempts have already been made to systematically and objectively evaluate the quality of explanations. An early, intuitive way of evaluating feature attributions was proposed by \citet{Samek2015}. In this approach, the top $k$ most important features are removed by replacing them with random noise. Consequently, the difference in output of the model is measured. If the most important features are truly important to the model, we expect a sharp drop in confidence for the predicted class.

A more general approach was proposed by \citet{Ancona2017}, called Sensitivity-$n$. This metric is computed by removing a number of random subsets of $n$ pixels from the image, and measuring the correlation of the difference in output with the sum of attribution values of those removed pixels. This allows one to assess the accuracy of the attribution values \textit{in general}, rather than just the top most important features.

A possible problem with the metrics mentioned above is the fact that masking inputs in images can introduce high-frequency artifacts, which can push the images outside of their normal data distribution. This can cause the model to produce arbitrary outputs. Although the exact impact of this problem on the scores produced by metrics is unclear, some efforts to resolving it have already been made, including the Remove And Retrain (ROAR) procedure \citep{Hooker2018}. Here, the authors attempt to resolve the OOD problem by modifying the Deletion metric by \citet{Samek2015} such that after every iteration, the model is retrained on the data where the top $k$ pixels are removed. The reasoning is that in this way, the model learns to regard the mean-valued pixels as uninformative.

However, we argue that this metric is not measuring the same kind of feature attribution as the original Deletion metric. Because the model is retrained after each iteration, it is able to detect and use different parts of the input to make a prediction. Also, there is no guarantee that the model, after retraining, will consider the masked pixels as uninformative: the shape or location of regions with that specific color (the dataset mean) can still be very informative. In other words, ROAR can only assess the ability of methods to map local, \textit{data-centric} feature attributions. Since the methods we are evaluating are designed to map local, model-centric attributions, we do not incorporate this metric in our analysis.

More recently, \citet{Yeh2019} proposed Infidelity and Max-Sensitivity, two complementary metrics that measure the accuracy of a method and its robustness against small, insignificant perturbations, respectively. Recent work has shown that some feature attribution methods, much like neural networks themselves, are vulnerable to adversarial attacks \citep{Ghorbani2019}. This makes the robustness of explanations an interesting property to measure in addition to the accuracy.

\citet{Yang2019} proposed a synthetic data approach, where objects from MSCOCO \citep{lin2015} were pasted into background images from MiniPlaces \citep{zhou2017places}. A model is then trained to classify either the background or the object in the image. Because it is known where in the image the object was pasted, a relative form of ground truth is available. From this, a number of metrics are derived. However, as opposed to the previously proposed metrics, these metrics are tied to a specific dataset, and cannot be calculated for any given dataset and model. For this reason, we do not consider this approach in our work.

Another related approach was proposed by \citet{Adebayo2018a}. In this work, a relative form of ground truth is created by randomizing the parameter values of the network, layer by layer. The assumption is that the feature attribution map should be significantly different for a trained model vs. a randomized model. Methods that return the same attribution map for both models, appear to be independent of the model parameters. This approach does not provide a numerical value that captures the quality of the explanation, but rather acts as a pass/fail ``sanity check''.

Recently, \citet{Tomsett} have shown that some of these metrics are very dependent on implementation details, and do not appear to be measuring the same underlying properties of explanations. This is shown by measuring the correlation between different metric scores. The authors find that details such as how pixels are masked (by setting them to 0 vs. replacing them with random noise), or in what order they are masked (by decreasing or increasing importance), have a great influence on the quality scores given by the metric. This suggests that these metrics, although they are all designed to measure the ``accuracy'' of explanations, appear to be measuring different underlying properties. We build upon this work by applying a similar but more extensive analysis on a larger number of metrics and methods. In doing so, we can draw more global conclusions about how different methods and metrics relate to each other, and which methods and metrics may be most desirable for specific use cases.

\section{Definitions and Notation}
\label{sec:notation}
We define an \textit{instance} as a vector $\mathbf{x} \in \mathcal{X} \subseteq \mathbb{R}^d$, where $d$ is the number of inputs (pixels, or color values of pixels in the case of RGB images). A \textit{model} is defined as a function $m: \mathcal{X} \to \mathbb{R}^o$, where $o$ is the number of output classes. Note that the output of the model for a given class $c$ can be any real number. In many cases, the output of the model is followed by a softmax function $\sigma$, mapping the outputs from $\mathbb{R}^o$ to $[0,1]^o$. In this case, the original outputs are called \textit{logits}. In this work, we consider the logit values as the actual output of the model. We write $m(\mathbf{x})_c$ as the $c$-th component of the output of model $m$ on instance $\mathbf{x}$.

We denote the \textit{model class} as $\mathcal{M} = \{m\}$, this represents the set of possible model instantiations (for example, the set of all possible neural networks, or all possible neural networks of a given architecture). An \textit{attribution method} is a function $E \in \mathcal{E}: \mathcal{M} \times \mathcal{X} \times \{1, \dots, o\} \to \mathbb{R}^d$. The result of this function is called an \textit{attribution map}. We explicitly mention the model $m \in \mathcal{M}$ as an argument of this function to indicate that we consider local, model-centric attributions, which are dependent on a specific combination of instance and model. The output class is also an argument of the attribution method, as attributions can be calculated for each output of the model. Finally, we define an \textit{attribution metric} as a function $M: \mathcal{M} \times \mathcal{X} \times \mathbb{R}^d \times \{1, \dots, o\} \rightarrow \mathbb{R}$ (type I) or $M: \mathcal{M} \times \mathcal{X} \times \mathcal{E} \times \{1, \dots, o\} \rightarrow \mathbb{R}$ (type II), mapping a model $m$, an instance $\mathbf{x}$, an attribution map $\mathbf{e}$ (type I) or attribution method $E$ (type II), and an output class $c$ to a single real number which represents the quality of the attributions given by $\mathbf{e}$ or $E$ for output $c$ of model $m$ on instance $\mathbf{x}$.

For an instance $\mathbf{x}$ and attribution map $\mathbf{e}$, we will denote $\mathbf{x}_k^\mathbf{e}$ and $\mathbf{x}_{-k}^\mathbf{e}$ as the instance $\mathbf{x}$ where respectively the $k$ most or least important inputs are removed according to $\mathbf{e}$. This removal, or ``masking out'' of features can be implemented in a number of different ways, which will be discussed in detail in Section \ref{sec:Masking}. In the case of color images, we define the attribution value of a pixel as the average value of its color components, and proceed analogously.

Finally, we will denote $S = \{S^l\}_{l=1}^L, S^l \in \{0,1\}^d$ as the set of segments of an input sample $\mathbf{x}$ as produced by a given segmentation algorithm, where $S^l_i = 1$ if the $i$th input feature is part of segment $l$, and $S^l_i = 0$ otherwise.
The attribution value of a segment $S^l$ can then simply be computed as the average attribution value of its input features: $\mathbf{e}_{S^l} := \frac{\| \mathbf{e} \odot S^l \|_1}{\| S^l \|_1}$, where $\odot$ indicates element-wise multiplication.
For an instance $\mathbf{x}$, a corresponding segmentation $S$, and an attribution map $\mathbf{e}$, we will denote $\mathbf{x}_{k_S}^\mathbf{e}$ (resp. $\mathbf{x}_{-k_S}^\mathbf{e}$) as the same sample $\mathbf{x}$ where the $k$ most (resp. least) important segments are masked out.

\section{Attribution Metrics}
We now describe the different quality metrics that were used to evaluate the attribution methods described in Section \ref{sec:methods}.
A summary of general properties can be seen in Table \ref{tbl:metrics}:

\begin{itemize}
	\item \textbf{Attribution range:} Indicates which parts of the attribution map the metric actually evaluates: we denote metrics that evaluate the most important, least important, or all inputs as \textit{high-end}, \textit{low-end} or \textit{overall} metrics, respectively. For example: $Del_{MoRF}$ and $Del_{LeRF}$ evaluate the high- and low-end, respectively, because they measure the influence of removing the top and bottom $k$ features, respectively (see further).
	\item \textbf{Masking:} Indicates whether the metric relies on masking inputs in its implementation. Metrics that do, can be implemented in different ways, as the choice of a neutral value to replace features with is not obvious (see Section \ref{sec:Masking}).
	\item \textbf{Data type:} Indicates which types of data the metric can be applied to. In our case, a metric can either be applied to any kind of data, or only to image data (for example, because it relies on an image segmentation algorithm \citep{Rieger}, or an adversarial patch \citep{QiuLin}).
	\item \textbf{Complexity:} Indicates the computational complexity of the metric expressed as a number of forward passes through the model. $C_{mth}$ is the complexity of the attribution method being evaluated, also expressed as a number of forward/backward passes.
	\item \textbf{Interface:} We define two interfaces for attribution metrics:
	      \begin{itemize}
		      \item \textbf{Type I:} $M: \mathcal{M} \times \mathcal{X} \times \mathbb{R}^d \times \{1, \dots, o\} \rightarrow \mathbb{R}$. A type I metric accepts an \textit{attribution map} $\mathbf{e} \in \mathbb{R}^d$ to evaluate. This allows one to compute the metric result for any attribution map, regardless of whether the implementation of the attribution method that generated it is available or not.
		      \item \textbf{Type II:} $M: \mathcal{M} \times \mathcal{X} \times \mathcal{E} \times \{1, \dots, o\} \rightarrow \mathbb{R}$. A type II metric needs access to the \textit{attribution method} under evaluation $E \in \mathcal{E}$, because the method needs to be re-executed at some point in the computation of the metric. If the implementation of the attribution method is not available, this type of metric cannot be computed.
	      \end{itemize}
\end{itemize}

\begin{table}
	\centering
	\begin{center}
		\begin{tabular}{||l c c c c c ||}
			\hline
			Metric        & Range    & Masking      & Data type & Complexity                 & Interface \\ [0.5ex] 
			\hline\hline
			$Del_{MoRF}$  & high-end & $\checkmark$ & any       & $\mathcal{O}(L)$           & Type I    \\
			\hline
			$Del_{LeRF}$  & low-end  & $\checkmark$ & any       & $\mathcal{O}(L)$           & Type I    \\
			\hline
			$Ins_{MoRF}$  & high-end & $\checkmark$ & any       & $\mathcal{O}(L)$           & Type I    \\
			\hline
			$Ins_{LeRF}$  & low-end  & $\checkmark$ & any       & $\mathcal{O}(L)$           & Type I    \\
			\hline
			$MS_{Del}$    & high-end & $\checkmark$ & any       & $\mathcal{O}(d)$           & Type I    \\
			\hline
			$MS_{Ins}$    & high-end & $\checkmark$ & any       & $\mathcal{O}(d)$           & Type I    \\
			\hline
			$IROF_{MoRF}$ & high-end & $\checkmark$ & image     & $\mathcal{O}(|S|)$         & Type I    \\
			\hline
			$IROF_{LeRF}$ & low-end  & $\checkmark$ & image     & $\mathcal{O}(|S|)$         & Type I    \\
			\hline
			$Sens_n$      & overall  & $\checkmark$ & any       & $\mathcal{O}(k)$           & Type I    \\
			\hline
			$SegSens_n$   & overall  & $\checkmark$ & image     & $\mathcal{O}(k)$           & Type I    \\
			\hline
			$INFD$        & overall  & $\checkmark$ & any       & $\mathcal{O}(k)$           & Type I    \\
			\hline
			$SENS_{MAX}$  & overall  &              & any       & $\mathcal{O}(k * C_{mth})$ & Type II   \\
			\hline
			$COV$         & high-end &              & image     & $C_{mth}$ $^1$             & Type II   \\
			\hline
			$PR$          & overall  &              & any       & $C_{mth}$                  & Type II   \\
			\hline
		\end{tabular}
		\\
		\footnotesize{$^1$: provided that an adversarial patch is already available.}
	\end{center}
	\caption{Summary of metrics.}
	\label{tbl:metrics}
\end{table}

\subsection{Deletion}
The first and most widely known metric is Deletion \citep{Samek2015}. This metric works by iteratively removing the top $k$ most important features from an input sample. This is done by masking the feature with some value (see further: \ref{sec:Masking}).

An ordering of features where the most important features are ranked highest will cause a steep decrease in the output confidence of the model. This can be summarized by computing the area under the perturbation curve, where a low AUC corresponds to a good explanation.
\citet{Samek2015} also introduces an alternative variant of the Deletion metric, where the features are masked in reversed order of importance. In that case, a high AUC value indicates a good attribution map. We call this variant Deletion$_{LeRF}$ (Least Relevant First), and the original Deletion$_{MoRF}$ (Most Relevant First).

\begin{align*}
	Del_{MoRF}(\mathbf{x}, m, \mathbf{e}, c) & = \frac{1}{L} \sum_{k=1}^L m(\mathbf{x}^\mathbf{e}_{k})_c  \\
	Del_{LeRF}(\mathbf{x}, m, \mathbf{e}, c) & = \frac{1}{L} \sum_{k=1}^L m(\mathbf{x}^\mathbf{e}_{-k})_c
\end{align*}

Where $L$ is the maximum number of inputs masked, and $c$ is the output that the attribution is intended to explain (usually this is the highest output of the model, which corresponds to the class that the model assigned to $\mathbf{x}$). For large images, we can approximate this value by taking
a fixed number of steps with a constant step size. The MoRF-variant evaluates the high end of the attribution map, whereas the LeRF-variant evaluates the low end. We choose $L$ such that at most 15\% of pixels are masked, which corresponds to the original approach in \citet{Samek2015}. This limits the influence of out-of-distribution effects: as more pixels are removed, the image gets further removed from the original data manifold, making the result less representative. This metric scales linearly in the number of steps $L$ taken to compute the AUC.

\subsection{Insertion}

A simple variant of the Deletion metric is Insertion \citep{Petsiuk}. This metric works entirely analogously to Deletion, but instead of iteratively removing features from the original input sample, we now iteratively insert pixels of the original image onto a blank background (which is again defined by the masking procedure).

Analogously to the Deletion metric, we can again define two variants of Insertion: Insertion$_{LeRF}$ and Insertion$_{MoRF}$, where resp. the least and most relevant features are inserted first. Since inserting the $k$ most important features is the same as removing the $d-k$ least important ones, we can define Insertion as follows:

\begin{align*}
	Ins_{MoRF}(\mathbf{x}, m, \mathbf{e}, c) & = \frac{1}{L} \sum_{k=1}^L m(\mathbf{x}^\mathbf{e}_{-(d-k)})_c \\
	Ins_{LeRF}(\mathbf{x}, m, \mathbf{e}, c) & = \frac{1}{L} \sum_{k=1}^L m(\mathbf{x}^\mathbf{e}_{d-k})_c
\end{align*}

Note that if $L = d$, $Ins_{MoRF} = Del_{LeRF}$ and $Ins_{LeRF} = Del_{MoRF}$.
Again, the MoRF- and LeRF-variants measure the high and low end respectively. This metric also scales linearly with the number of steps hyperparameter $L$.

\subsection{Minimal Subset}

The previously mentioned Deletion and Insertion metrics only take into account the model's confidence in the originally predicted class $c$. However, the actual prediction of the model is also dependent on the confidence of the other classes.
The removal of certain features could, for example, hardly influence the output confidence in $c$, but drastically change the confidence of another class $c'$, causing the model to change its overall prediction.
To mitigate this problem, we introduce Minimal Subset Deletion and Minimal Subset Insertion.

These metrics work by iteratively removing (resp. inserting) the top $k$ most important features from the input sample, and recording the smallest value for $k$ that causes the prediction of the model to change. For Minimal Subset Insertion specifically, the prediction must change into the originally predicted class $c$.

\begin{align*}
	MS_{Del}(\mathbf{x}, m, \mathbf{e}, c) & = \argmin_{k \in \{1, \dots, d\}} (\argmax(m(\mathbf{x}^\mathbf{e}_{k})) \neq \argmax(m(\mathbf{x})))   \\
	MS_{Ins}(\mathbf{x}, m, \mathbf{e}, c) & = \argmin_{k \in \{1, \dots, d\}} (\argmax(m(\mathbf{x}^\mathbf{e}_{-(d-k)})) = \argmax(m(\mathbf{x})))
\end{align*}

For analogous reasons as with Deletion/Insertion, this metric evaluates the high end of the attribution map. Both variants scale linearly with the amount of dimensions $d$.

\subsection{IROF}
Iterative Removal Of Features (IROF) \citep{Rieger} is an extension of Deletion, where a segmentation $S$ of the input sample $\mathbf{x}$ is used.
Instead of iteratively masking the $k$ most important inputs, we now mask the $k$ most important \textit{segments}.
This can reduce the number of forward passes needed, and can provide insight into the quality of an attribution at a larger scale: an algorithm that is able to find the top few pixels that maximally perturb the network when removed, might score
very well on Deletion, but not so much on IROF. If another algorithm correctly identifies the most important ``regions'', it might score better on IROF and worse on Deletion. In some cases, the latter might be more interesting, as this would likely correspond to explanations that are less noisy and more easily readable.

We can define IROF$_{MoRF/LeRF}$ analogously to Deletion$_{MoRF/LeRF}$, leading to the following definitions:

\begin{align*}
	IROF_{MoRF}(\mathbf{x}, m, \mathbf{e} ,c) & = \frac{1}{|S|} \sum_{k=1}^{|S|} m(\mathbf{x}^\mathbf{e}_{k_S})_c  \\
	IROF_{LeRF}(\mathbf{x}, m, \mathbf{e}, c) & = \frac{1}{|S|} \sum_{k=1}^{|S|} m(\mathbf{x}^\mathbf{e}_{-k_S})_c
\end{align*}

Note that, even though all segments are removed in IROF, we classify this metric as high-end. This is because the metric score still depends most on the top most important image segments: if those are identified correctly, the model output will decrease quickly, and the other segments will have little influence on the metric score. IROF scales linearly with the amount of segments $|S|$, and is only applicable to image data because of the dependence on an image segmentation algorithm.
We implement IROF using the SLIC algorithm \citep{achanta2010slic}, with an approximate number of segments of 100.

\subsection{Sensitivity-n}
Previous metrics have only considered the most or least important features. This can be a problem if the inputs contain a large number of features, in which case a large proportion of the features is hardly evaluated, or has a small influence on the evaluation.
To get a more global assessment of the quality of feature attributions, Sensitivity-n was introduced \citep{Ancona2017}. Formally, Sensitivity-n is defined as follows \citep[quoted from][where mathematical notation was adjusted to conform to ours]{Ancona2017}:

\begin{displayquote}
	An attribution method satisfies Sensitivity-$n$ when the sum of the attributions for any subset of features of cardinality $n$ is equal to the variation of the output $m(\mathbf{x})_c$ caused by removing the features in the subset. 
\end{displayquote}

Since no attribution method exactly satisfies Sensitivity-$n$ for all values of $n$, the metric instead measures how well the sum of attributions $\sum_{s \in S} \mathbf{e}_{s}$ correlates with the difference in output $m(\mathbf{x})_c - m(\mathbf{x}_{S})_c$, using the Pearson correlation coefficient (where $\mathbf{x}_S$ denotes the instance $\mathbf{x}$ with all features in $S$ removed, and $\mathbf{e}_s$ denotes the attribution of feature $s$ according to attribution map $\mathbf{e}$). We can compute Sensitivity-$n$ as:

$$
	Sens_n(\mathbf{x}, m, \mathbf{e}, c) = r\left(\sum_{s \in S_i} \mathbf{e}_{s}, m(\mathbf{x})_c - m(\mathbf{x}_{S_i})_c\right)
$$

Where $S_i$ is a random subset of inputs of size $n$, and $r(X,Y)$ is the Pearson correlation coefficient between variables $X$ and $Y$. The correlation is computed using $k$ randomly selected subsets $S_i$. We choose $k=100$, which corresponds to the configuration in \citet{Ancona2017}.

The number of possible subsets of features grows exponentially with $d$. Because of this, the approximation made by this metric will get exponentially worse for increasing image size.
To mitigate this problem, we introduce a segmented variant of Sensitivity-n, called \textbf{Seg-Sensitivity-n}. This metric works by first segmenting the input image $\mathbf{x}$ into segments $S$, and then removing random subsets of segments instead of features. Since the amount of segments is drastically lower than the number of features, selecting 100 random subsets gives a more representative sample, which we expect will increase the signal-to-noise ratio of this metric.

$$
	SegSens_n(\mathbf{x}, m, \mathbf{e}, c, S) = r\left(\sum_{l \in L_i} \mathbf{e}_{S^l}, m(\mathbf{x})_c - m(\mathbf{x}_{S^{L_i}})_c\right)
$$

Where $S$ is the segmentation of instance $\mathbf{x}$ (represented as a set of segments $\{S^l\}$), and $\mathbf{x}_{S^{L_i}}$ denotes the instance $\mathbf{x}$ with all segments in $L_i$ removed. The correlation is now computed using $k=100$ randomly selected subsets of segments $L_i$. Since the subsets of features/segments are chosen randomly, Sensitivity-n and Seg-Sensitivity-n evaluate the overall attribution map. Both metrics scale linearly in the number of subsets $k$.

\subsection{Infidelity}

Infidelity \citep{Yeh2019} generalizes the previous metrics from perturbation by masking to general perturbations. This is done by comparing the difference in output after an arbitrary perturbation with the dot product of the perturbation vector $\mathbf{I}$ and the attribution map $\mathbf{e}$.
The perturbation vector is a random variable $\mathbf{I} \in \mathbb{R}^d$ with probability measure $\mu_\mathbf{I}$. The infidelity of an attribution map $\mathbf{e}$ for an input sample $\mathbf{x}$ and class $c$ is then defined as follows:

\begin{align*}
	INFD(\mathbf{x}, m, \mathbf{e}, c) & = \mathbb{E}_{\mathbf{I} \sim \mu_\mathbf{I}} \left[ (\beta\mathbf{I}^T \mathbf{e} - (m(\mathbf{x})_c - m(\mathbf{x} - \mathbf{I})_c))^2 \right]         \\
	\beta                              & = \frac{\mathbb{E}_{I \sim \mu_I} [ I^T \mathbf{e} (m(\mathbf{x})_c - m(\mathbf{x} - \mathbf{I})_c) ]}{\mathbb{E}_{I \sim \mu_I} [ (I^T \mathbf{e})^2 ]}
\end{align*}

Here, $\beta$ acts as a normalizing term (called \textit{optimal scaling} in the original paper) to make the values for different explanation methods comparable.
We use two variants of Infidelity proposed in \citet{Yeh2019}, defined by their perturbation vectors:
\begin{itemize}
	\item Difference to noisy baseline ($INFD_{NB}$): $\mathbf{I} = \mathbf{x} - \epsilon$, where $\epsilon \sim \mathcal{N}(0, \sigma^2)$. This corresponds to a robust variant of the completeness axiom \citep{Lundberg2017}, where we take a Gaussian random vector centered around a zero baseline, instead of a constant zero baseline.
	\item Square removal ($INFD_{SQ}$): in this case, $\mathbf{I}$ has a uniform distribution over square patches of the image $\mathbf{x}$ of some predefined size. This can better capture spatial relationships in the images, as the removal of single pixels actually removes very little information if the surrounding pixels are still intact.
\end{itemize}
Since the perturbations happen on the entire image or on randomly selected squares, respectively, they evaluate the overall attribution map. Infidelity scales linearly with the number of samples $k$ used to approximate the expected value. In this work, $k = 1000$, which corresponds to the original implementation by \citet{Yeh2019}.

\subsection{Max-Sensitivity}
Max-Sensitivity \citep{Yeh2019} is the only metric that isn't designed to evaluate the \textit{correctness} of an attribution map, but rather the \textit{robustness} of the attribution map against small perturbations. It does this by adding small perturbations to the sample and recomputing the attribution map on the perturbed samples. The maximum value of the $L_\infty$-norm of the difference between the original and perturbed attribution map is measured. To make different attribution methods comparable, the attribution maps are normalized to unit norm before computing Max-Sensitivity.

$$
	SENS_{MAX}(\mathbf{x}, m, E, c) = \max_{\lVert \mathbf{y} - \mathbf{x} \rVert \leq r} \lVert E(m,\mathbf{x},c) - E(m,\mathbf{y},c) \rVert
$$

Where $r$ is the maximum size of the added perturbation. We choose $r = 0.1$, as in \citep{Yeh2019}. Note that this is a type II-metric, meaning that it needs access to the attribution method $E \in \mathcal{E}$ rather than just the attribution map $e \in \mathbb{R}^d$. This metric scales linearly with the number of samples $k$ (here chosen to be 50, as in \citet{Yeh2019}) used to approximate the maximum value, \textit{and} the number of forward/backward passes necessary to compute the attribution map $C_{mth}$. Note that this can result in very large runtimes when evaluating computationally complex methods.

\subsection{Impact Coverage}
\label{sec:impact-coverage}
Impact Coverage \citep{QiuLin} works by applying an adversarial patch to the image, and computing feature attributions on the adversarially attacked image. If the adversarial attack was successful, we would expect a large proportion of the attribution to be inside of the adversarial patch, as the patch caused the model to change its output.

We quantify this by computing the intersection-over-union (IOU) between the $k$ most important pixels according to the attribution map $E(m,\mathbf{x},c)$ (denoted here as the set $T$), where $k$ is the number of pixels covered by the patch, and the patch $P$ itself.
A score of 1 would indicate that the most important pixels perfectly cover the adversarial patch.

$$
	COV(\mathbf{x}, m, E, c) = \frac{\mid T \cap P\mid}{\mid T \cup P\mid}
$$

Where $P$ is the set of features that were covered by the adversarial patch. Note that this metric, like Max-Sensitivity, is also a type II metric. Impact Coverage evaluates only the high end of the attribution map. Since the attribution method needs to be executed on the attacked image, this metric has the same complexity as the method being evaluated. Note that an adversarial patch is also needed to compute this metric, meaning that this complexity is only valid when the adversarial patch is given (that is, when evaluating this metric on a large number of attribution maps for the same model). Impact Coverage can only be computed for image data with a sufficiently high resolution, such that an adversarial patch can be generated successfully.

Impact Coverage stands out from the other metrics because of its causal interpretation. As the adversarial patch was added to the image and the model was re-evaluated, we can be sure that the patch caused the change in the model output. This acts as a form of ground truth, although it is
an incomplete form: it is not guaranteed that the entire patch was necessary to change the output. Also, adversarial patches typically have strongly contrasting, high-frequency structure. This means that an attribution method that simply identifies highly contrasting, high-frequency regions in the image will likely score well on this metric, even though it might not be a good explanation of the model's behavior. Therefore, despite the causal interpretation of this metric, a good score is not a completely necessary or sufficient condition for good performance of an attribution method.

\subsection{Parameter Randomization}
The final metric we consider is the Parameter Randomization test \citep{Adebayo2018a}. This metric acts as a \textit{sanity check:} rather than scoring each explanation, the results of Parameter Randomization should be interpreted as a pass/fail-test, where passing is a minimal requirement for any method to be considered valuable. The Parameter Randomization test works by randomizing the parameters of the model and recomputing the attribution map for the randomized model. As the attribution map should highlight the features that were important to a specific model, we expect it to be dependent on the model parameters. Therefore, we expect the attribution map to change drastically when the parameters are randomized.

However, \citet{Adebayo2018a} warn against a visual inspection of the resulting attribution maps, as it is possible that features with a formerly strongly positive attribution value receive a strongly negative attribution value after randomization. In this case, a visual inspection (which in many cases shows absolute attribution values) can be misleading, as the same features can seem important after randomization, even though their attribution value has changed drastically. Therefore, the change in attribution map is quantified using the absolute value of the Spearman rank correlation coefficient between the attribution maps for the original and the randomized model. If this value is close to zero, then the method is said to pass the sanity check. Note that the authors also introduce variants of this metric using the Structural Similarity Index \citep{wang2004ssim} and Histogram of Oriented Gradients \citep{dalal2005hogs}. However, we will not consider these variants in this work, as they require the image to be divided into patches, which is not always possible (for example, when evaluating attribution methods on tabular data or low-resolution images).

Like Impact Coverage, the Parameter Randomization metric also has a causal interpretation: by randomizing the parameters, we intervene on the model, which allows us to define a form of ground truth. However, recent work suggests some possible limitations of this metric as well. \citet{yona2021} model the metric using a causal DAG, and suggest that the task on which the model was trained might act as a confounder in the causal diagram of the metric. This would imply that whether a given explanation method passes or fails the sanity check could depend on the specific task or dataset. \citet{binder2023} demonstrate that, even after partial randomization of the network, channels with high activations are still likely to have a strong contribution to the output. For this reason, we only compare the original explanation with an explanation generated for a fully randomized model. Finally, \citet{hedstrom2023a} show that the similarity metrics employed in \citet{Adebayo2018a} are minimized by a statistically uncorrelated random process. This implies that intrinsically noisy explanations, such as gradient-based methods which can be subject to shattered gradient noise \citep{balduzzi2017shattered}, might be favoured by the Parameter Randomization test.

Note that the underlying assumption from \citet{Adebayo2018a}, i.e. the idea that any useful attribution method should be sensitive to the model parameters, is not being called into question by any of these works. Instead, the works demonstrate that methods could fail the sanity checks for other reasons than invariance to model parameters, and the outcome of the sanity check might depend on the specific dataset. For this reason, we will also compute the sanity check for each dataset separately.

\section{Attribution Methods}
\label{sec:methods}
An overview of the attribution methods included in this study can be seen in Table \ref{tbl:methods}. We divide the methods in three types: Gradient-based, CAM-based and Perturbation-based. We also mention if the method requires the model to be differentiable, convolutional, or if it has no requirements about the model. The implementation used for GradCAM++ \citep{chattopadhay2018gradcampp} and ScoreCAM \citep{Wang2020ScoreCAM} is available 
in the torch-cam package \citep{Fernandez_TorchCAM_class_activation_2021}. For XRAI \citep{Kapishnikov_2019_ICCV},
an implementation is available in the Saliency package provided by PAIR\footnote{\url{https://github.com/pair-code/saliency}}. For all other methods, the Captum package
\citep{kokhlikyan2020captum} was used. For more details on the methods, we refer to
the original papers in Table \ref{tbl:methods}.

\begin{table}
	\centering
	\begin{tabular}{||c c c c||}
		\hline
		Method                            & Type         & Complexity        & Model requirements \\ [0.5ex] 
		\hline\hline
		Gradient                          & Gradient     & $\mathcal{O}(1)$  & Differentiable     \\ 
		\citep{Simonyan2014}              &              &                   &                    \\
		\hline
		InputXGradient                    & Gradient     & $\mathcal{O}(1)$  & Differentiable     \\
		\citep{Shrikumar2017}             &              &                   &                    \\
		\hline
		Deconvolution                     & Gradient     & $\mathcal{O}(1)$  & Differentiable     \\
		\citep{Zeiler2014}                &              &                   &                    \\
		\hline
		Guided Backpropagation            & Gradient     & $\mathcal{O}(1)$  & Differentiable     \\
		\citep{Springenberg2014}          &              &                   &                    \\
		\hline
		DeepLIFT                          & Gradient     & $\mathcal{O}(1)$  & Differentiable     \\
		\citep{Shrikumar2017}             &              &                   &                    \\
		\hline
		Integrated Gradients              & Gradient     & $\mathcal{O}(n)$  & Differentiable     \\
		\citep{Sundararajan2017}          &              &                   &                    \\
		\hline
		XRAI                              & Gradient     & $\mathcal{O}(n)$  & Differentiable     \\
		\citep{Kapishnikov_2019_ICCV}     &              &                   &                    \\
		\hline
		Expected Gradients                & Gradient     & $\mathcal{O}(nm)$ & Differentiable     \\
		\citep{erion2020}                 &              &                   &                    \\
		\hline
		SmoothGrad                        & Gradient     & $\mathcal{O}(m)$  & Differentiable     \\
		\citep{Smilkov2017}               &              &                   &                    \\
		\hline
		VarGrad                           & Gradient     & $\mathcal{O}(m)$  & Differentiable     \\
		\citep{Adebayo2018a}              &              &                   &                    \\
		\hline
		DeepSHAP                          & Gradient     & $\mathcal{O}(m)$  & Differentiable     \\
		\citep{Lundberg2017}              &              &                   &                    \\
		\hline
		KernelSHAP                        & Perturbation & $\mathcal{O}(m)$  & None               \\
		\citep{Lundberg2017}              &              &                   &                    \\
		\hline
		LIME                              & Perturbation & $\mathcal{O}(m)$  & None               \\
		\citep{Ribeiro2016}               &              &                   &                    \\
		\hline
		GradCAM                           & CAM          & $\mathcal{O}(1)$  & Convolutional      \\
		\citep{Selvaraju2017}             &              &                   &                    \\
		\hline
		Guided GradCAM                    & CAM          & $\mathcal{O}(1)$  & Convolutional      \\
		\citep{Selvaraju2017}             &              &                   &                    \\
		\hline
		GradCAM++                         & CAM          & $\mathcal{O}(1)$  & Convolutional      \\
		\citep{chattopadhay2018gradcampp} &              &                   &                    \\
		\hline
		ScoreCAM                          & CAM          & $\mathcal{O}(c)$  & Convolutional      \\
		\citep{Wang2020ScoreCAM}          &              &                   &                    \\
		\hline
	\end{tabular}
	\caption{Summary of methods. Complexity is expressed as number of executions of the model. $n$, $m$ and $c$ are path length, number of perturbed samples/baselines, and number of channels in the final layer, respectively. $n$ and $m$ are hyperparameters of the method, $c$ depends on the model being explained. Note that, even though many methods have the same asymptotic complexity, the typical values of hyperparameters can vary a lot, for example DeepSHAP usually needs much fewer samples than KernelSHAP or LIME, making it computationally less expensive.}
	\label{tbl:methods}
\end{table}
\section{Experimental Setup}
In this section, we describe the datasets used in the experiments, the different implementations of feature masking, and the methods of statistical analysis that we performed on the metric scores. For a demonstration of the methodology on tabular datasets, see Appendix \ref{app:tabular}.

\subsection{Datasets}
All experiments were conducted on 14 attribution methods and 8 datasets. The datasets can be divided into three groups:

\begin{itemize}
	\item Low-dimensional datasets (28x28x1): MNIST \citep{lecun1998}, FashionMNIST \citep{xiao2017}
	\item Medium-dimensional datasets (32x32x3): CIFAR-10, CIFAR-100 \citep{krizhevsky2009}, SVHN \citep{netzer2011}
	\item High-dimensional datasets (224x224x3): ImageNet \citep{deng2009}, Caltech-256 \citep{griffin2022}, Places-365 \citep{zhou2017places}
\end{itemize}

For the low-dimensional datasets, a simple CNN architecture (2 convolutional layers with 32 and 64 channels, followed by a fully connected hidden layer with 128 nodes) was trained. For the medium- and high-dimensional datasets, we used Resnet20 and Resnet18, respectively. The models for the low- and medium-dimensional datasets were trained up to a test set accuracy of at least 90\%, except for CIFAR-100, where a top-five accuracy of 90.6\% was reached. For Caltech-256 and Places-365, the models were trained up to a top-five test set accuracy of 91.6\% and 83.7\%, respectively. For ImageNet, the built-in Resnet18 model of torchvision\footnote{\href{https://pytorch.org/vision/stable/models.html}{https://pytorch.org/vision/stable/models.html}} was used, obtaining a top-five accuracy of 89.08\%. The metric scores were computed for all attribution methods on 256 correctly-classified samples for each dataset. Note that an adversarial patch was only generated for the high-dimensional datasets (ImageNet, Caltech-256, Places-365), which means that the Impact Coverage could only be computed for these datasets.

\subsection{Masking}
\label{sec:Masking}
Except for Infidelity, Impact Coverage, and the Parameter Randomization test, all metrics depend in some way on the \textit{masking} of features to remove information. When masking features, we try to replace the feature value with some ``neutral'' value that is expected to remove the original information contained in the feature. However, the choice of this neutral value is not obvious \citep{Sturmfels2020}.
We consider three options:

\begin{itemize}
	\item \textbf{Dataset mean:} The first and simplest way of masking is by replacing the feature with a constant zero value (in the case of color images, we do the same for each color channel). Since the data is $z$-normalized to have $\mu=0$ and $\sigma=1$, this is equivalent to changing the feature into the average feature value over the training dataset. A disadvantage of this technique is that, if the original feature was already close to the average value, the value remains nearly unchanged after masking and the information might not be properly destroyed. Specifically for image data, masking out large regions with a constant value can preserve some of the spatial information in the image. Additionally, masking features with any constant value can introduce high-frequency artifacts to the image in question, driving the input away from the data manifold \citep{Fong2017}.
	\item \textbf{Uniform random:} To mitigate some of the problems of the dataset mean value, we can also draw values from a standard uniform distribution $\mathcal{U}(0,1)$. That way, masked out features are less likely to remain unchanged after masking, and spatial information is likely to be successfully destroyed if larger regions of the image need to be masked. However, using a uniform distribution to mask out features introduces even more adversarial high-frequency artifacts than using a constant dataset mean.
	\item \textbf{Blur:} To reduce the high-frequency artifact problem of the first two masking procedures, pixels can instead be masked out using blurring \citep{Fong2017}. We use the OpenCV normalized box filter \citep{opencv_library} with kernel size $k=0.5$ to produce a blurred version of the original image. Pixels are then masked out by replacing them by their blurred equivalents. Although this technique mitigates the high-frequency artifact problem, it again has the disadvantage that spatial information might not be completely destroyed after masking.
\end{itemize}

\subsection{Statistical Analysis}
In this section we provide a brief overview of the statistical techniques used to analyse our results. We first use a paired t-test to identify which methods outperform a basic random baseline on the metrics. Next, we compute the correlations of scores between different metrics, which allows us to measure which metrics are or are not measuring the same underlying properties. We then study the consistency of method rankings as given by each metric, which can be viewed as a quality check for the metrics themselves. Finally, we propose a technique to compare two methods in more detail.

\subsubsection{Statistical Significance Testing}
\label{sec:analysis:stat_sig}
For each metric and each method, we use a paired t-test to verify if the method performs significantly better than a uniform baseline on the given metric. More specifically, we test if the difference in metric score for the explanation method is significantly larger/smaller (depending on the metric in question) than the score obtained by the uniform baseline. The uniform baseline is defined as a ``pseudo-method'', which simply assigns random values $u \sim \mathcal{U}(0,1)$ to each feature. This baseline is computed once for every input sample, such that the same baseline attribution map is compared to each of the attribution maps computed by the explanation methods. Note that a different, more informative baseline method could also be used. For example, a simple edge detection algorithm could be used to establish a more competitive baseline, while retaining the property that any valid explanation method should be expected to outperform it. Alternatively, an existing explanation method could be used as the baseline, for example to test whether some other method specifically outperforms that baseline method. We leave a further investigation of different baseline methods to future work.

Because for each metric, multiple methods are tested against the random baseline, we use Bonferroni multiple testing correction \citep{Bland170}. If the result of the test is significant after correction ($p < 0.01$), we report the Cohen's $d$ effect size for paired t-tests \citep{Cohen1988}:
$$
	\frac{\mu_d}{\sigma_d}
$$
which is simply the average difference in metric scores divided by the standard deviation of the differences. Since the absolute values of most metrics carry little to no semantic meaning, these effect sizes are only relevant relative to each other. For this reason, the effect sizes are scaled to $[0,1]$ for each metric, such that the best-performing method has an effect size of 1.

\subsubsection{Inter-Metric Correlation}
Inter-metric correlations are computed as the Spearman rank correlation between metric scores, averaged over all methods (except the random baseline). These correlations allow us to identify which metrics are measuring different underlying aspects, and which metrics are mutually redundant.

\subsubsection{Ranking Consistency}
Ranking consistency assesses how consistent a metric is in ranking the methods across the different images. This is measured using Krippendorff's $\alpha$ \citep{krippendorff2018content}. Krippendorff's $\alpha$ is a statistic usually used to measure \textit{inter-rater reliability}: the degree to which different \textit{raters} (for example, for a psychological test) agree in their assessments. Krippendorff's $\alpha$ is defined as follows:
$$
	\alpha = 1 - \frac{D_o}{D_e}
$$
where $D_o$ is the \textit{observed disagreement}, and $D_e$ is the \textit{disagreement expected by chance}. 
If $\alpha = 1$, then the ranking is perfectly consistent: the ranking of methods produced by the metric is identical for each image.
If $\alpha = 0$, then the ranking is completely random.
For more details on how these values are computed, we refer the reader to \citet{krippendorff2011computing}.

\subsubsection{Pairwise Comparison Using PoS}
\label{sec:cles}
Once a global overview of method performance has been established using the statistical significance test (Section \ref{sec:analysis:stat_sig}), two or more methods can be selected for a more detailed comparison. Such a comparison is then made by performing a new statistical test, this time comparing the methods to each other, rather than to a trivial random baseline.

In this case, we use the Power of Superiority (PoS) effect size to measure the difference between methods. This measure is simply the fraction of images where method A outperforms method B. This effect size measure is less informative when comparing methods to the random baseline, as we expect methods to at least consistently outperform the baseline, leading to a saturated effect size of 1. If two methods are selected that are more similar in their performance, the PoS can give an intuition to how often one method (usually a more computationally complex one) outperforms the other. 

For example, if the difference in metric scores is very large, but the PoS is only slightly larger than 0.5, this would mean that method A outperforms method B only in a small majority of images. If computational cost is a concern, this can make it more interesting to choose for the computationally cheaper method.

\section{Results}
In this section, we describe the results of paired t-tests, inter-metric correlations, and ranking consistency of metric scores on the different datasets. Masking is done using the dataset mean approach unless stated otherwise. Finally, we perform a pairwise comparison of DeepSHAP and DeepLIFT on MNIST, CIFAR-10 and ImageNet.

\subsection{Paired t-tests}
\label{sec:results:ttest}

\begin{figure}
	\centering
	\begin{subfigure}[b]{0.49\textwidth}
		\includegraphics[width=\textwidth]{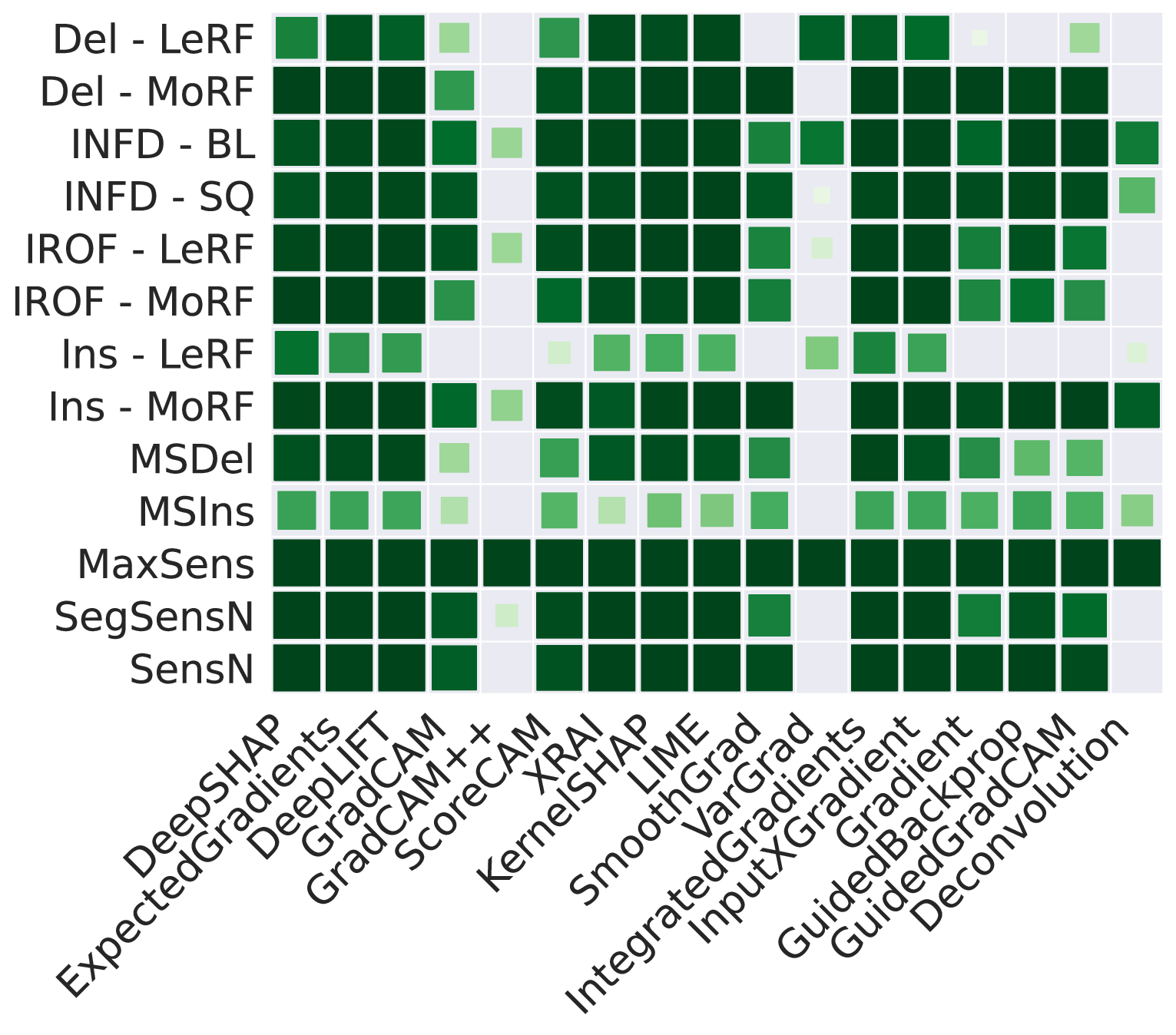}
		\caption{MNIST}
		\label{fig:wsp-mnist}
	\end{subfigure}
	\begin{subfigure}[b]{0.49\textwidth}
		\includegraphics[width=\textwidth]{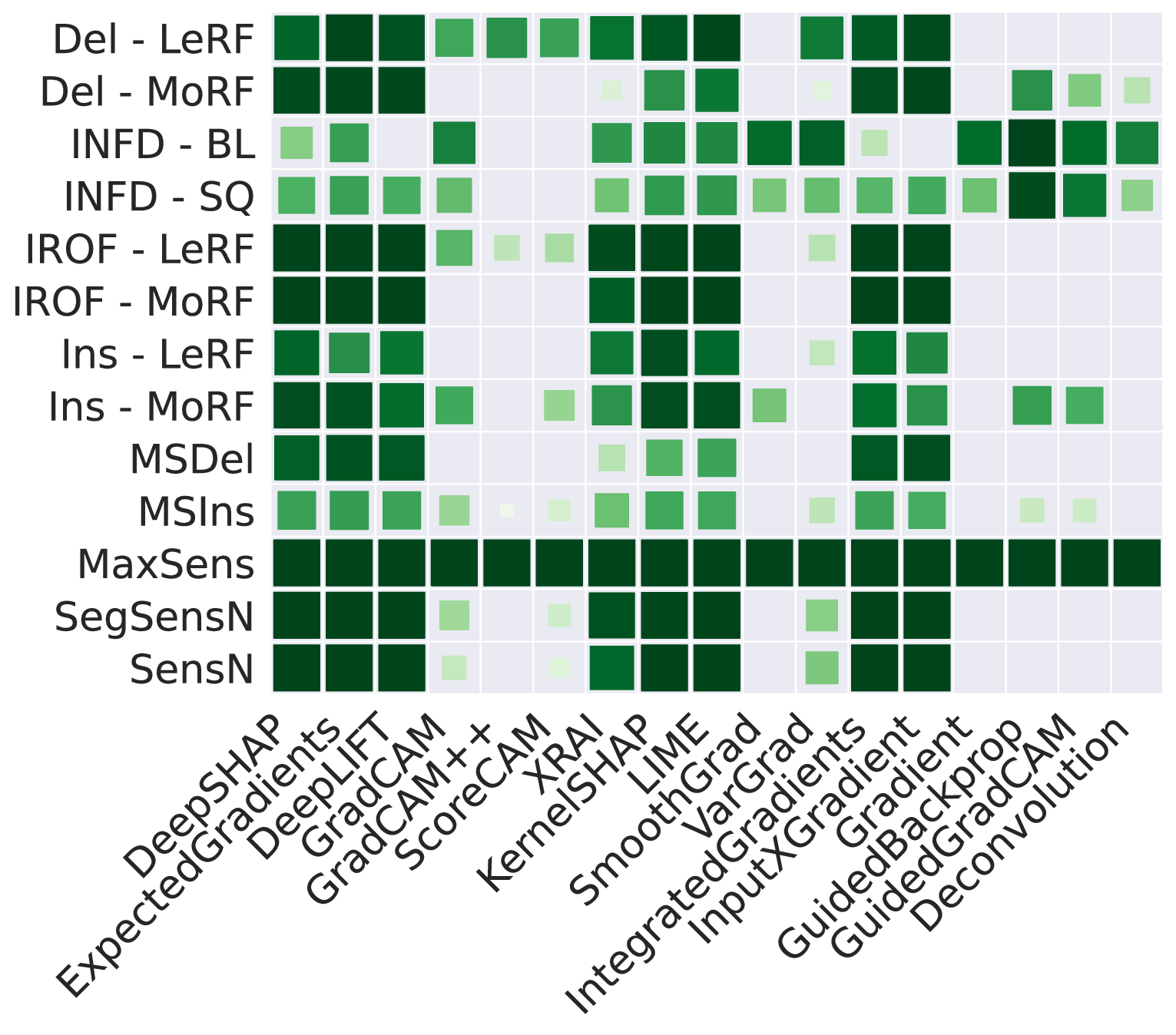}
		\caption{FashionMNIST}
		\label{fig:wsp-fashionmnist}
	\end{subfigure}
	\caption{Results of paired t-tests (low-dimensional datasets). A square is only drawn if the corresponding result was significant after Bonferroni correction ($p < 0.01$).}
	\label{fig:wsp-low}
\end{figure}

\begin{figure}
	\centering
	\begin{subfigure}[b]{0.49\textwidth}
		\includegraphics[width=\textwidth]{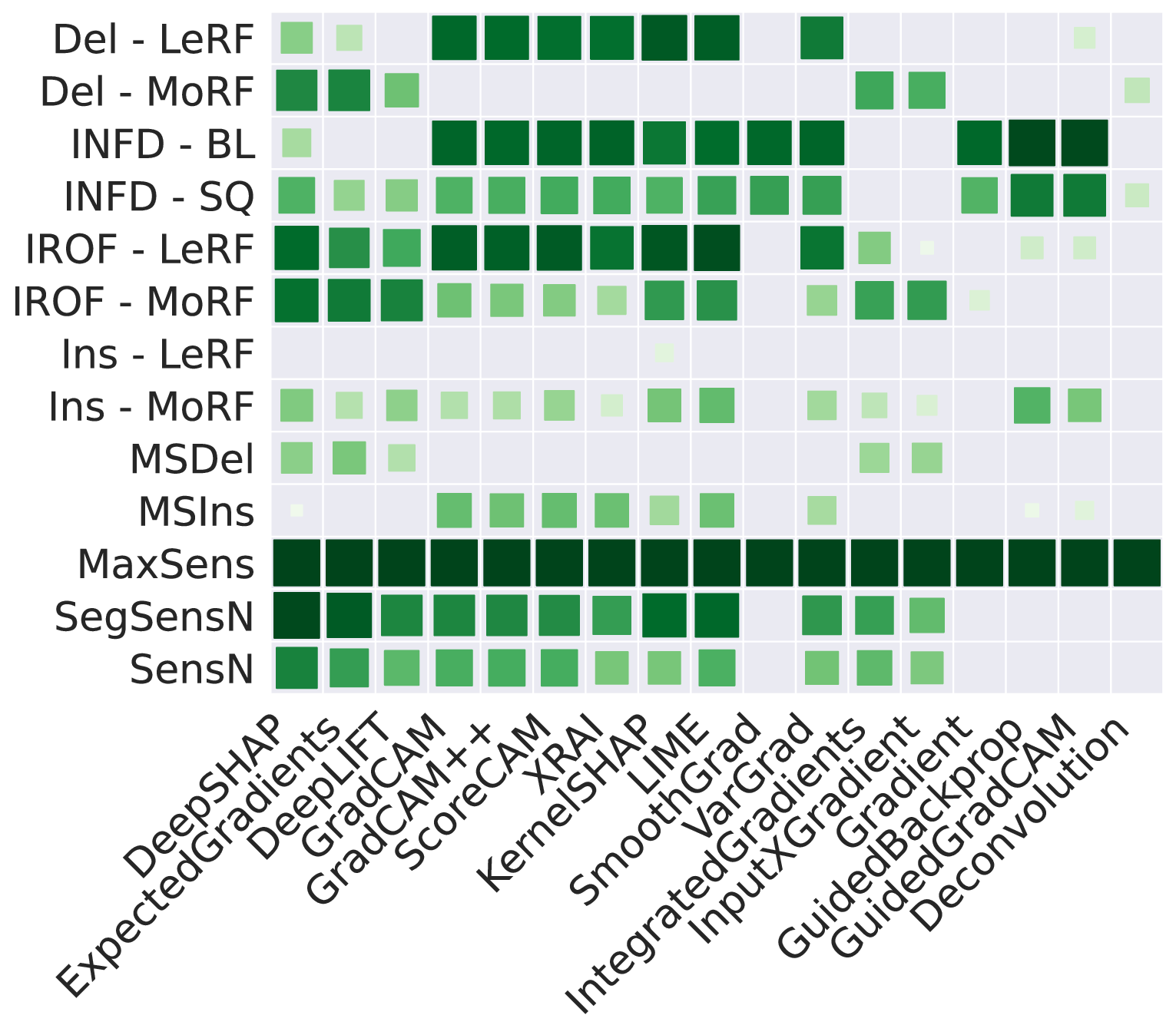}
		\caption{CIFAR-10}
		\label{fig:wsp-cifar10}
	\end{subfigure}
	\begin{subfigure}[b]{0.49\textwidth}
		\includegraphics[width=\textwidth]{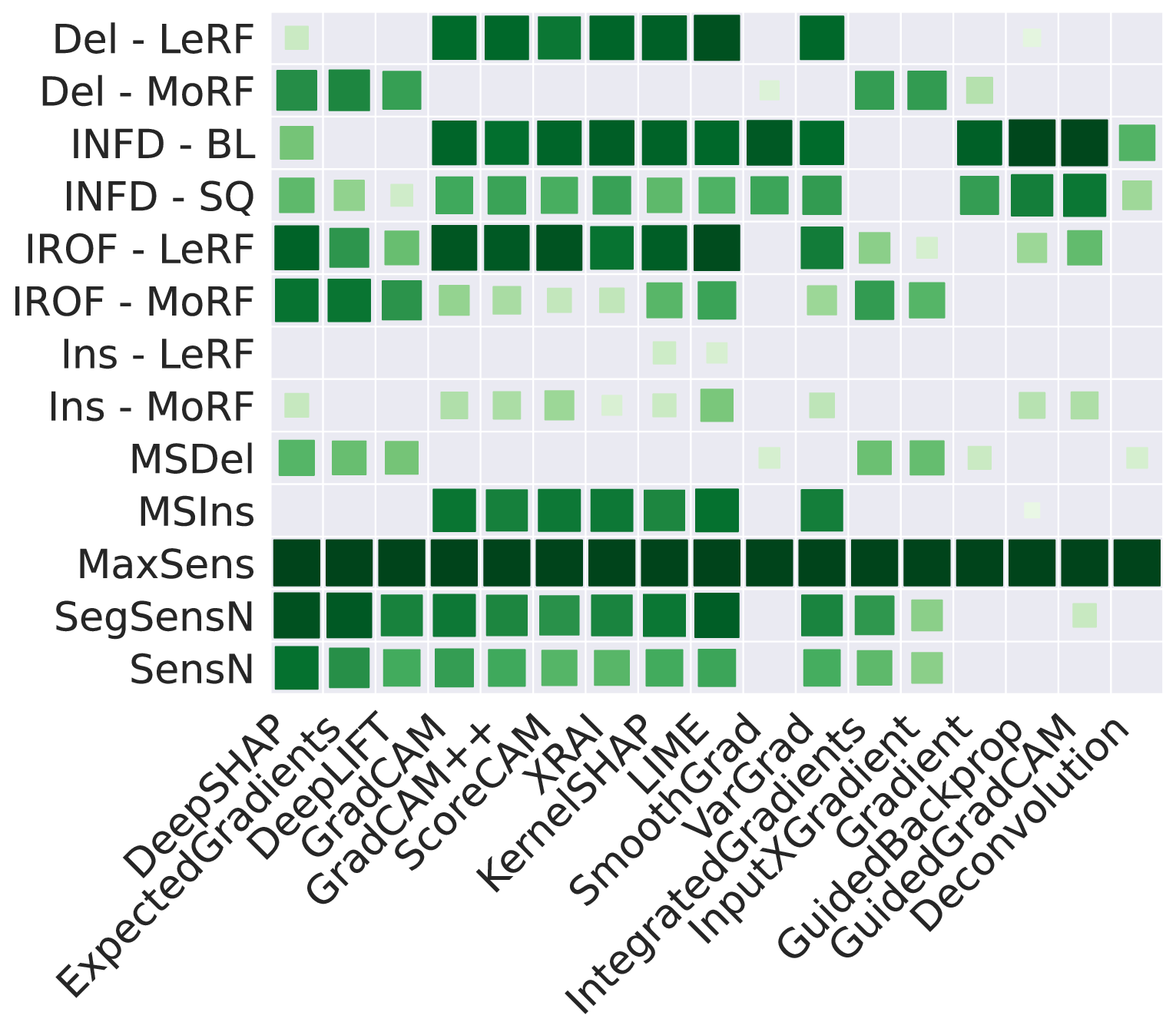}
		\caption{CIFAR-100}
		\label{fig:wsp-cifar100}
	\end{subfigure}
	\begin{subfigure}[b]{0.49\textwidth}
		\includegraphics[width=\textwidth]{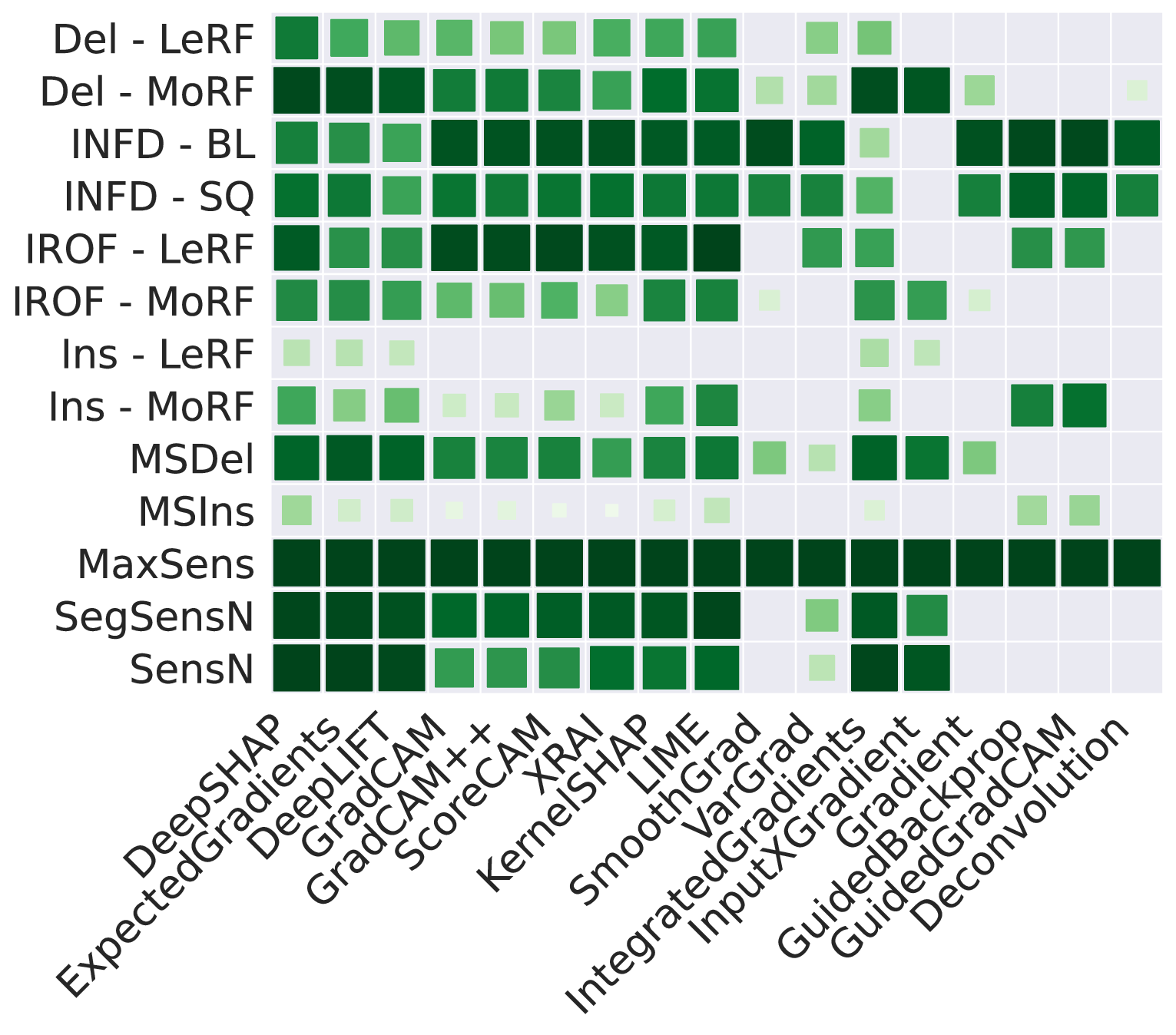}
		\caption{SVHN}
		\label{fig:wsp-SVHN}
	\end{subfigure}
	\caption{Results of paired t-tests (medium-dimensional datasets). A square is only drawn if the corresponding result was significant after Bonferroni correction ($p < 0.01$).}
	\label{fig:wsp-medium}
\end{figure}

\begin{figure}
	\centering
	\begin{subfigure}[b]{0.49\textwidth}
		\includegraphics[width=\textwidth]{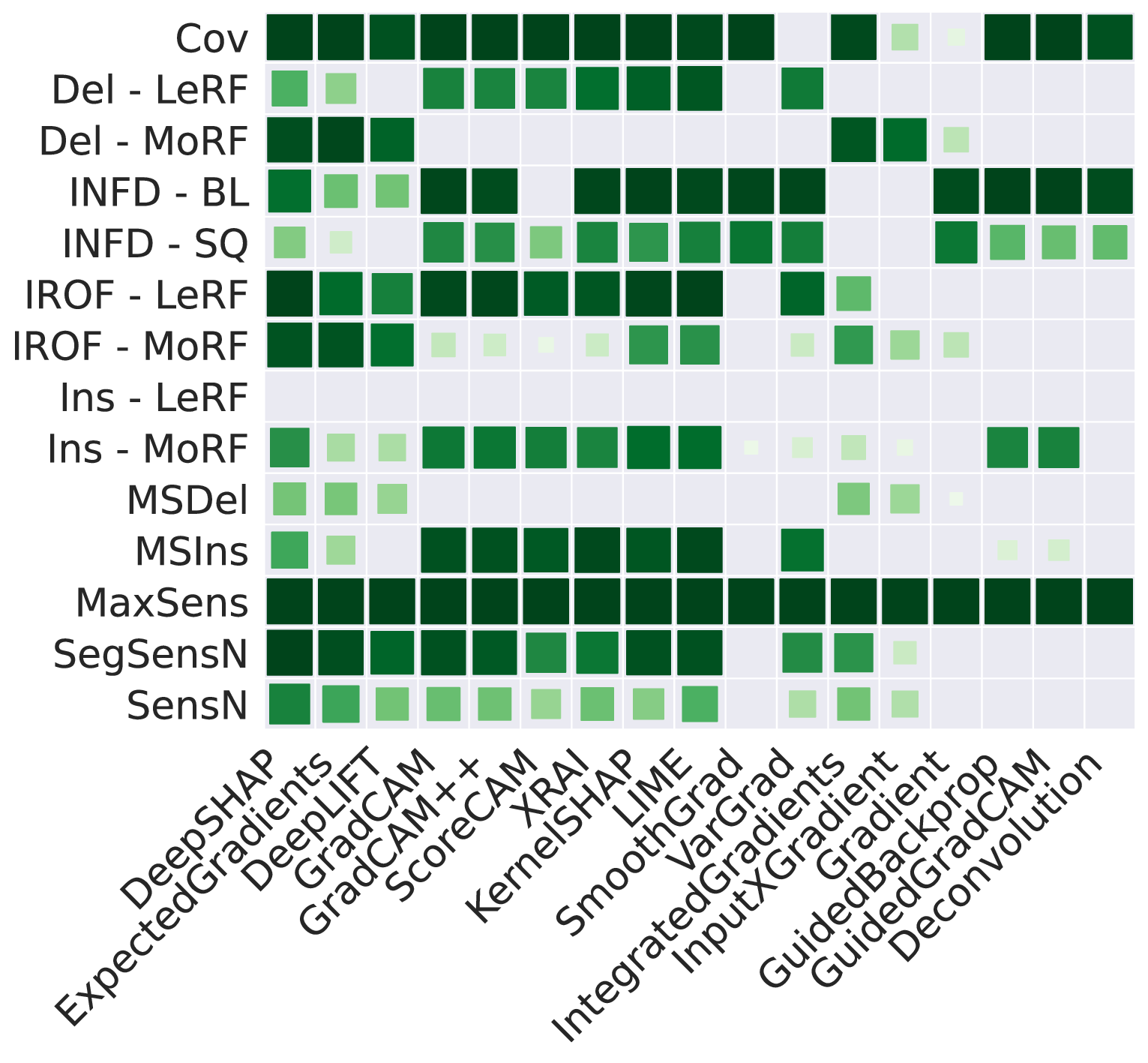}
		\caption{ImageNet}
		\label{fig:wsp-imagenet}
	\end{subfigure}
	\begin{subfigure}[b]{0.49\textwidth}
		\includegraphics[width=\textwidth]{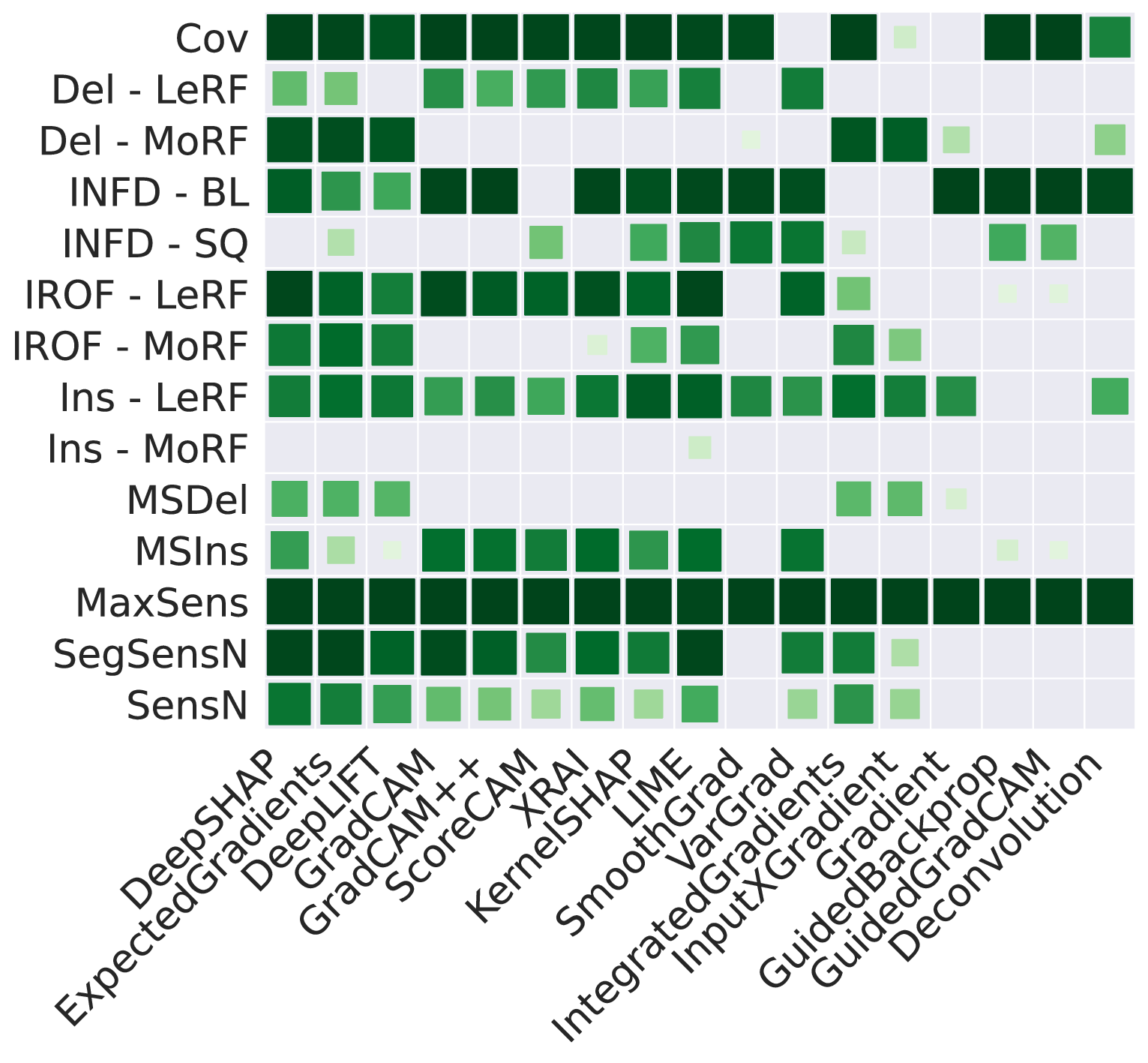}
		\caption{Caltech-256}
		\label{fig:wsp-caltech}
	\end{subfigure}
	\begin{subfigure}[b]{0.49\textwidth}
		\includegraphics[width=\textwidth]{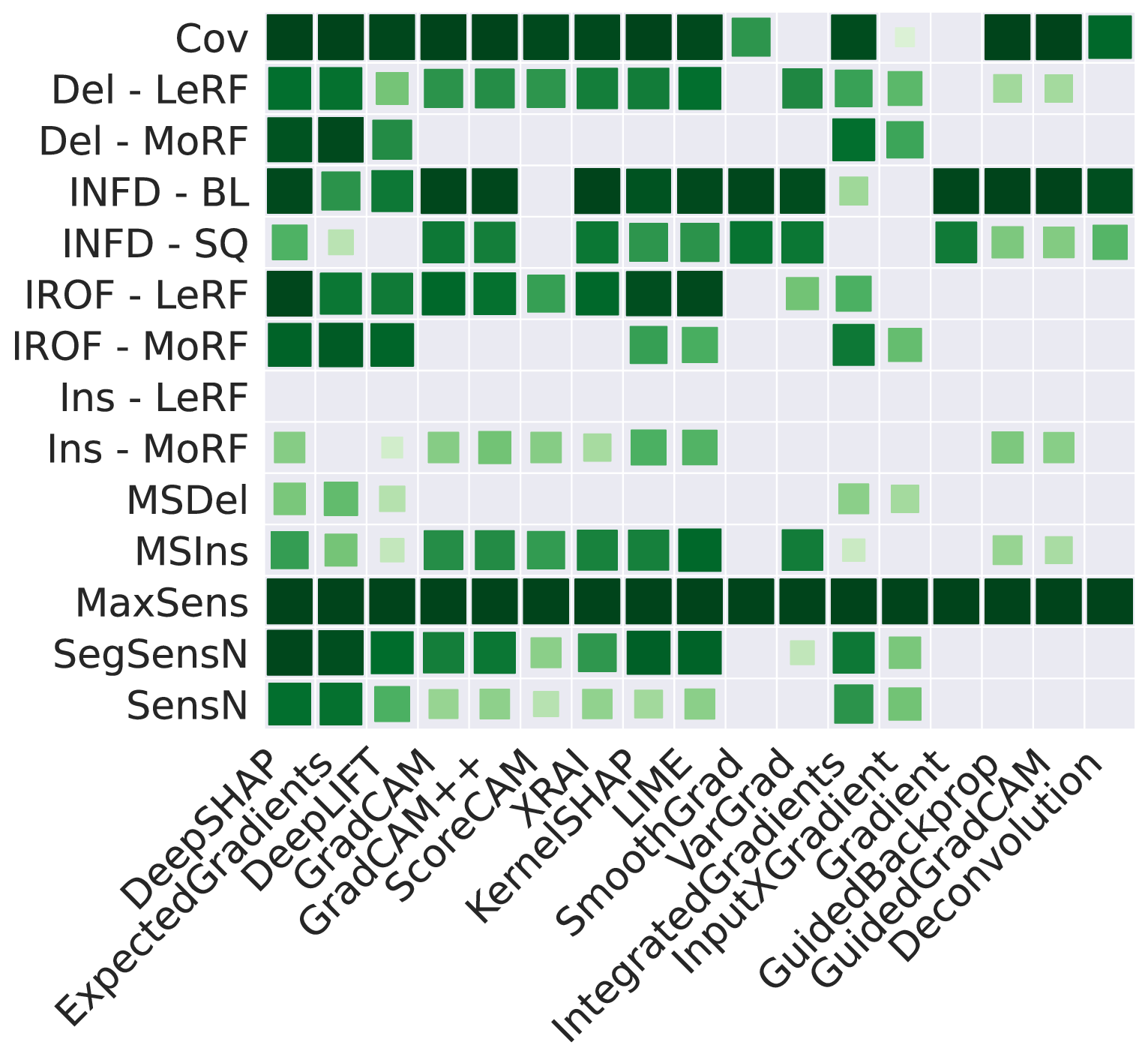}
		\caption{Places-365}
		\label{fig:wsp-places}
	\end{subfigure}
	\caption{Results of paired t-tests (high-dimensional datasets). A square is only drawn if the corresponding result was significant after Bonferroni correction ($p < 0.01$).}
	\label{fig:wsp-high}
\end{figure}

The results of the paired t-tests are shown in Figures \ref{fig:wsp-low}, \ref{fig:wsp-medium} and \ref{fig:wsp-high}, for the low-, medium- and high-dimensional datasets, respectively. For each method-metric pair, a square is drawn if the result of the paired t-test is significant after Bonferroni correction for multiple testing ($p < 0.01$), with the size and color of the square indicating the effect size (Cohen's $d$). Effect sizes are normalized such that a value of 1 corresponds to the largest effect size for a given metric.

On MNIST, most methods significantly outperform the random baseline on nearly all metrics. On FashionMNIST however, CAM-based methods, Guided Backpropagation, Deconvolution, SmoothGrad and VarGrad perform significantly worse than the others. 

On the medium-dimensional datasets, we see more complementarity in the results, although this still depends on the dataset. On CIFAR-10 and CIFAR-100, we notice that the CAM-based methods along with XRAI, KernelSHAP and LIME perform very similarly, with this group of metrics outperforming DeepSHAP, DeepLIFT and ExpectedGradients on some metrics and vice versa. This similarity in behaviour can be linked to the fact that these methods produce more coarse-grained attribution maps, as all CAM-based methods rely on upsampling the final convolutional layer, and XRAI, KernelSHAP and LIME rely on image segmentation. Conversely, DeepLIFT, DeepSHAP and ExpectedGradients are all based on modified versions of the gradient, which tends to produce very granular attribution maps. This difference in granularity could be the source of the observed complementarity. For SVHN, a number of methods significantly outperform the baseline across all metrics.

In the high-dimensional case, we see fewer differences between the datasets. The same complementarity between the coarse-grained and fine-grained methods is again noticeable for all three datasets, suggesting that it is linked to the complexity or dimensionality of the classification problem. 

We also note that the coarse-grained methods tend to outperform the others on Impact Coverage (COV), which could only be computed for high-dimensional datasets. Interestingly, the results for Impact Coverage seem to be complementary to those of Deletion-MoRF and Minimal Subset Deletion. A possible explanation is the fact that Impact Coverage makes the implicit assumption that the entire adversarial patch is equally important, which is not necessarily the case, as discussed in Section \ref{sec:impact-coverage}. This might bias the metric towards coarse-grained attribution methods. If only a few pixels in the adversarial patch are truly important, then fine-grained attribution maps might highlight only those few pixels, resulting in a low Impact Coverage score. Further research is needed to confirm or refute this hypothesis.

We draw three conclusions from these results:
\begin{enumerate}
	\item Depending on the dataset, very simple and computationally cheap methods can perform nearly as well as computationally more expensive methods.
	\item Complementarity between methods, where some methods outperform other methods on a subset of metrics and vice versa, suggests that a combination of attribution maps given by different methods might provide more information than the individual attribution maps. This is related to the idea proposed in \citet{Tomsett} that different metrics might be measuring different underlying aspects of the attribution maps.
	\item The complementarity of results between coarse-grained and fine-grained attribution maps suggests that these methods might have to be evaluated in fundamentally different ways, focusing on single-pixel importance for fine-grained maps and on a more high-level view for coarse-grained maps.
\end{enumerate}

\subsection{Inter-Metric Correlations}
\label{sec:results-inter-metric-correlation}

\begin{figure}
	\centering
	\captionsetup[subfigure]{justification=centering}
	\begin{subfigure}[b]{0.32\textwidth}
		\includegraphics[width=\textwidth]{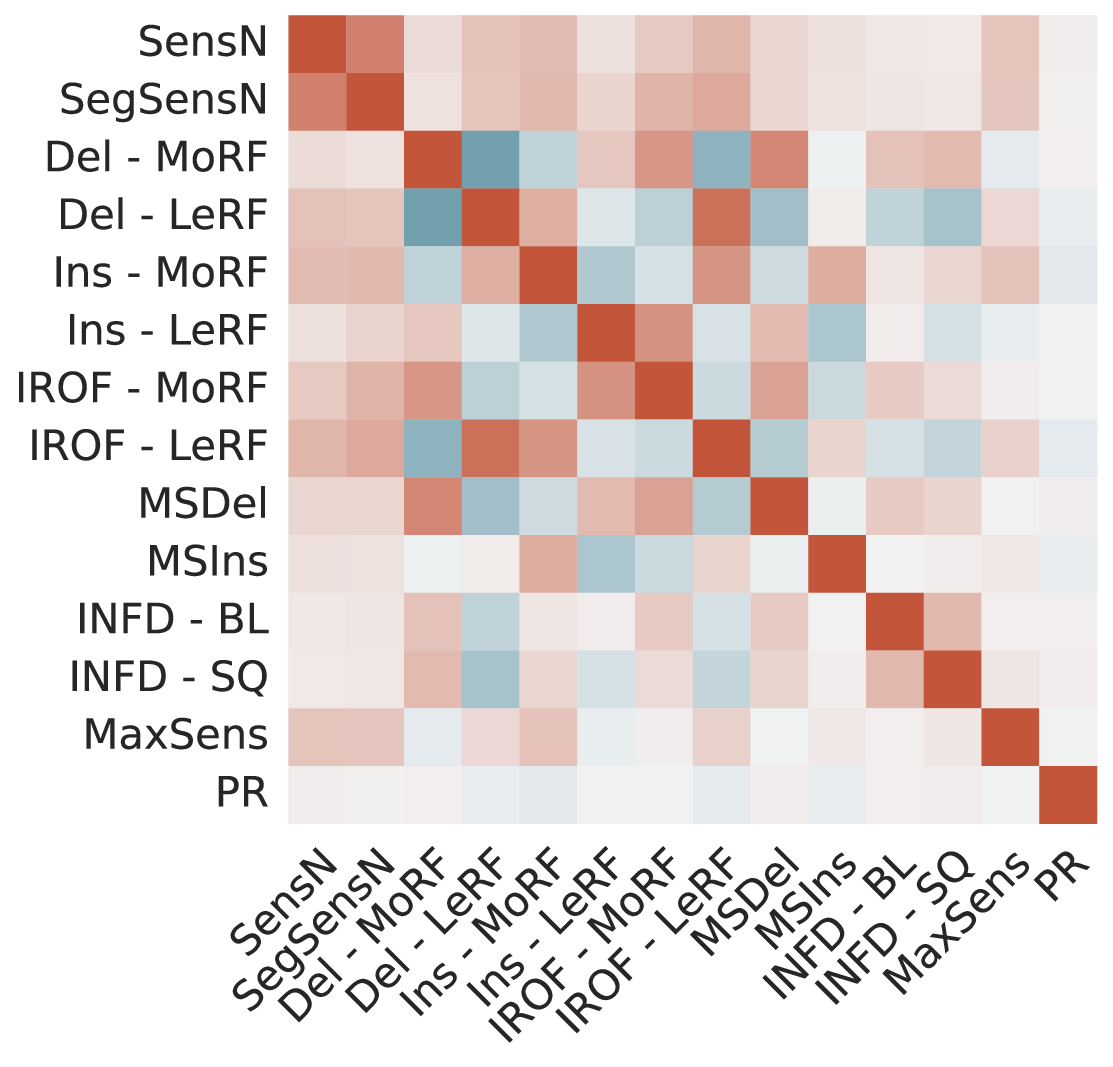}
		\caption{Low-dimensional datasets}
		\label{fig:mtr-corr-low}
	\end{subfigure}
	\begin{subfigure}[b]{0.32\textwidth}
		\includegraphics[width=\textwidth]{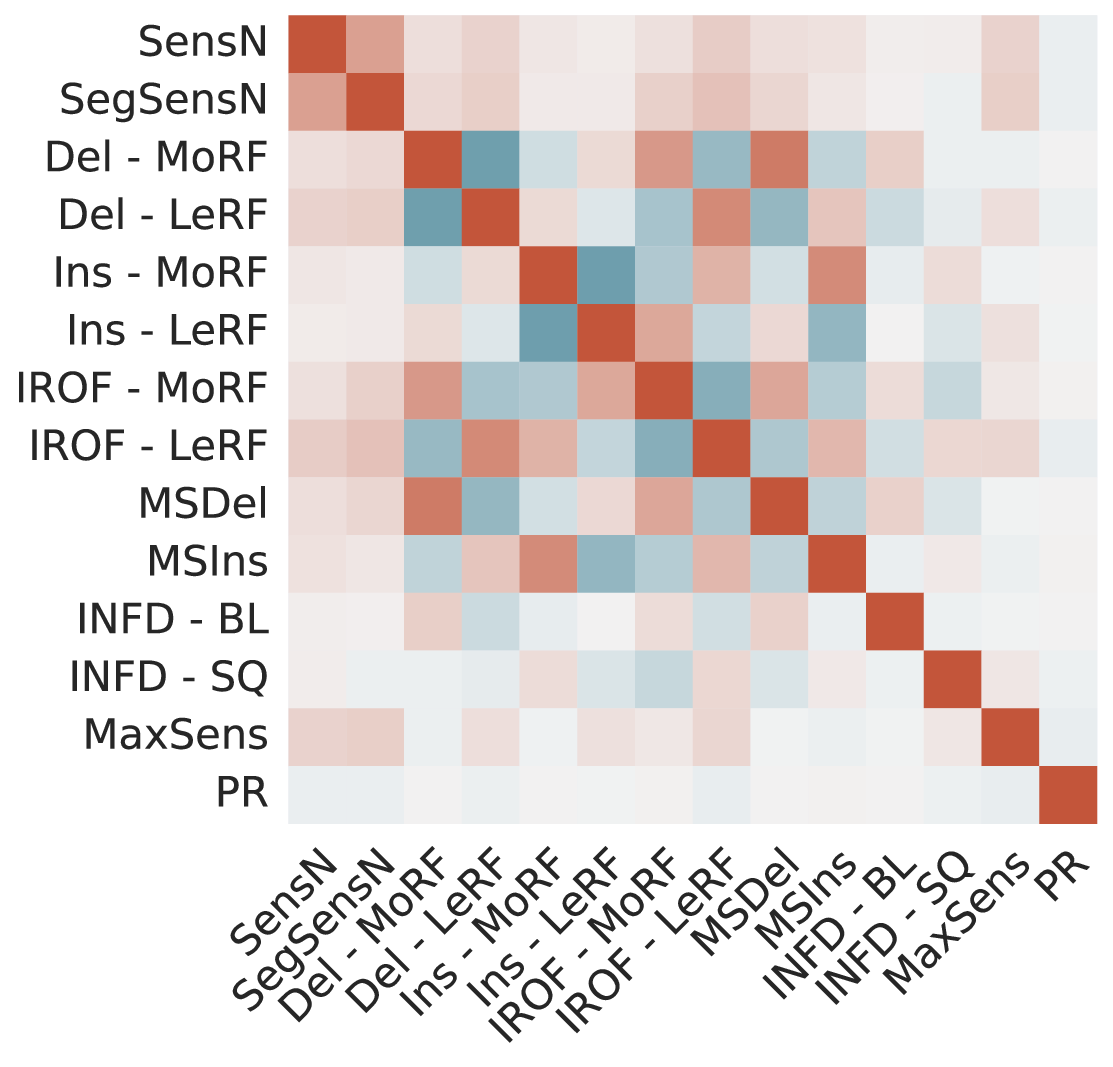}
		\caption{Medium-dimensional datasets}
		\label{fig:mtr-corr-med}
	\end{subfigure}
	\begin{subfigure}[b]{0.32\textwidth}
		\includegraphics[width=\textwidth]{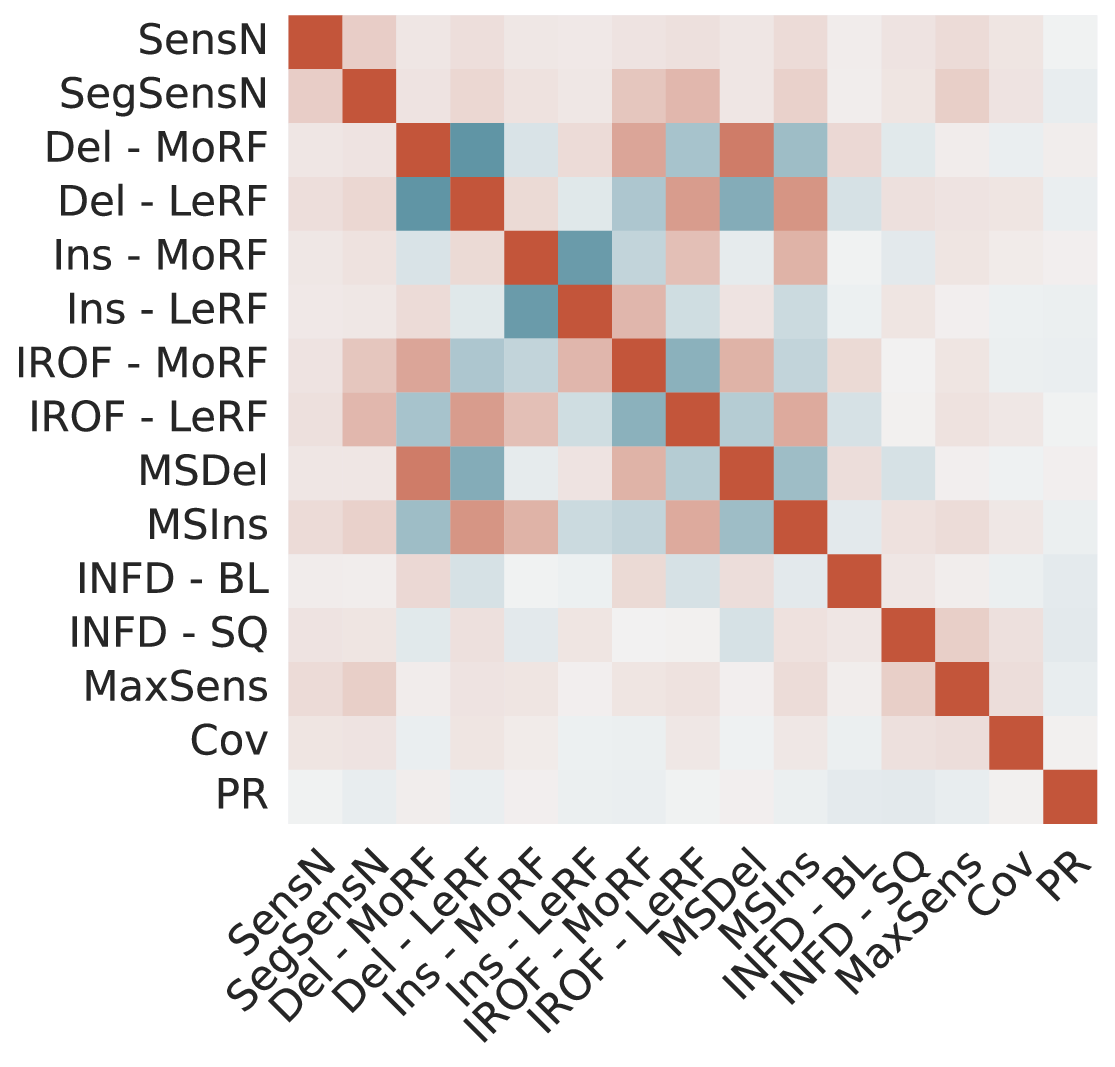}
		\caption{High-dimensional datasets}
		\label{fig:mtr-corr-high}
	\end{subfigure}
	\caption{Average inter-metric correlations for low-, medium- and high-dimensional datasets. Impact Coverage (Cov) was only computed for the high-dimensional datasets due to the requirement of an adversarial patch (see Section \ref{sec:impact-coverage})}
	\label{fig:mtr-corr}
\end{figure}

Figure \ref{fig:mtr-corr} shows average inter-metric correlations for the low-, medium- and high-dimensional datasets. For specific correlations per dataset, see Appendix \ref{app:corr-all}.
In general, we note similar patterns of correlations for the three dimensionalities. Most metrics have relatively low correlations, suggesting that they might be measuring different underlying aspects of the attribution maps, as proposed in \citet{Tomsett}. We also note strong negative correlations between certain pairs of metrics, more specifically MoRF/LeRF-pairs, which suggests that MoRF/LeRF-pairs contain largely redundant information. This insight can be used to reduce computational cost in future benchmarking efforts, by selecting only MoRF or LeRF metrics. Interestingly, figure \ref{fig:mtr-corr-high} shows very little correlation between the Impact Coverage and Parameter Randomization metrics. This is notable, as both of these metrics have a causal interpretation, even though those causal interpretations are different: Impact Coverage intervenes on the data, whereas Parameter Randomization intervenes on the model. Finally, we note that correlations between segmented and non-segmented metrics (for example, Deletion and IROF) are stronger for low-dimensional datasets. This is to be expected, since the low dimensionality of the data causes segments to be composed only of a few pixels.

Table \ref{tbl:mtr-corr-masking} shows inter-metric correlations of different metric implementations on ImageNet (results on the other datasets were generally similar). We note that, although different metrics have relatively low correlations, correlations between different implementations of the same metric are generally quite high. We can conclude from this that different implementations of the same metric generally provide redundant information. We recommend first deciding which masking procedure makes most sense for a given dataset and/or model, rather than performing full measurements using a large number of masking procedures.

\begin{table}
	\centering
	\begin{tabular}{l || c c c | c c c | c c c | c c c | }
		  & \multicolumn{3}{|c|}{$Del_{MoRF}$}  & \multicolumn{3}{|c|}{$Del_{LeRF}$}  & \multicolumn{3}{|c|}{$Ins_{MoRF}$} & \multicolumn{3}{|c|}{$Ins_{LeRF}$}                                                         \\
		\hline
		  & C                                   & B                                   & R                                  & C                                  & B    & R    & C    & B    & R    & C    & B    & R    \\
		\hline
		C & 1.00                                &                                     &                                    & 1.00                               &      &      & 1.00 &      &      & 1.00 &      &      \\
		\hline
		B & 0.87                                & 1.00                                &                                    & 0.83                               & 1.00 &      & 0.89 & 1.00 &      & 0.89 & 1.00 &      \\
		\hline
		R & 0.85                                & 0.75                                & 1.00                               & 0.82                               & 0.69 & 1.00 & 0.54 & 0.60 & 1.00 & 0.56 & 0.62 & 1.00 \\
		
		\multicolumn{13}{ c }{ }                                                                                                                                                                                        \\
		
		  & \multicolumn{3}{|c|}{$IROF_{MoRF}$} & \multicolumn{3}{|c|}{$IROF_{LeRF}$} & \multicolumn{3}{|c|}{$MS_{Del}$}   & \multicolumn{3}{|c|}{$MS_{Ins}$}                                                           \\
		\hline
		  & C                                   & B                                   & R                                  & C                                  & B    & R    & C    & B    & R    & C    & B    & R    \\
		\hline
		C & 1.00                                &                                     &                                    & 1.00                               &      &      & 1.00 &      &      & 1.00 &      &      \\
		\hline
		B & 0.92                                & 1.00                                &                                    & 0.90                               & 1.00 &      & 0.81 & 1.00 &      & 0.78 & 1.00 &      \\
		\hline
		R & 0.79                                & 0.76                                & 1.00                               & 0.75                               & 0.72 & 1.00 & 0.83 & 0.70 & 1.00 & 0.70 & 0.60 & 1.00 \\
	\end{tabular}
	\caption{Inter-metric correlations of different implementations of metrics on ImageNet. C, B and R stand for Constant, Blur and Random masking, respectively.}
	\label{tbl:mtr-corr-masking}
\end{table}

\subsection{Ranking Consistency}

\begin{figure}
	\centering
	\includegraphics[width=\textwidth]{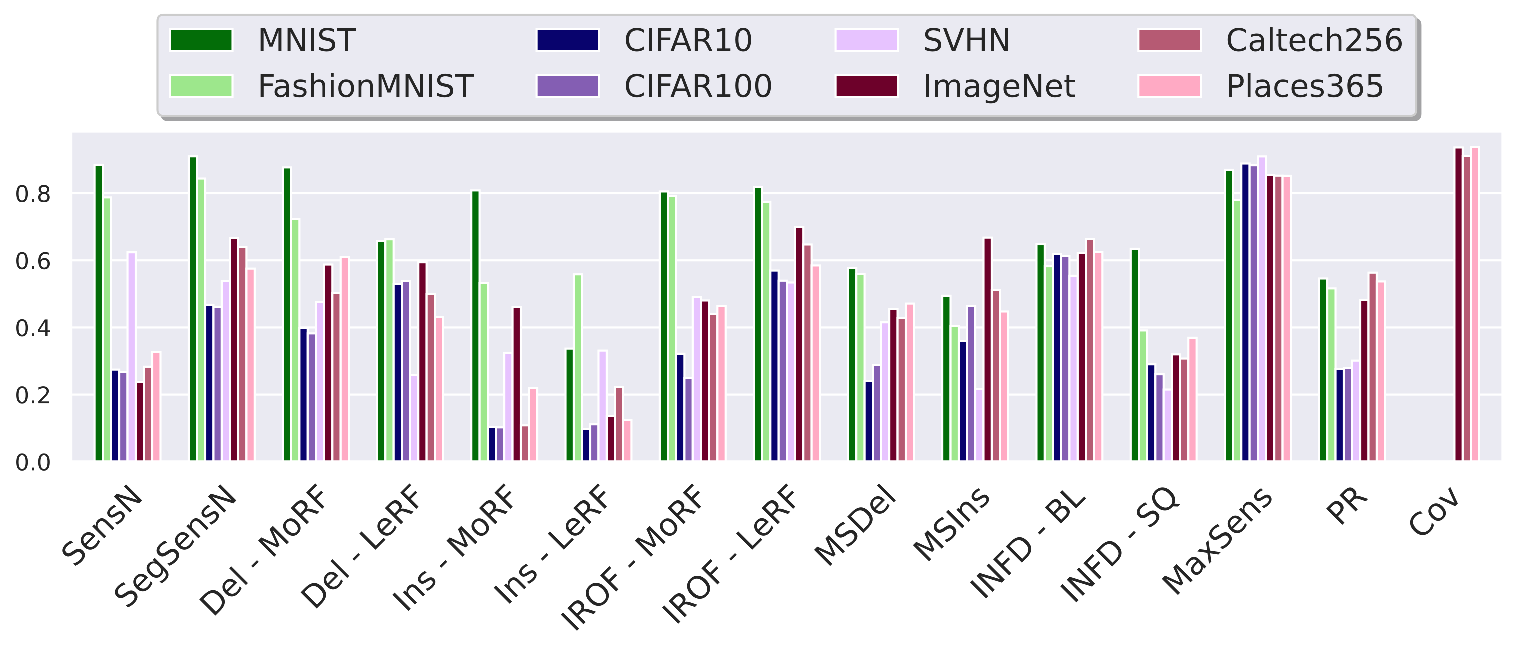}
	\caption{Krippendorff's $\alpha$ for default implementations of different metrics on all datasets. Low-, medium- and high-dimensional datasets are indicated in green, blue and red tones, respectively.  Impact Coverage (Cov) was only computed for the high-dimensional datasets due to the requirement of an adversarial patch (see Section \ref{sec:impact-coverage})}
	\label{fig:alpha-bar}
\end{figure}

The values of $\alpha$ for all datasets are shown in Figure \ref{fig:alpha-bar}. It can be observed that most of the metrics are most consistent on the low-dimensional datasets (MNIST, FashionMNIST). Impact Coverage was only measured for high-dimensional datasets because of the reliance on an adversarial patch, and has the highest values of $\alpha$.
We also see that there is no clear pattern between the medium- and high-dimensional datasets, implying that $\alpha$ doesn't simply decrease with increasing dimensionality. We note that our proposed segmented variant of Sensitivity-n has a higher $\alpha$ for high-dimensional datasets, confirming the intuition that this metric has a higher signal-to-noise ratio for high-dimensional data.

Although there are some metrics that have a significantly lower $\alpha$ value across all or almost all datasets, such as $Ins_{MoRF}$, $Ins_{LeRF}$, there is no clear subset of metrics that is generally superior to all others in terms of ranking consistency, with two exceptions: Max-Sensitivity and Impact Coverage. However, Max-Sensitivity measures robustness of explanations rather than correctness, and Impact Coverage can only be computed for high-dimensional datasets. From these results, we conclude that the ideal subset of metrics to measure depends on the dataset and model. Different implementations of the same metric (using different masking procedures) generally have similar values for $\alpha$. An overview of Krippendorff $\alpha$ for all metric implementations is given in Appendix \ref{app:alpha}.

\subsection{Pairwise Comparison of Methods}
We use the proposed framework in Section \ref{sec:cles} to compare the performance of DeepSHAP and DeepLIFT on MNIST, CIFAR-10 and ImageNet.
We choose these two methods because they have very similar results across all datasets in Figures \ref{fig:wsp-low}, \ref{fig:wsp-medium} and \ref{fig:wsp-high}, which is to be expected as DeepSHAP is based on DeepLIFT. However, DeepSHAP is computationally much more expensive than DeepLIFT, so if the fraction of images where it outperforms DeepLIFT is relatively small, it might not be worth the cost. Note that this specific choice was made merely for demonstration purposes. In practice, we recommend that practitioners select methods to compare based on their results on the paired t-tests and other relevant factors such as computational complexity, difficulty of implementation, etc.

\begin{figure}
	\centering
	\begin{subfigure}[b]{0.49\textwidth}
		\includegraphics[width=\textwidth]{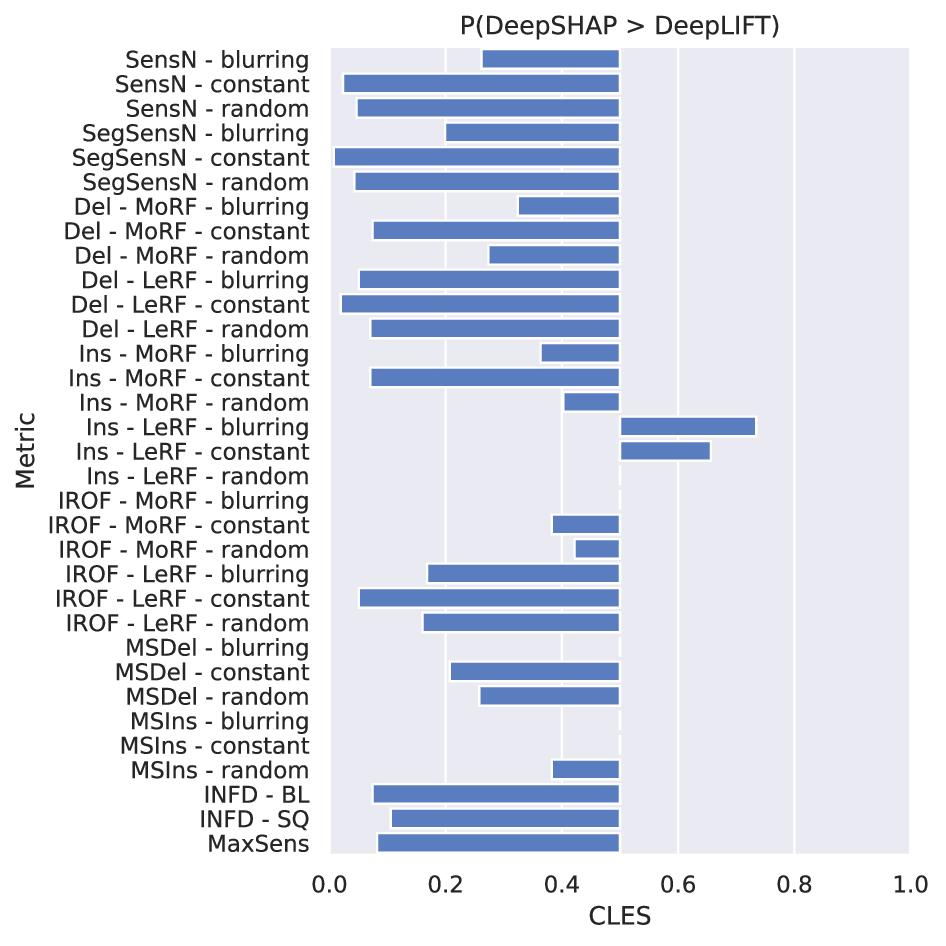}
		\caption{MNIST}
	\end{subfigure}
	\begin{subfigure}[b]{0.49\textwidth}
		\includegraphics[width=\textwidth]{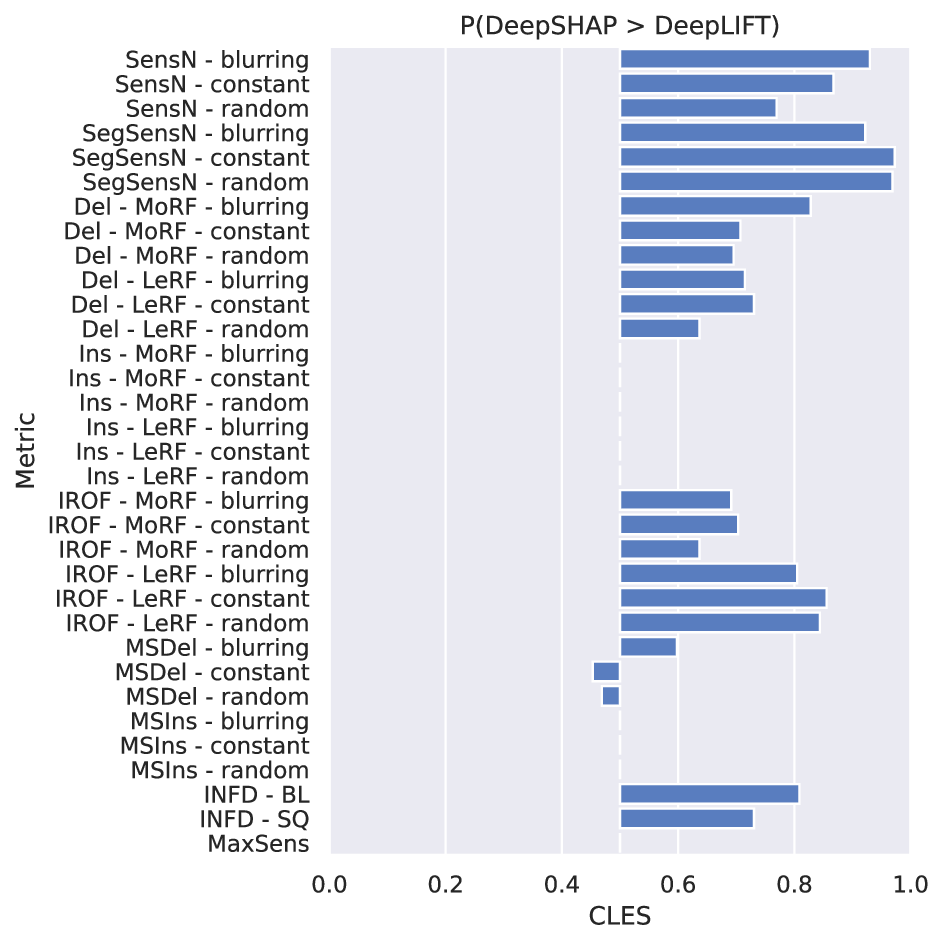}
		\caption{CIFAR-10}
	\end{subfigure}
	\begin{subfigure}[b]{0.49\textwidth}
		\includegraphics[width=\textwidth]{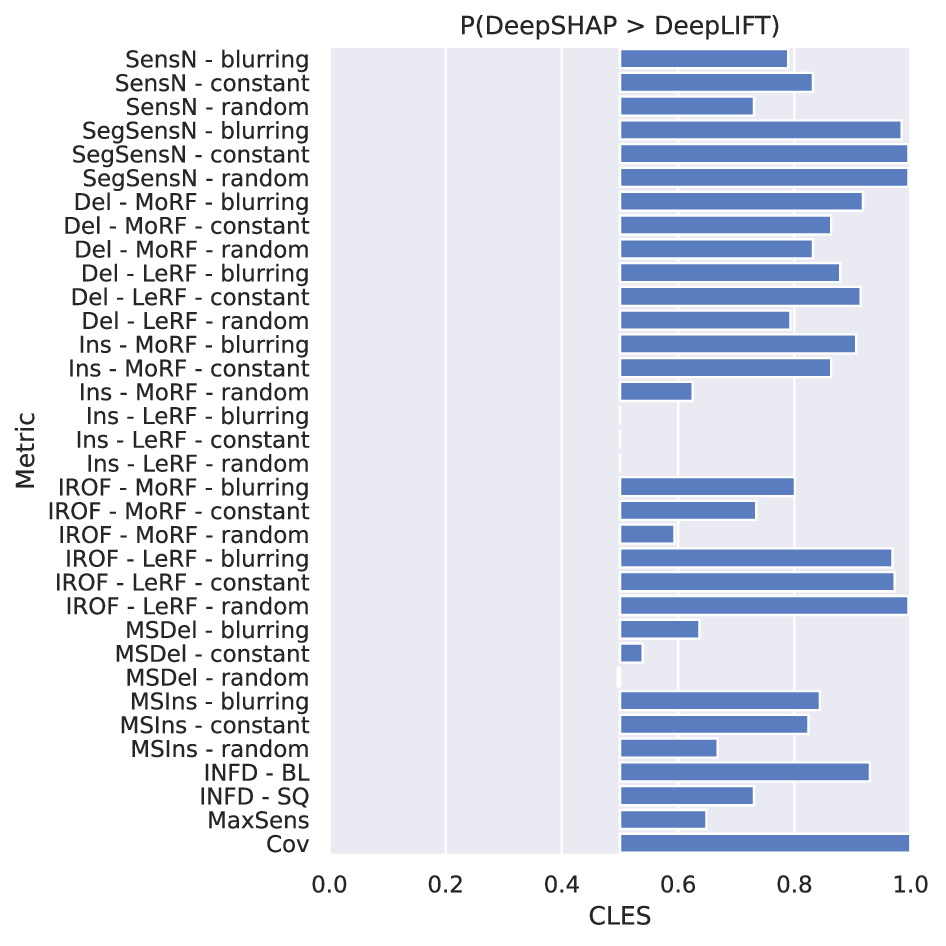}
		\caption{ImageNet}
	\end{subfigure}
	\caption{Comparison of DeepSHAP vs. DeepLIFT using Common Language Effect Size.}
	\label{fig:cles}
\end{figure}

The results are shown in Figure \ref{fig:cles}. Each bar corresponds to a paired t-test between the results for DeepSHAP and DeepLIFT on a single metric. A bar is only drawn if the corresponding result was significant ($p < 0.01$). The bars are centered on 0.5, since a Probability of Superiority of 0.5 would indicate that both methods are equivalent, each outperforming the other in 50\% of cases.

We see that, although performance in terms of absolute metric scores is very similar between the two methods (as shown in Section \ref{sec:results:ttest}), the Probability of Superiority (PoS) varies greatly depending on the dataset. On ImageNet, DeepSHAP outperforms DeepLIFT for most images, with the PoS ranging between 60-80\% for most metrics. On CIFAR-10 however, the difference between the two methods is much smaller. Finally, on MNIST, DeepSHAP is outperformed by DeepLIFT on a majority of images, for almost all metrics. This indicates that the relative performance of methods is strongly dependent of the dataset in question.

\subsection{Sensitivity-n vs. Seg-Sensitivity-n}
\begin{figure}
	\centering
	\includegraphics[width=\textwidth]{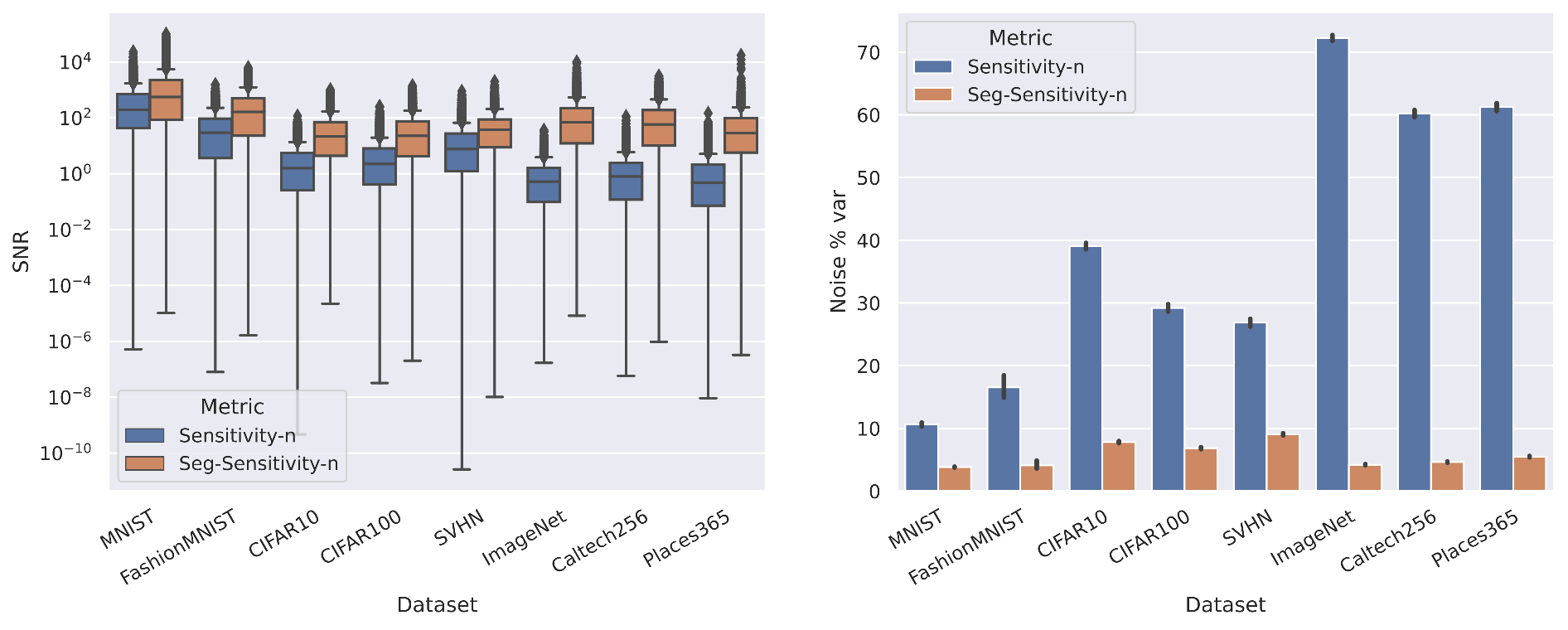}
	\caption{Comparison between Sensitivity-$n$ and Seg-Sensitivity-$n$. Left: Signal-to-noise ratio (logarithmic scale). Right: noise fraction of variance.}
	\label{fig:sens_n}
\end{figure}

To compare our proposed metric Seg-Sensitivity-$n$ to the original Sensitivity-$n$, we measure the stability of both metrics in two ways. First, we measure the signal-to-noise ratio (SNR) of both metrics. We repeatedly compute both Seg-Sensitivity-$n$ and Sensitivity-$n$ scores 100 times on 256 images (where the same images were used for both metrics). We then compute the SNR ratio of the metric for each image as $\frac{\mu^2}{\sigma^2}$, where $\mu$ is the mean of the 100 metric values, and $\sigma$ is the standard deviation. The results are shown in the left part of Figure \ref{fig:sens_n}. Note that the SNR of Seg-Sensitivity-$n$ for high-dimensional datasets (ImageNet, Caltech-256 and Places-365) is significantly higher than the SNR of Sensitivity-$n$. On the other datasets, the SNR is also larger for Seg-Sensitivity-$n$, although the difference is smaller.

A different way to measure the stability is to look at the \textit{noise fraction of variance}. To compute this, we compute the ratio of the within-sample variance (the variance of the 100 repeated measurements for each sample) to the between-sample variance (the total variance of all measurements on all samples). A low noise fraction of variance corresponds to a clear signal. These results are shown in the right plot of Figure \ref{fig:sens_n}. We see again that the noise fraction of variance for Sensitivity-$n$ is much larger than for Seg-Sensitivity-$n$, especially on the high-dimensional datasets.

\subsection{Parameter Randomization}
\begin{figure}
	\centering
	\includegraphics[width=\textwidth]{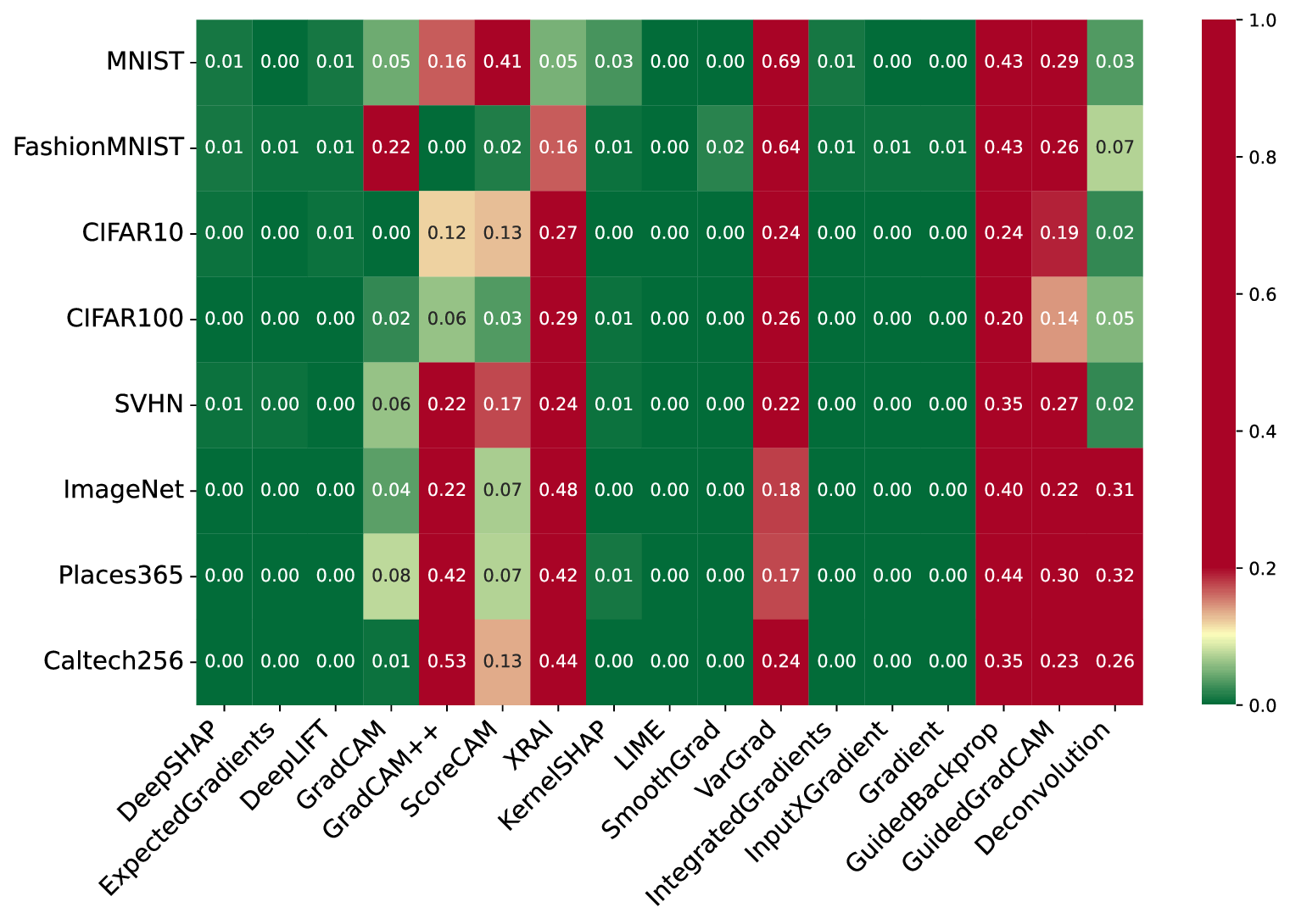}
	\caption{Results of the Parameter Randomization metric.}
	\label{fig:parameter-randomization-imaging}
\end{figure}

Results of the Parameter Randomization test are shown in Figure \ref{fig:parameter-randomization-imaging}. We consider any method that has an absolute rank correlation larger than 0.2 as failing the test. Note that these rank correlations are between the attribution maps produced by the same method before and after randomization of the model parameters. They should not be confused with the inter-metric correlations discussed in Section \ref{sec:results-inter-metric-correlation}.

We notice that the methods that were identified in \citep{Adebayo2018a} as failing the test (Guided Backpropagation and Guided Grad-CAM) also have relatively large correlation values across all datasets, implying that they indeed fail the sanity checks on all these datasets. Additionally, methods that were shown by \citet{Adebayo2018a} to pass the test (Gradient, Integrated Gradients and InputXGradient) obtain low correlation scores across all datasets as well. We can therefore confirm many of the experimental findings of \citet{Adebayo2018a} across a wider selection of datasets. 

Interestingly, ScoreCAM and XRAI have large correlation values across many datasets, even though they were described as passing the Parameter Randomization test by their original authors \citep{Wang2020ScoreCAM,Kapishnikov_2019_ICCV}. However, ScoreCAM was only inspected visually, and the test for XRAI was only conducted on MNIST. Our findings confirm that XRAI passes the test on MNIST, but show that the same method also fails on other datasets. We therefore conclude that the outcome of the Parameter Randomization test is indeed dependent on the dataset and/or model, which is in concordance with the findings of \citet{yona2021}. Based on these results, our recommendation to practitioners is to perform a quantitative analysis of the Parameter Randomization test on the specific dataset in question, rather than performing a simple visual inspection or assuming that a method will pass the test if it was shown to pass the test on a different dataset (such as MNIST).

\section{Conclusion and future work}
\label{sec:guidelines}
We have performed an extensive study of the behaviour of a large number of attribution metrics and methods, on a collection of image datasets with varying complexity and dimensionality. From this investigation, we draw the following general conclusions:

\begin{itemize}
	\item Metric scores vary strongly for different datasets. This implies that the performance of attribution methods should be measured for each specific use case, rather than drawing general conclusions from the results on a set of benchmark datasets.
	\item Most metrics tend to have low ranking consistency, shown by the relatively low values of Krippendorff $\alpha$. From this we conclude that a rigorous statistical testing approach is necessary to draw any dataset-wide conclusions. Also, the ranking consistency values of metrics themselves are not consistent across datasets, implying that there is no generally superior evaluation metric in terms of ranking consistency.
	\item We confirm the conclusion from \citet{Tomsett} that metrics do not necessarily measure the same underlying concept to a larger amount of metrics, and extend their findings to include Sensitivity-n \citep{Ancona2017}, Infidelity \citep{Yeh2019}, IROF \citep{Rieger} and Impact Coverage \citep{QiuLin}. This can be seen in the low inter-metric correlation values between these metrics.
	\item The result of the Parameter Randomization test, introduced as a sanity check in \citet{Adebayo2018a}, is dataset-dependent, meaning that whether a method passes or fails the sanity check depends on the dataset and/or the model that is being used. This implies that the Parameter Randomization test should also be performed for each specific use case. This experimentally confirms the hypothesis posed by \citet{yona2021} on natural image datasets.
	\item From the complementarity of results between coarse-grained and fine-grained attribution maps, we conclude that these methods might have to be evaluated in fundamentally different ways, focusing on single-pixel importance for fine-grained maps and on a more high-level view for coarse-grained maps. Further research is needed to verify or refute this hypothesis.
	\item Finally, we also introduce Seg-Sensitivity-n as an extension of Sensitivity-n \citep{Ancona2017}, and show that it has a higher signal-to-noise ratio than Sensitivity-n on high-dimensional datasets.
\end{itemize}

\begin{figure}
	\centering
	\includegraphics[width=\textwidth]{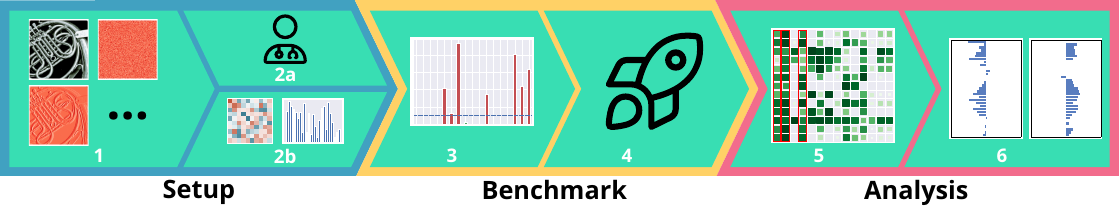}
	\caption{Visual overview of the proposed benchmarking guidelines. (1) Choose a baseline. This can be a random baseline, an edge detector, or some other baseline of choice. (2) Select metrics. This can be done manually (2a) or through a pilot study (2b) where metrics are selected based on inter-metric correlations and Krippendorff $\alpha$. (3) Compute the parameter randomization sanity check and discard methods that fail it. (4) Run the remaining methods on the selected metrics. (5) Perform statistical tests against the random baseline and make a selection of well-performing methods. (6) Perform pairwise comparisons between the selected methods.}
	\label{fig:infographic}
\end{figure}

From these conclusions, we propose a set of benchmarking guidelines for practitioners seeking to select the best feature attribution method for their specific use case (see Figure \ref{fig:infographic}). We note that these guidelines should be viewed as exploratory, as more research is needed into which specific aspects of explanation methods are evaluated by the different metrics. This means that it is still difficult to prove that one method is strictly ``better'' than another, especially if metrics contradict each other. Our recommendation to practitioners is to use these benchmarking guidelines to perform a first selection of candidate methods, and then select one or multiple explanation methods based on use case-specific properties, such as computational budget, access to model internals, fine- or coarse-grainedness, and/or others.

\begin{enumerate}
	\item \textbf{Baseline selection:} First, a baseline attribution method must be defined. In general, a uniform random baseline can be used, but more specific baselines can also be chosen depending on the use case, such as an edge detector or some specific explanation method that one hopes to outperform.
	\item \textbf{Metric selection:} Next, a selection of metric implementations must be made. This can be done manually, if such a selection of metrics is obvious from the use case and there is a clear approach to masking available (2a in Figure \ref{fig:infographic}). In MNIST for example, masking using the black background color might be an intuitive choice. Alternatively, a \textit{pilot study} can be performed (2b in Figure \ref{fig:infographic}). In such a pilot study, a large number of metrics and masking approaches are tested on a limited number of images. We then recommend computing inter-metric correlations and Krippendorff $\alpha$ values, and selecting those metric implementations that have high Krippendorff $\alpha$ and low inter-metric correlations. In this way, a minimal number of images can be used to draw dataset-wide conclusions, and metric scores will contain a minimal amount of redundant information.
	\item \textbf{Parameter Randomization test:} Before running a full benchmark on all available methods, we recommend first performing the Parameter Randomization test from \citet{Adebayo2018a} using the spearman rank correlation. One should avoid using absolute values of attribution scores when performing this test, as this has been shown to unfairly penalize certain methods \citep{binder2023}. Any methods that fail the test (by obtaining a correlation score larger than a chosen threshold value) should be discarded from further analysis.
	\item \textbf{Full benchmark:} Once a selection of metrics is made and methods that fail the Parameter Randomization test are discarded, metric scores can be computed for the remaining methods on a large enough number of samples.
	\item \textbf{Rough statistical analysis:} Once the metric scores are computed for all methods and the baseline, a rough overview of method performance can be made using a paired t-test for each metric between each method and the random baseline. For those methods that significantly outperform the baseline on a given metric, we recommend using the Cohen's $d$ effect size to quantify performance.
	\item \textbf{Detailed comparison:} Using the rough overview made in the previous step, we recommend selecting a smaller number of well-performing attribution methods, if possible with varying computational complexity. Those methods can then be compared in more detail using new paired t-tests, this time between the two methods rather than between a method and the baseline. We then recommend using the Probability of Superiority as an interpretable effect size measure to assess the fraction of cases where one method is superior to another. Based on these results, as well as other use-case specific constraints, a final selection of one or more ideal methods can then be made.
\end{enumerate}
An example application of these guidelines can be found in Appendix \ref{app:guidelines-imagenet}.

The results described in this paper leave a number of directions of future research. First of all, the observation that metrics do not necessarily measure the same underlying concept of feature attribution maps leads to the question of what those underlying concepts might be. A better understanding of those underlying concepts can lead to more directed benchmarking efforts and the development of better methods and/or metrics. A possible link can be made with the concepts of necessity and sufficiency, found in the literature of causality \citep{pearl2009causality}. Secondly, the complementarity of results for coarse- and fine-grained methods implies that a combination of different attribution maps can be more informative than a single one. This can lead to the development of new methods, generalizing the concept of feature attribution itself. Future work can also be done in designing specific evaluation metrics for coarse- or fine-grained attribution methods. Finally, the application of the benchmark procedure on new datasets can shed light on what the best attribution methods are for a given problem domain. An important example is the domain of biomedical imaging. Here, medical practitioners are often interested in what the most important regions of a radiographic image are for a specific prediction \citep{vandervelden2022}, in order to build trust in the model and identify when a model might be making a mistake. Application of the general guidelines given above can help developers choose the right attribution method in this case.

\begin{appendices}
	\section{Supplementary figures}
	\label{app:alpha}
\begin{figure}
		\centering
		\begin{subfigure}[b]{0.49\textwidth}
			\includegraphics[width=\textwidth]{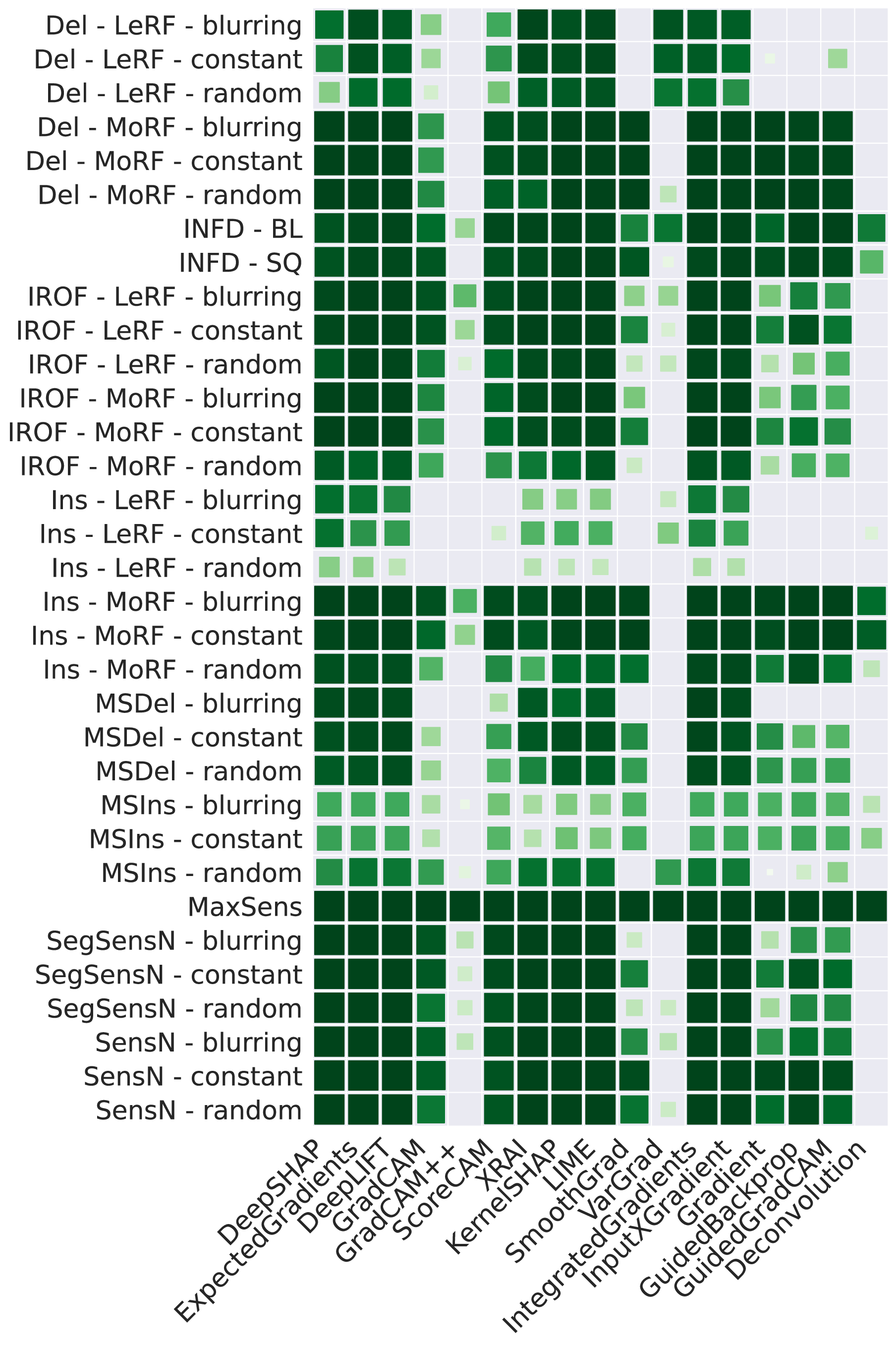}
			\caption{MNIST}
		\end{subfigure}
		\begin{subfigure}[b]{0.49\textwidth}
			\includegraphics[width=\textwidth]{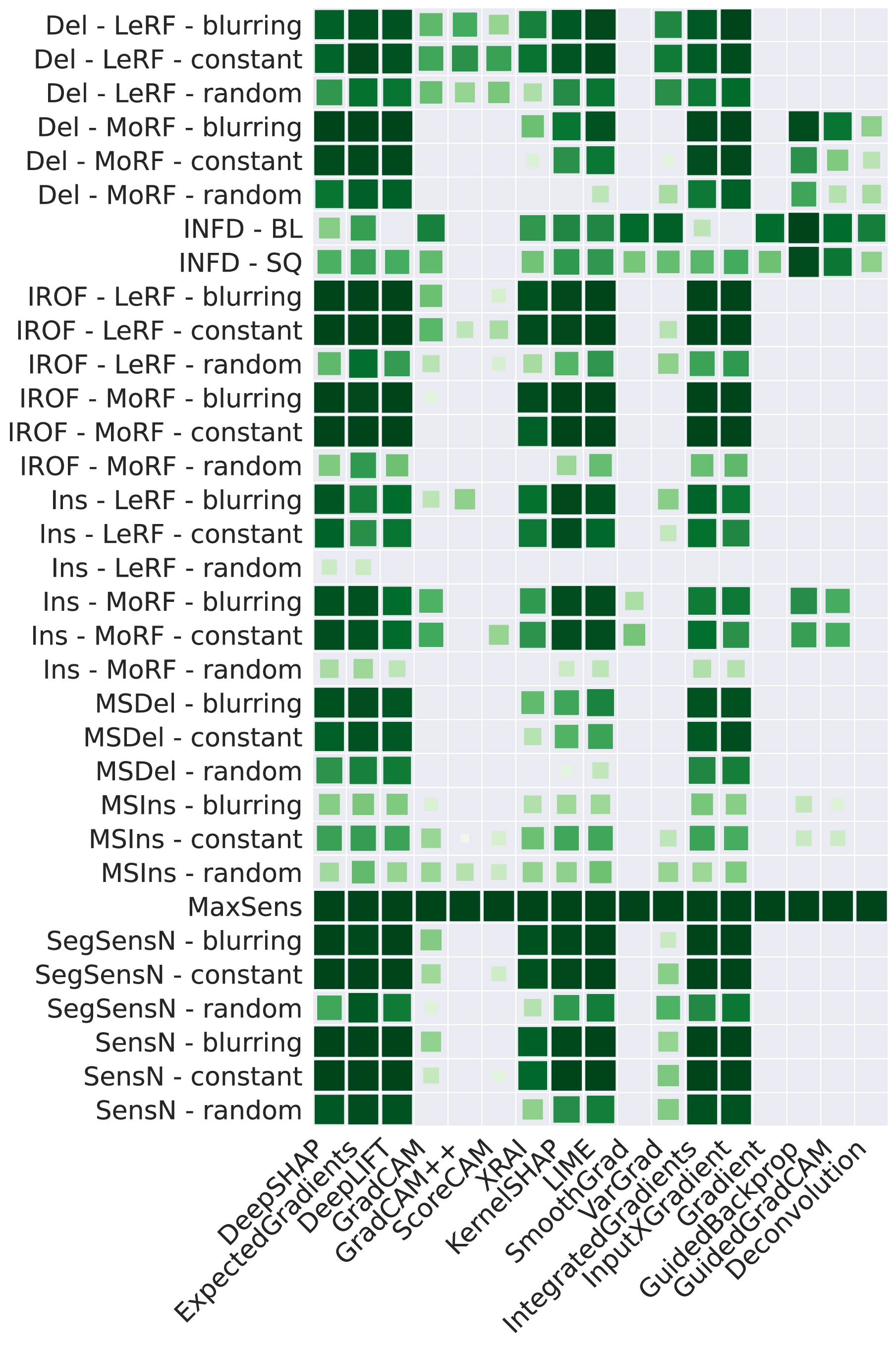}
			\caption{FashionMNIST}
		\end{subfigure}
		\caption{Results of paired t-tests for all metrics on low-dimensional datasets.}
		\label{fig:results:ttest-all-lowdim}
	\end{figure}

\begin{figure}
		\begin{subfigure}[b]{0.49\textwidth}
			\includegraphics[width=\textwidth]{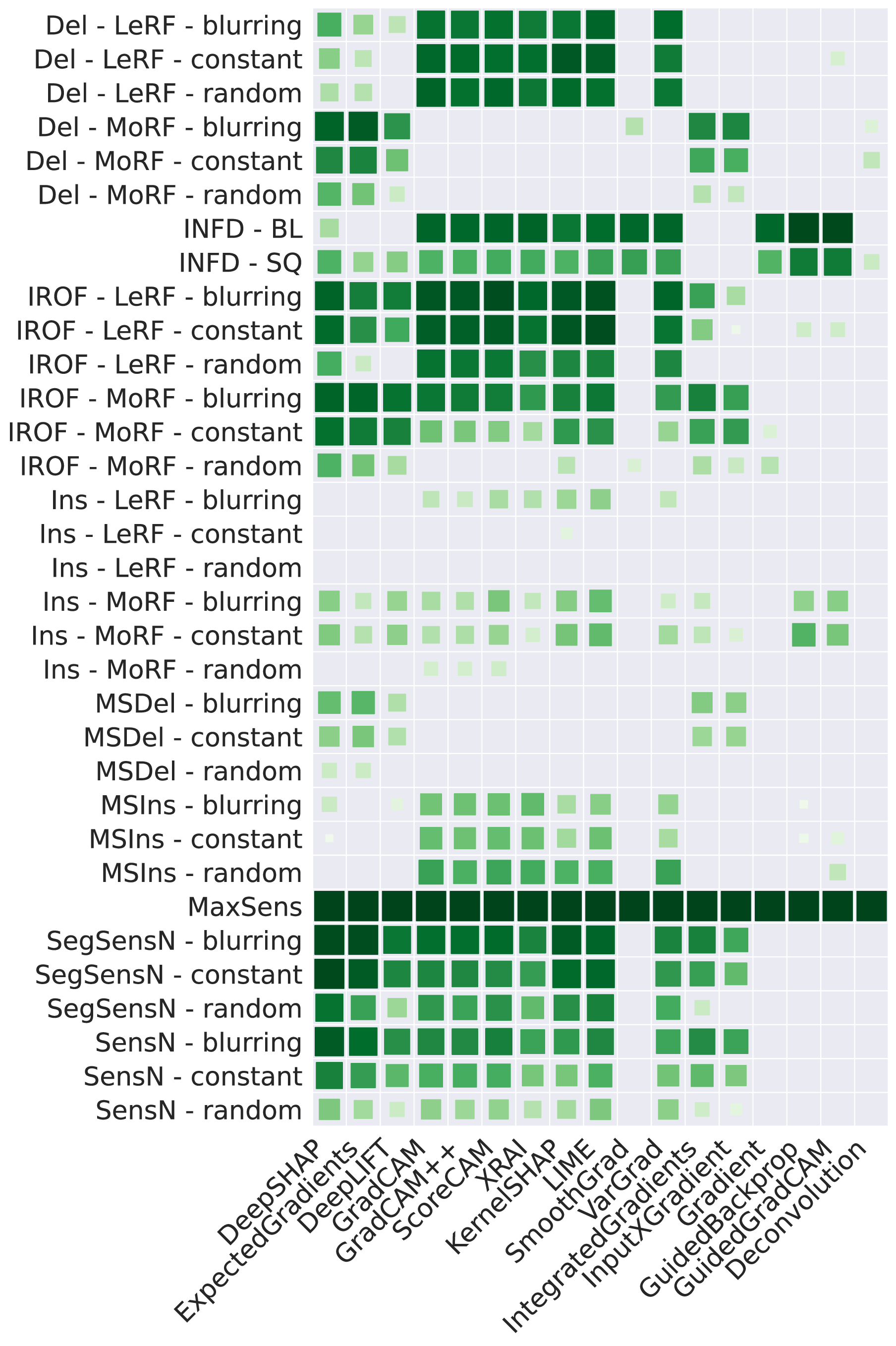}
			\caption{CIFAR-10}
		\end{subfigure}
		\begin{subfigure}[b]{0.49\textwidth}
			\includegraphics[width=\textwidth]{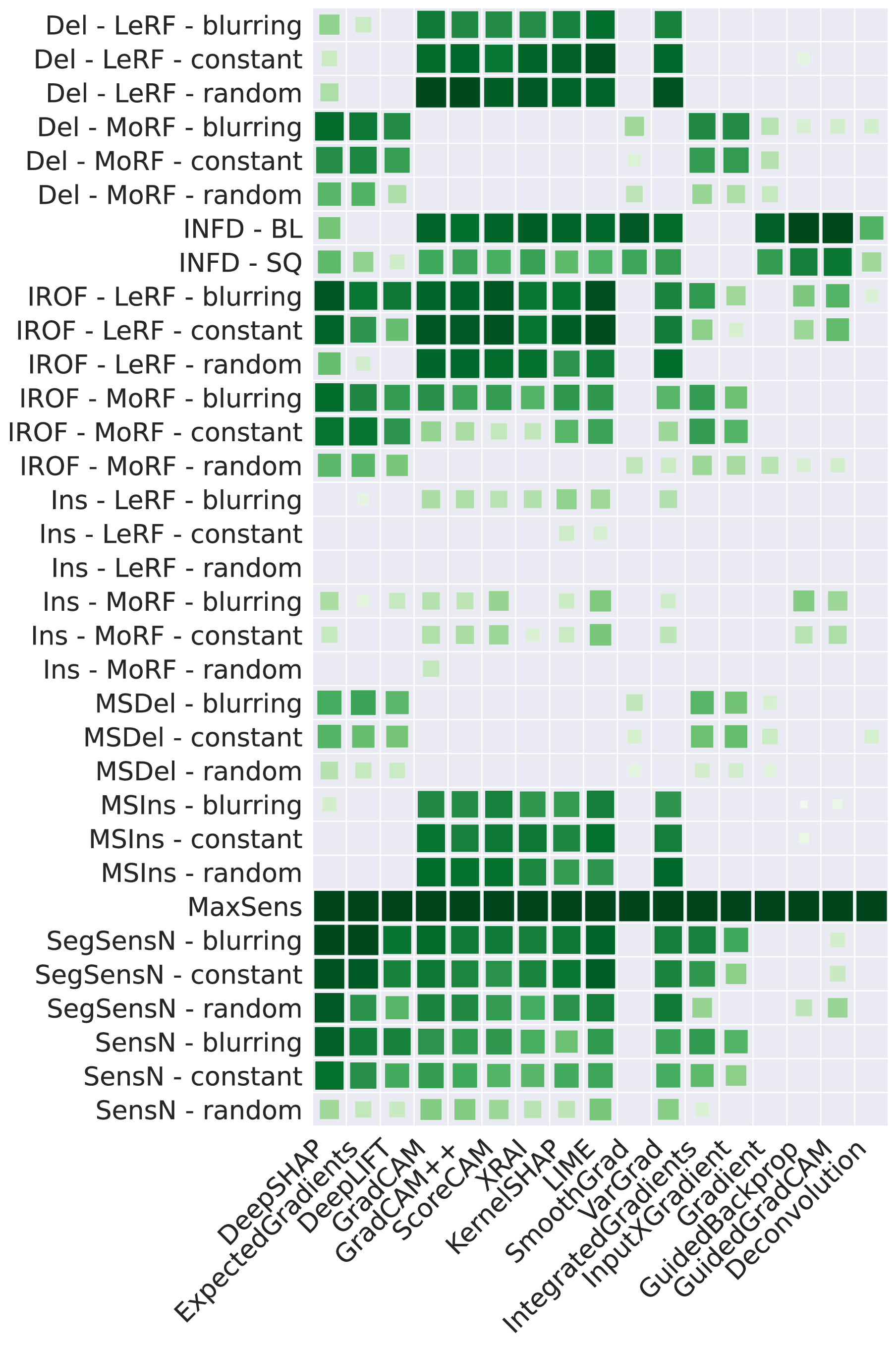}
			\caption{CIFAR-100}
		\end{subfigure}
		\begin{subfigure}[b]{0.49\textwidth}
			\includegraphics[width=\textwidth]{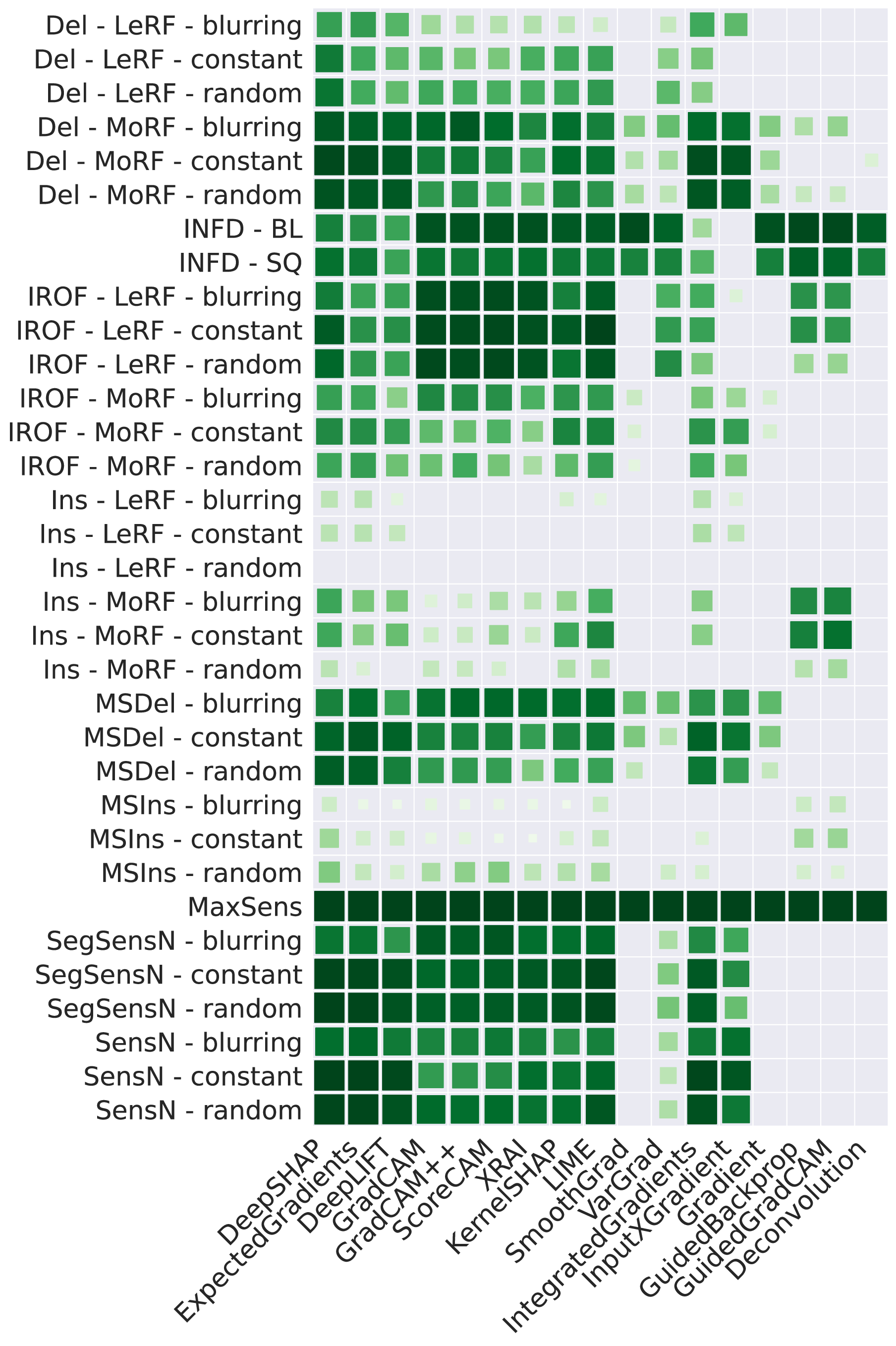}
			\caption{SVHN}
		\end{subfigure}
		\caption{Results of paired t-tests for all metrics on medium-dimensional datasets.}
		\label{fig:results:ttest-all-meddim}
	\end{figure}

\begin{figure}
		\begin{subfigure}[b]{0.49\textwidth}
			\includegraphics[width=\textwidth]{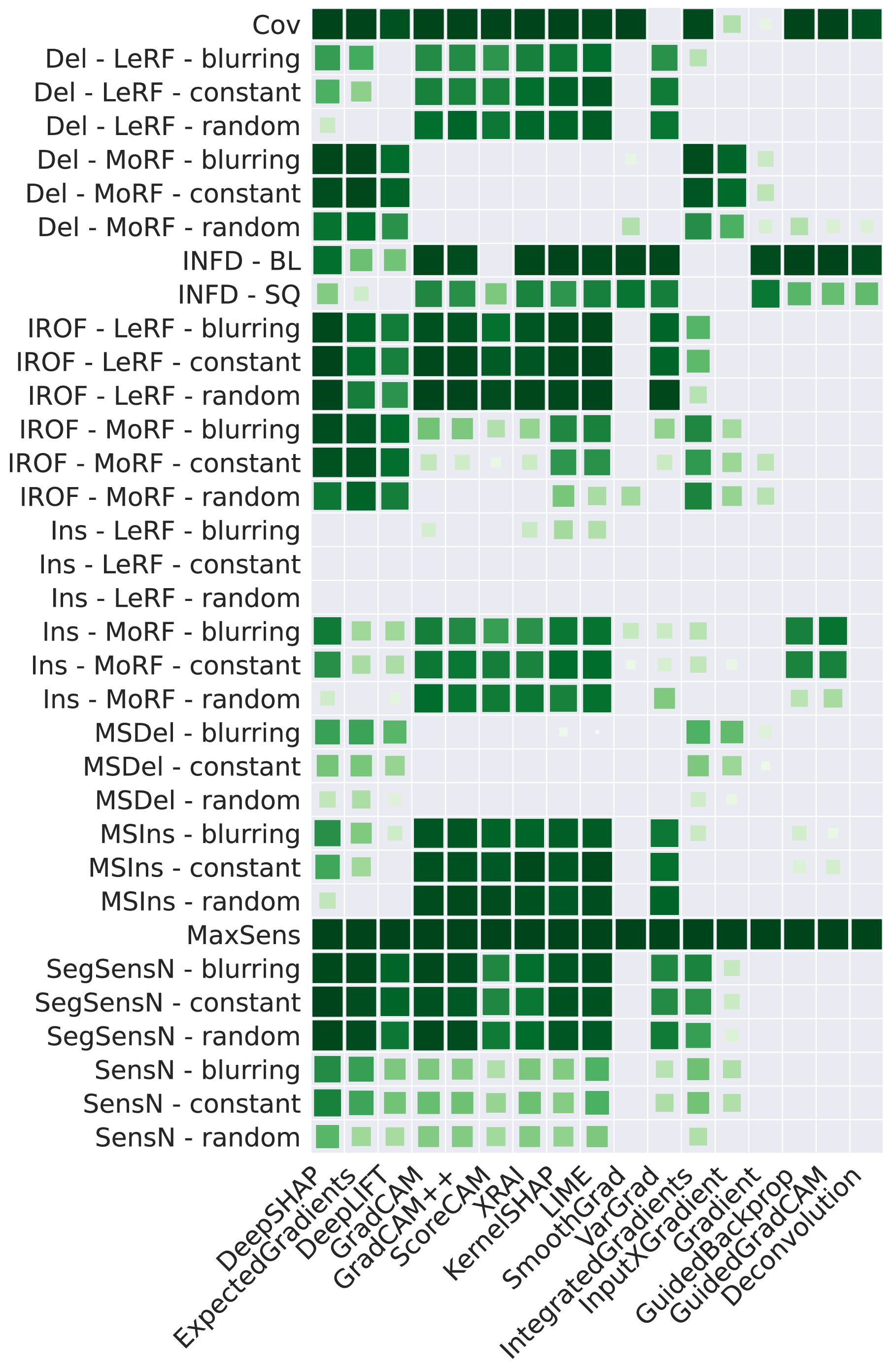}
			\caption{ImageNet}
		\end{subfigure}
		\begin{subfigure}[b]{0.49\textwidth}
			\includegraphics[width=\textwidth]{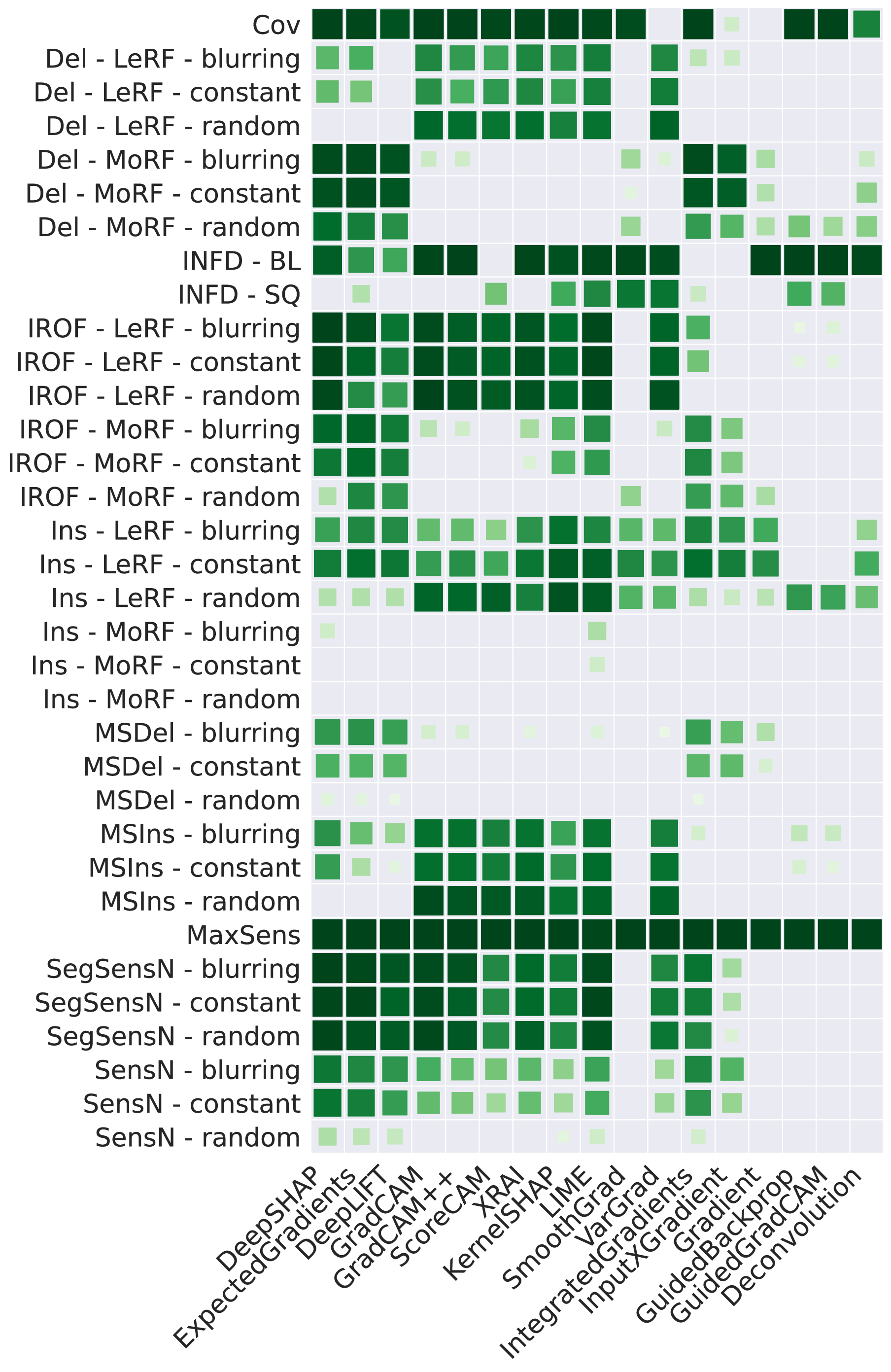}
			\caption{Caltech-256}
		\end{subfigure}
		\begin{subfigure}[b]{0.49\textwidth}
			\includegraphics[width=\textwidth]{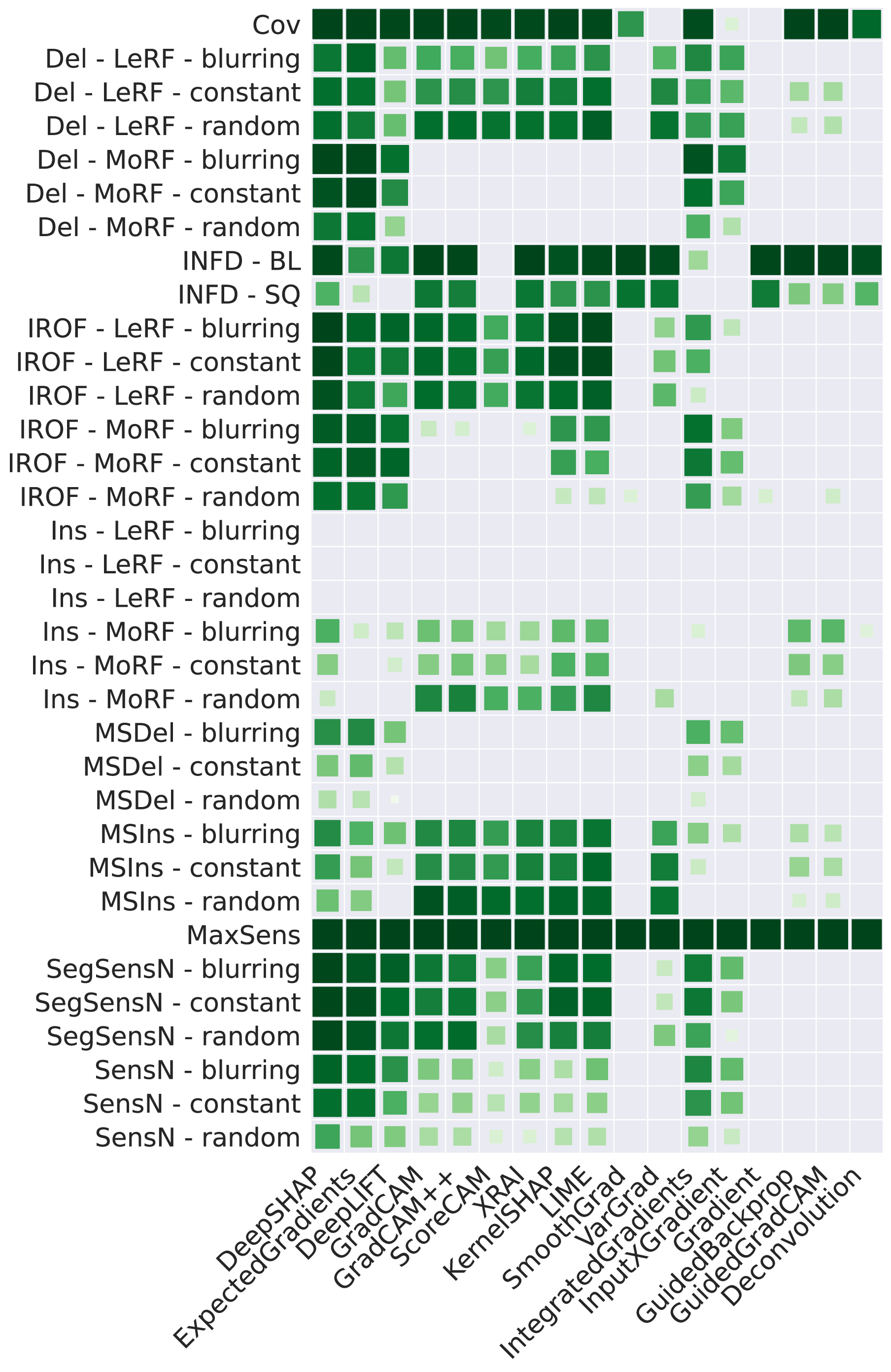}
			\caption{Places-365}
		\end{subfigure}
		\caption{Results of paired t-tests for all metrics on high-dimensional datasets.}
		\label{fig:results:ttest-all-highdim}
	\end{figure}

\begin{figure}
		\centering
		\includegraphics[height=0.7\textheight]{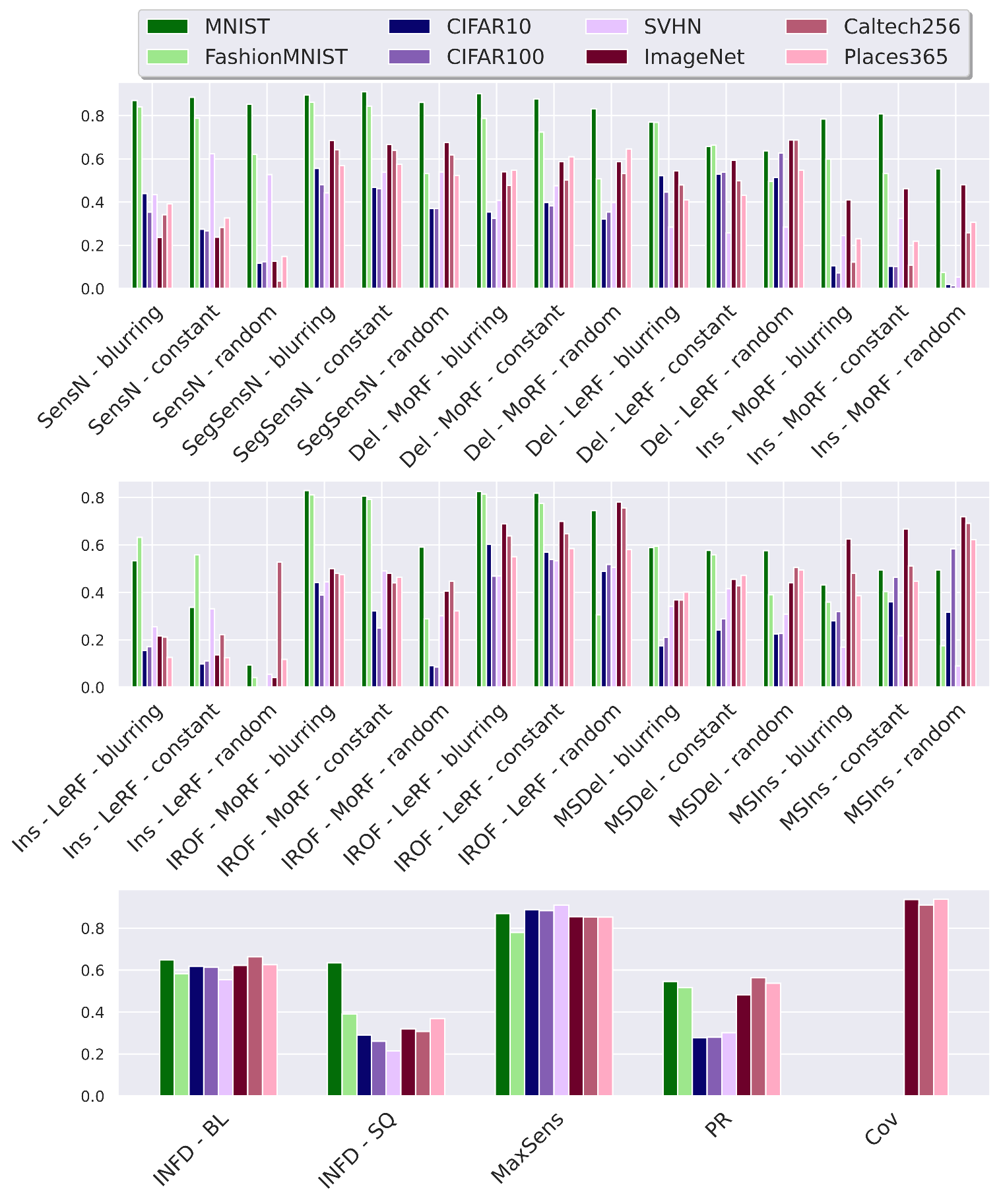}
		\caption{Krippendorff's $\alpha$ for all implementations of all metrics on all datasets. Low-, medium- and high-dimensional datasets are indicated in green, blue and red tones, respectively.}
		\label{fig:alpha-bar-all}
	\end{figure}

\begin{figure}
		\centering
		\begin{subfigure}[b]{0.32\textwidth}
			\includegraphics[width=\textwidth]{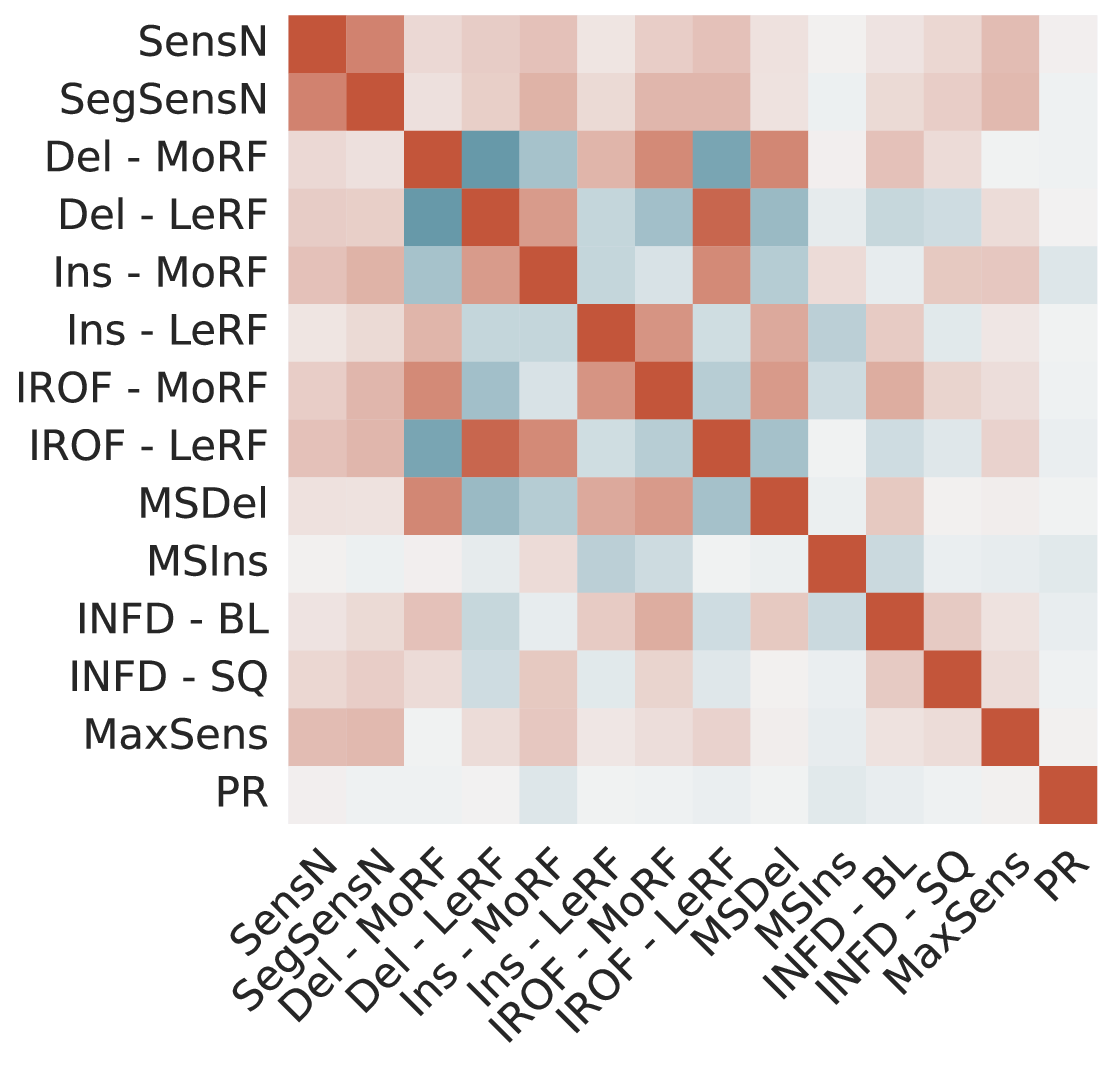}
			\caption{MNIST}
		\end{subfigure}
		\begin{subfigure}[b]{0.32\textwidth}
			\includegraphics[width=\textwidth]{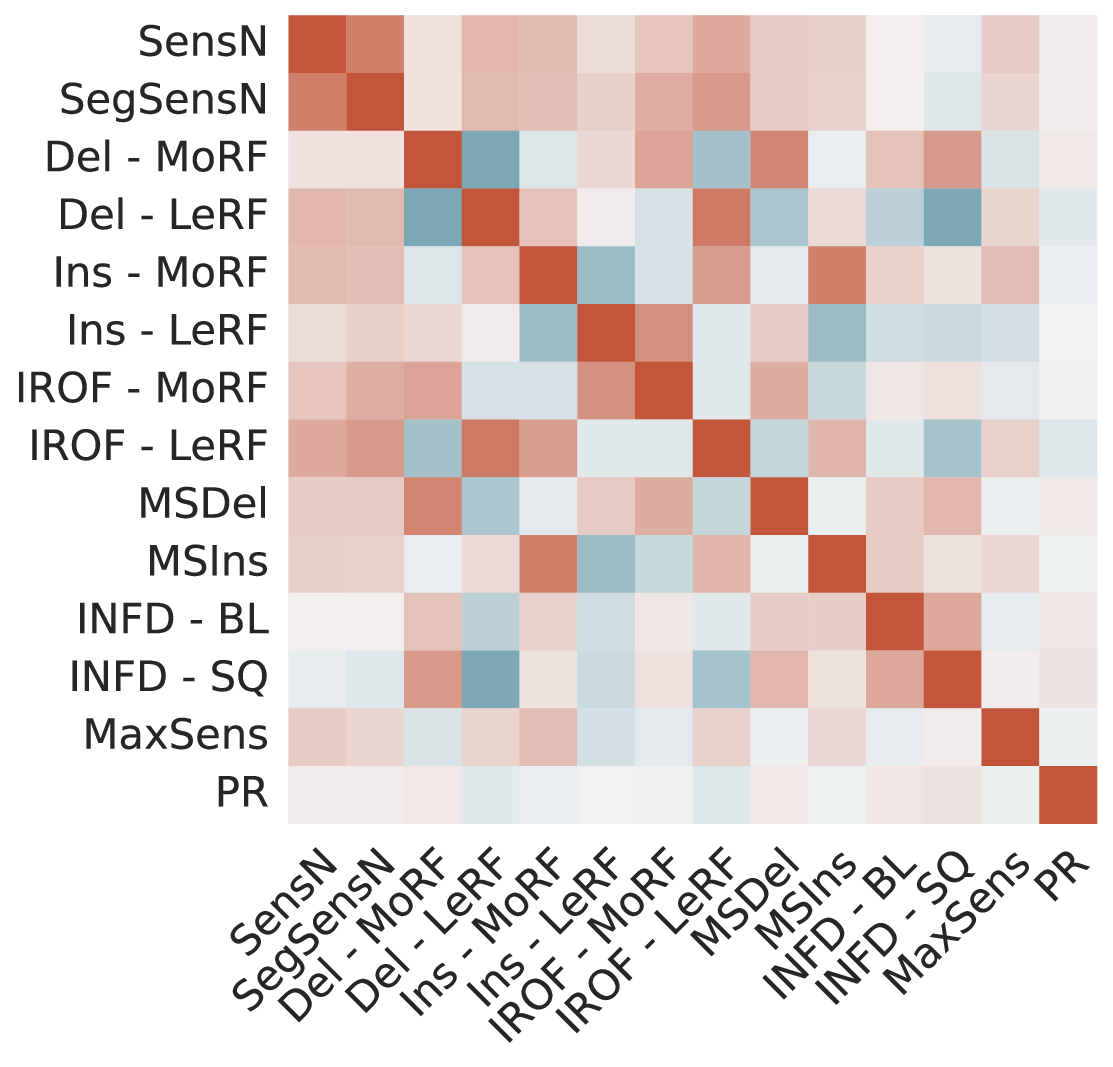}
			\caption{FashionMNIST}
		\end{subfigure}
		\begin{subfigure}[b]{0.32\textwidth}
			\includegraphics[width=\textwidth]{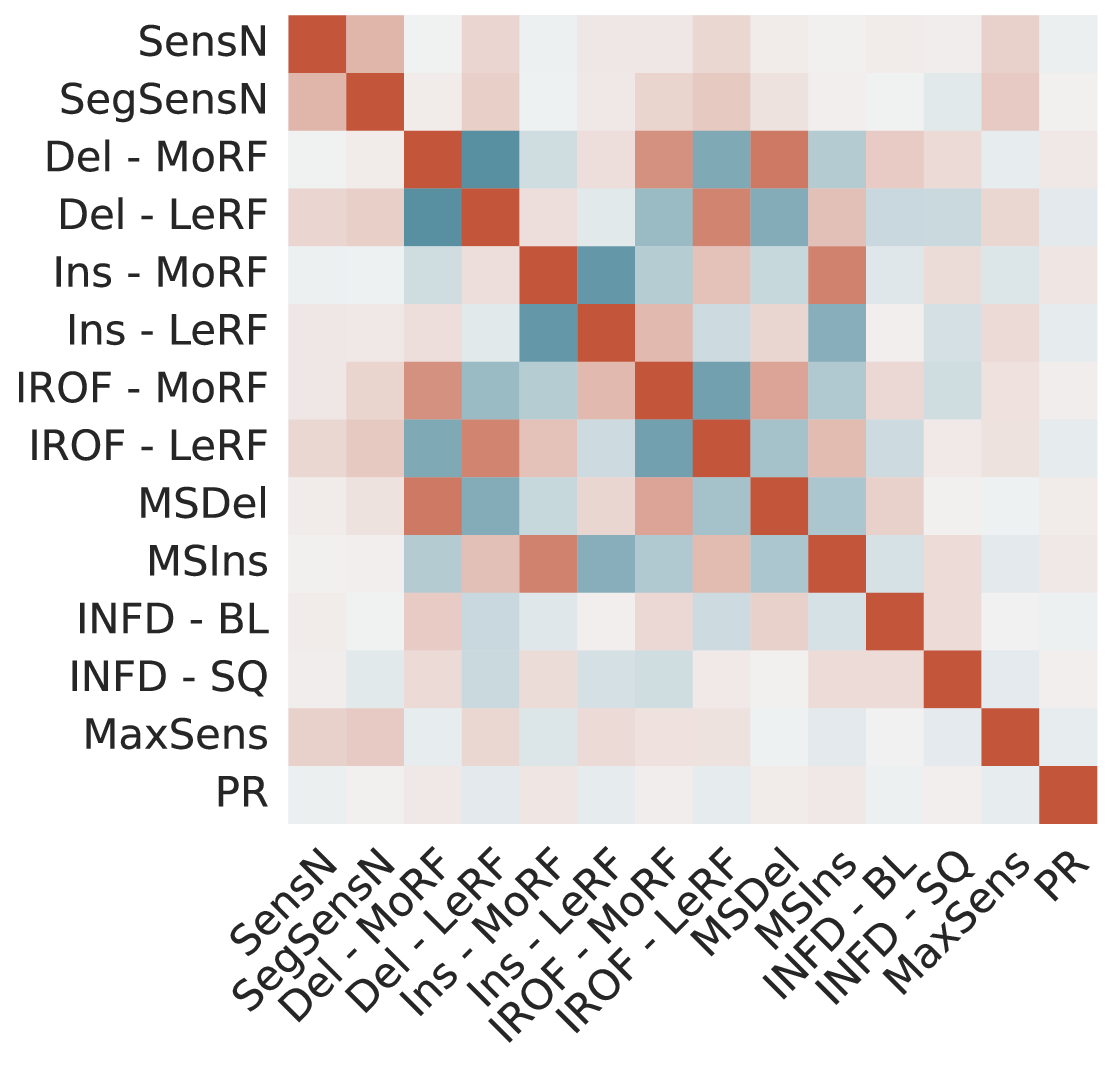}
			\caption{CIFAR-10}
		\end{subfigure}
		\begin{subfigure}[b]{0.32\textwidth}
			\includegraphics[width=\textwidth]{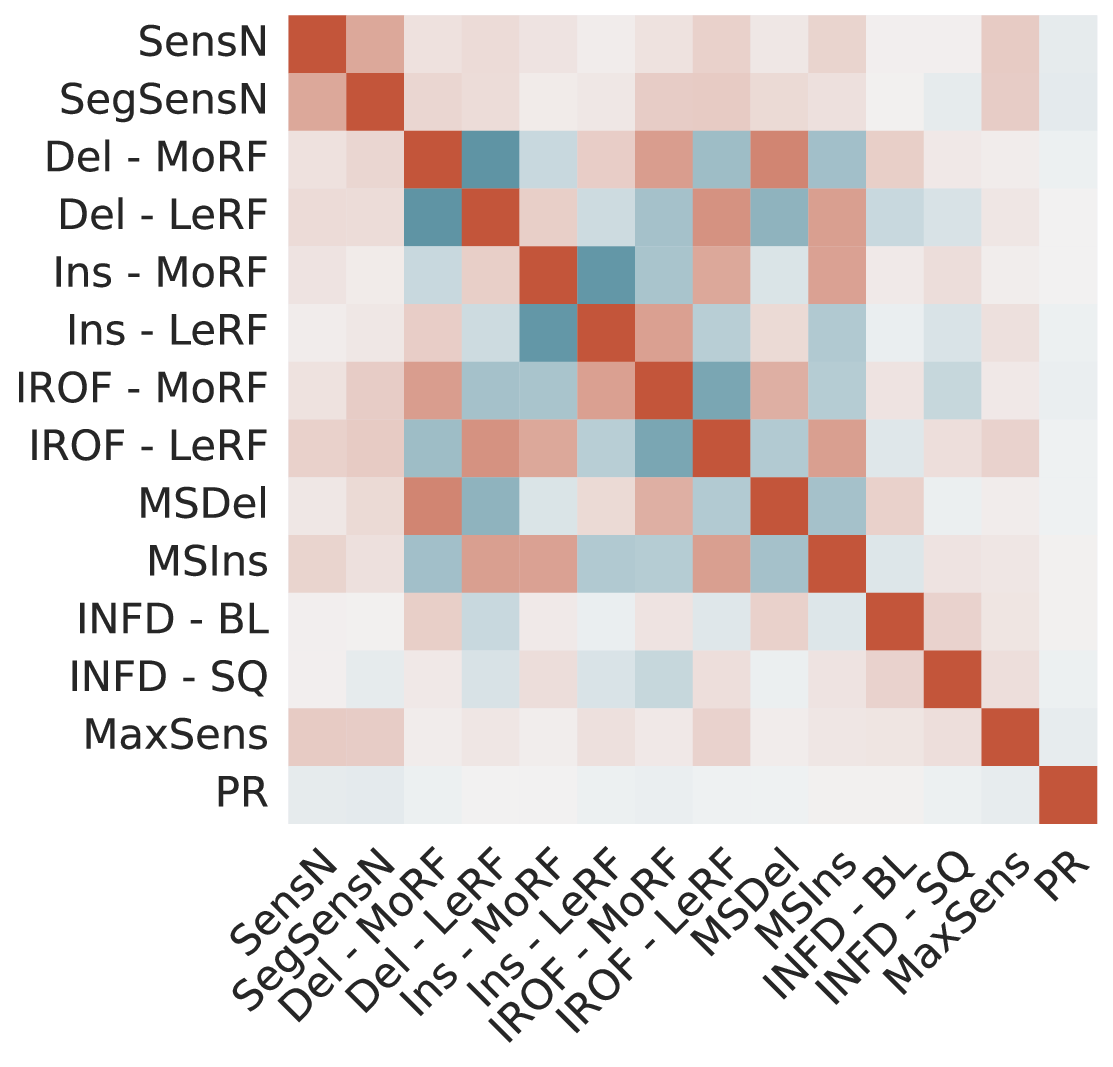}
			\caption{CIFAR-100}
		\end{subfigure}
		\begin{subfigure}[b]{0.32\textwidth}
			\includegraphics[width=\textwidth]{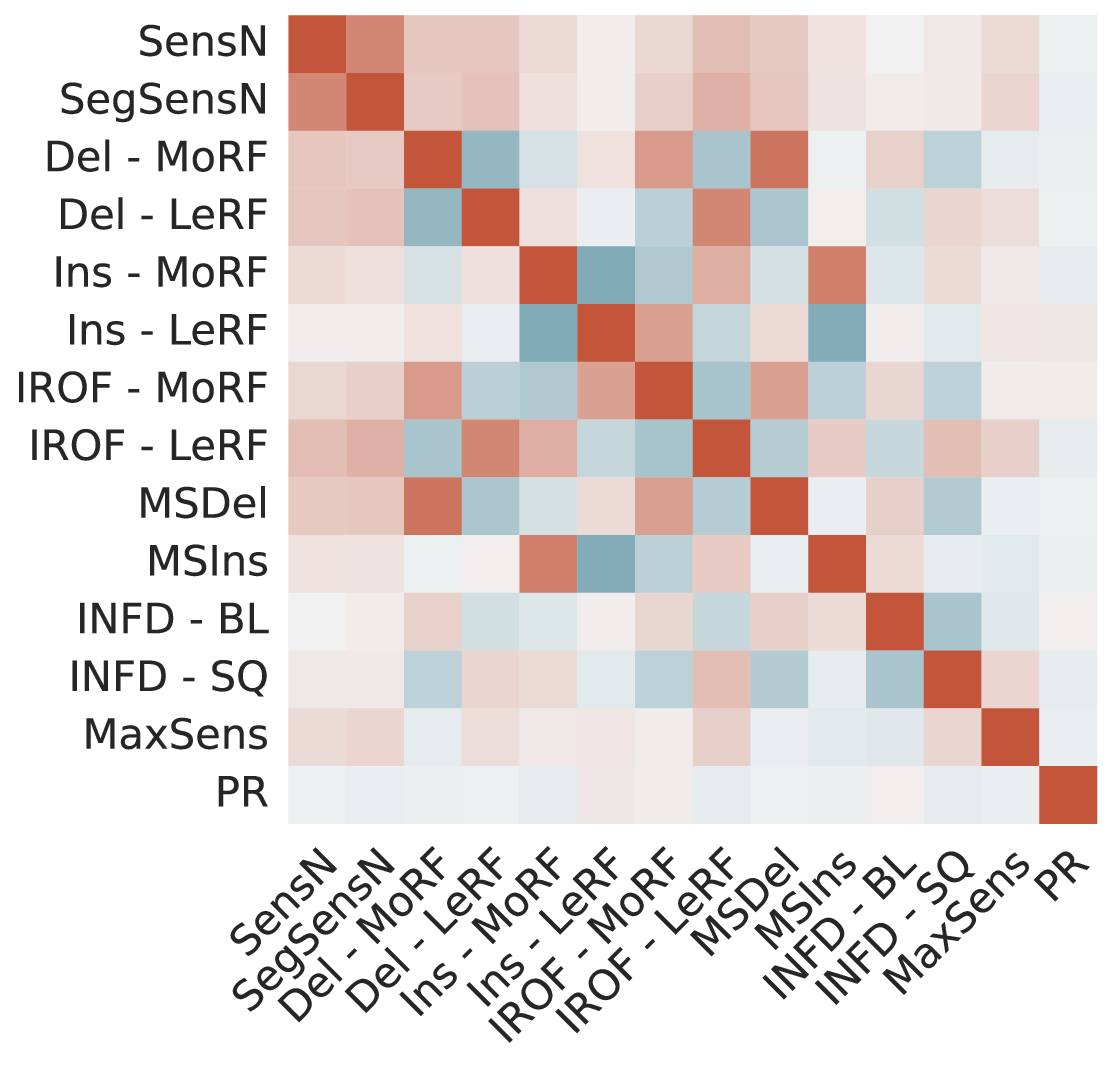}
			\caption{SVHN}
		\end{subfigure}
		\begin{subfigure}[b]{0.32\textwidth}
			\includegraphics[width=\textwidth]{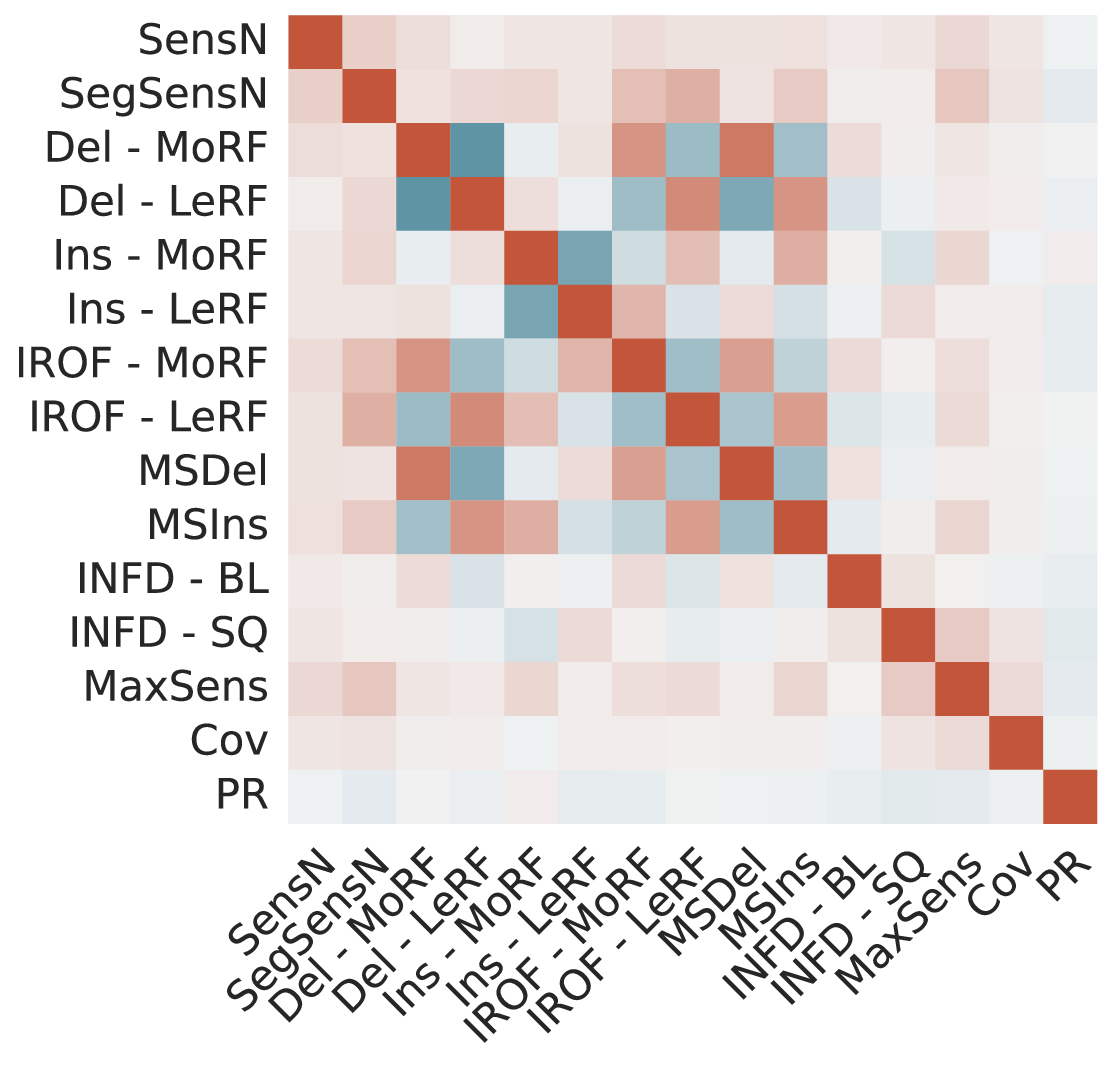}
			\caption{ImageNet}
		\end{subfigure}
		\begin{subfigure}[b]{0.32\textwidth}
			\includegraphics[width=\textwidth]{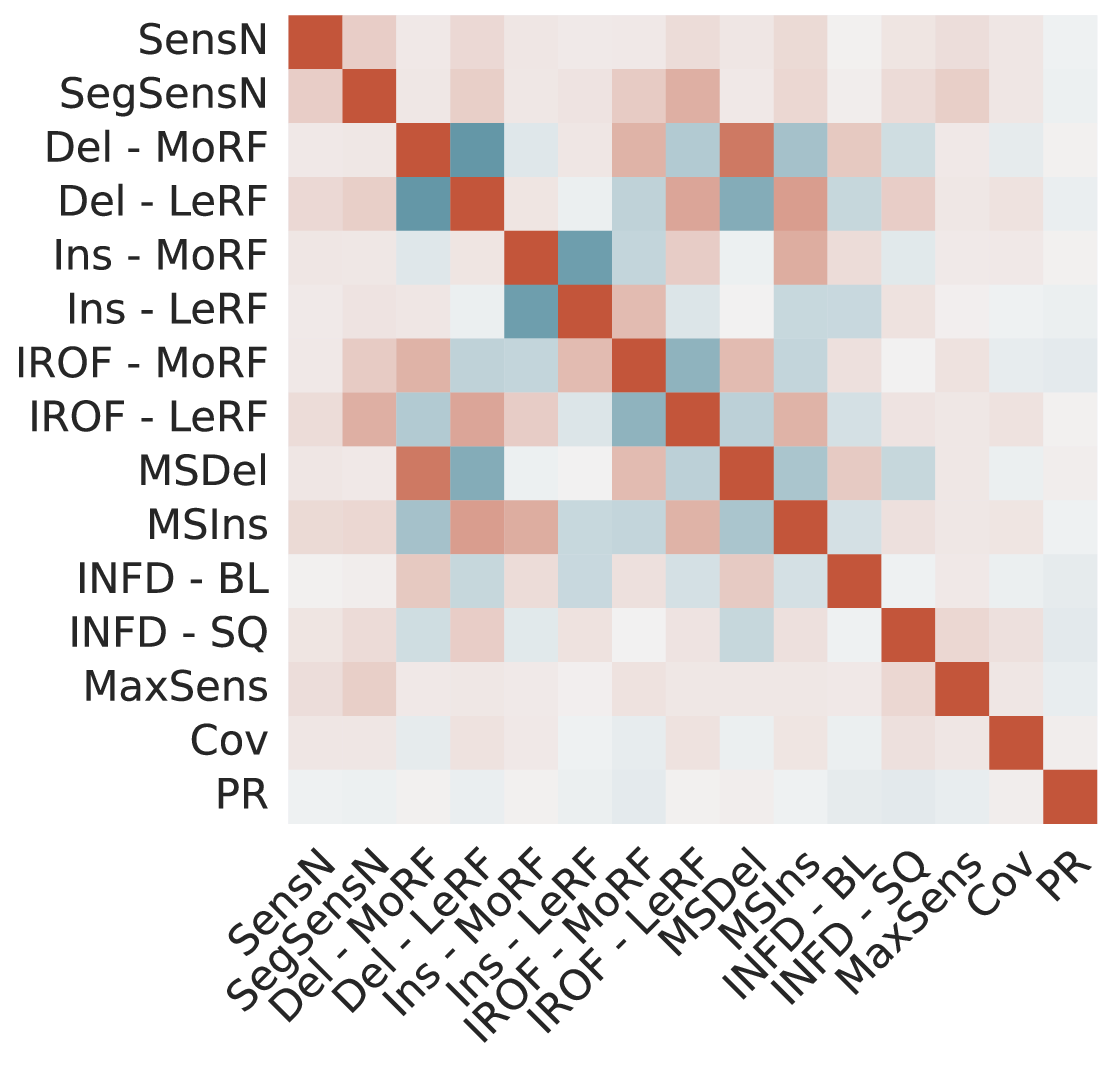}
			\caption{Caltech256}
		\end{subfigure}
		\begin{subfigure}[b]{0.32\textwidth}
			\includegraphics[width=\textwidth]{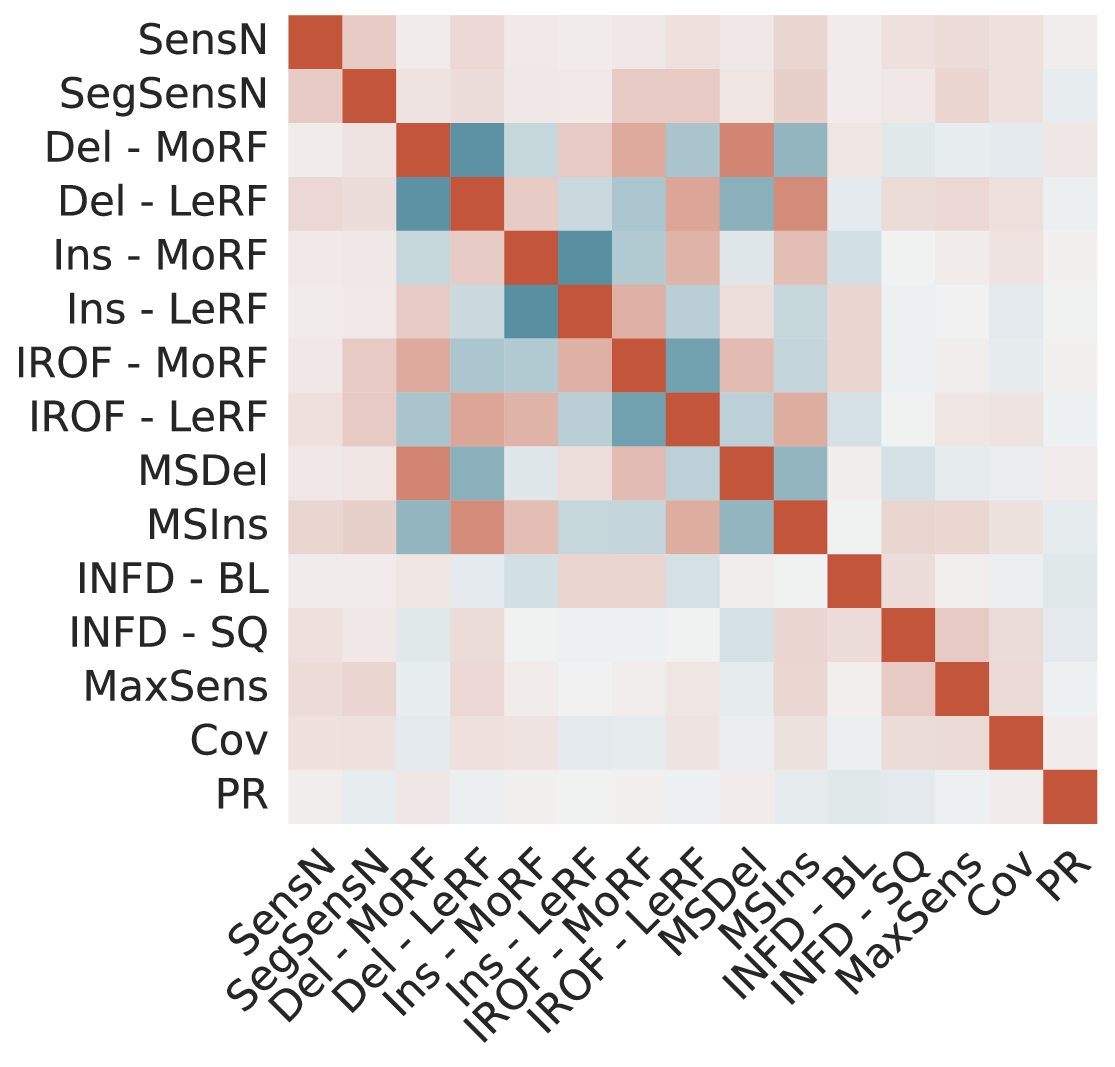}
			\caption{Places365}
		\end{subfigure}
		\caption{Inter-metric correlations for all datasets.}
		\label{app:corr-all}
	\end{figure}

	Full results of the paired t-tests for all metrics on all datasets are shown in Figures \ref{fig:results:ttest-all-lowdim}, \ref{fig:results:ttest-all-meddim} and \ref{fig:results:ttest-all-highdim} for the low-, medium- and high-dimensional datasets, respectively. Krippendorff $\alpha$ values for all metrics on all datasets are shown in Figure \ref{fig:alpha-bar-all}. Inter-metric correlations for all metrics on all datasets are shown in Figure \ref{app:corr-all}.
	
	\section{Applying the Guidelines to ImageNet}
	\label{app:guidelines-imagenet}
\begin{figure}
		\centering
		\includegraphics[height=0.5\textheight]{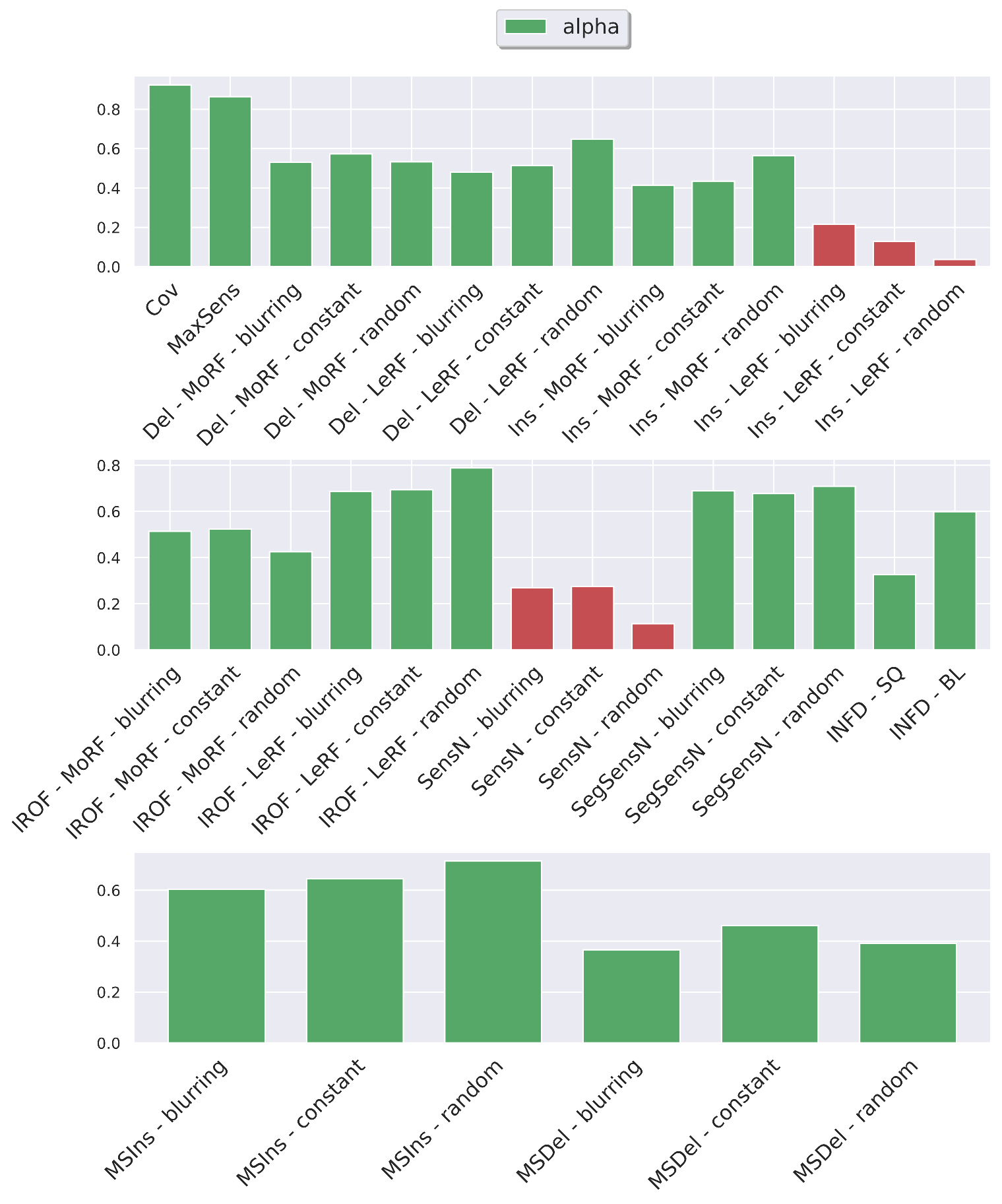}
		\caption{Krippendorff $\alpha$ for all metrics. Values above 0.3 are shown in green, others are shown in red.}
		\label{fig:krip-cs}
	\end{figure}

\begin{figure}
		\includegraphics[width=\textwidth]{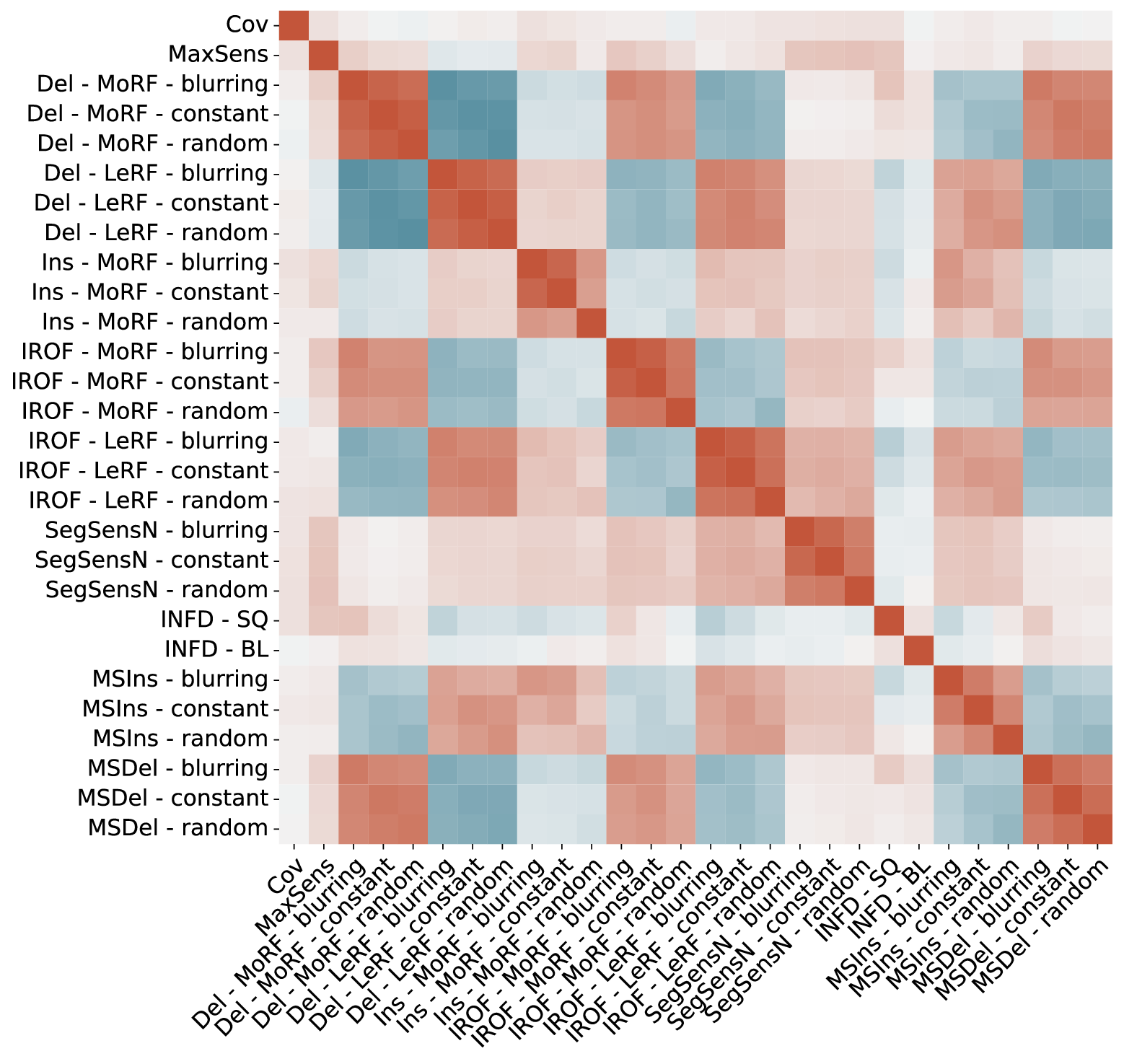}
		\caption{Inter-metric correlations for all metrics with Krippendorff $\alpha$ larger than 0.3.}
		\label{fig:imr-cs}
	\end{figure}

\begin{figure}
		\includegraphics[width=\textwidth]{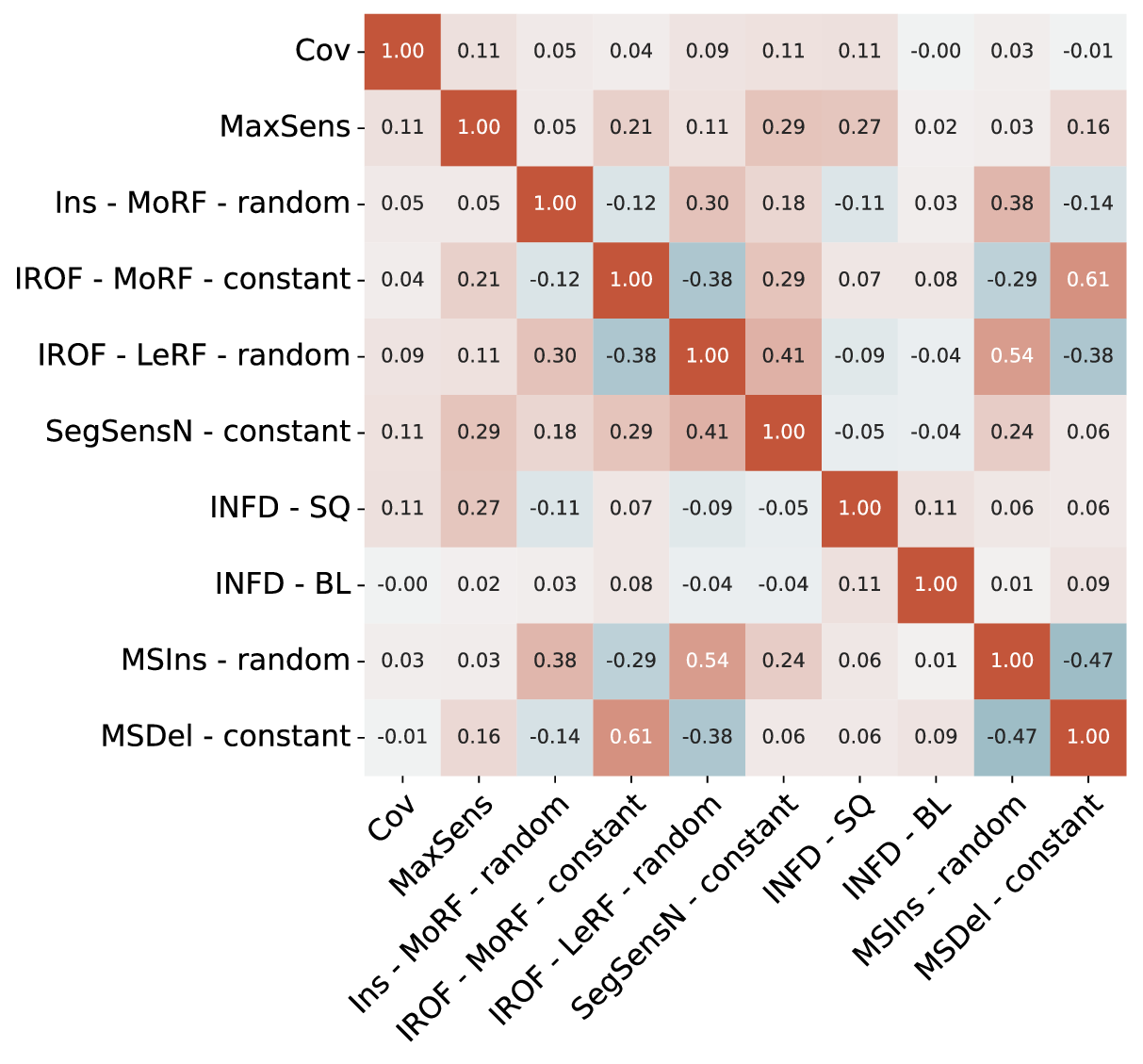}
		\caption{Inter-metric correlations of final selection of metrics.}
		\label{fig:imr-selected-cs}
	\end{figure}

\begin{figure}
		\includegraphics[width=\textwidth]{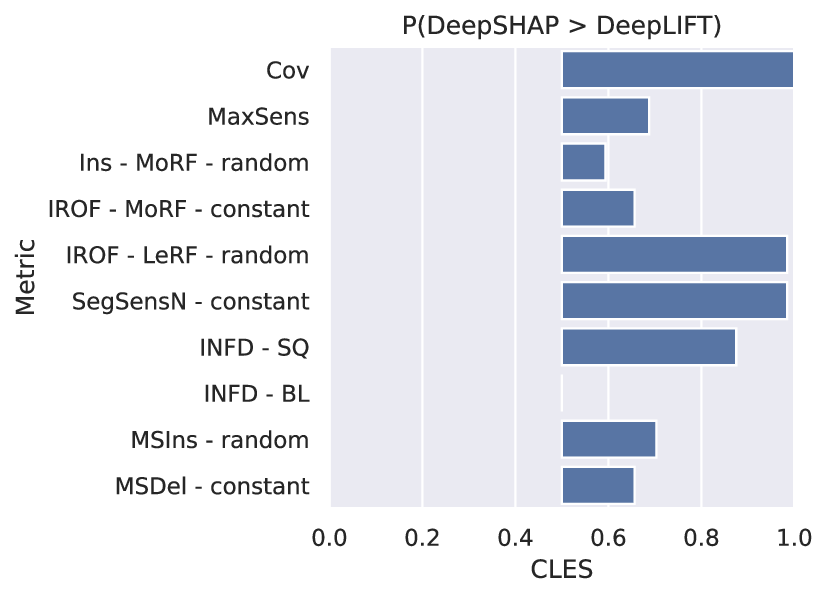}
		\caption{Pairwise comparison between DeepSHAP and DeepLIFT.}
		\label{fig:cles-cs-deepshap-deeplift}
	\end{figure}

\begin{figure}
		\includegraphics[width=\textwidth]{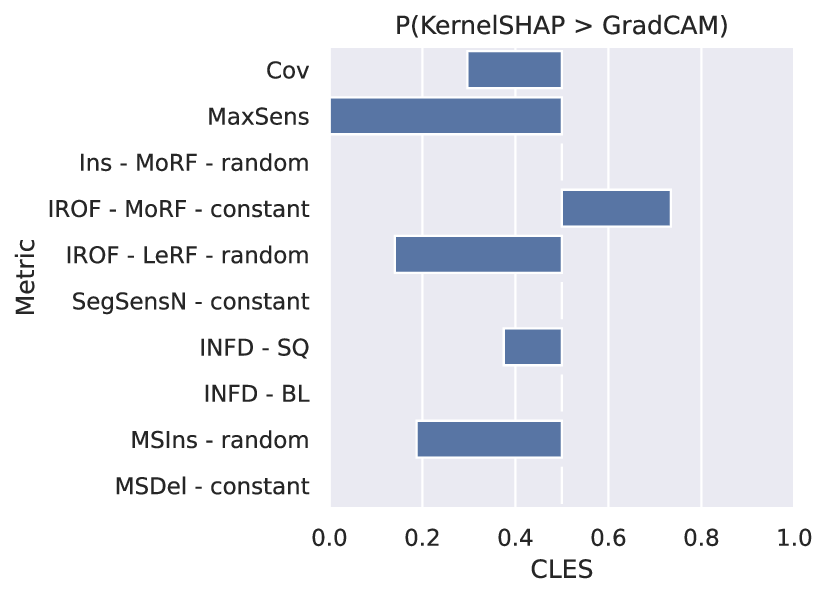}
		\caption{Pairwise comparison between KernelSHAP and GradCAM.}
		\label{fig:cles-cs-kernelshap-gradcam}
	\end{figure}

\begin{figure}
		\includegraphics[width=\textwidth]{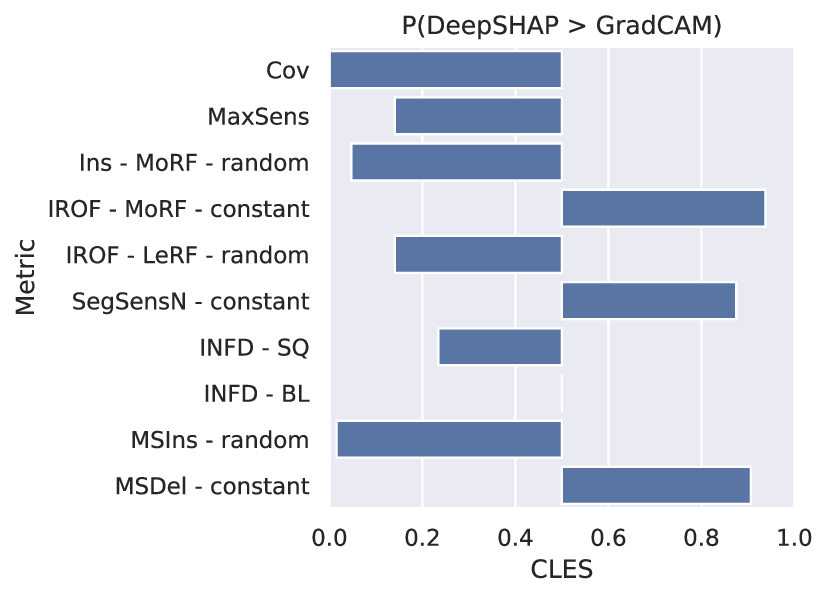}
		\caption{Pairwise comparison between DeepSHAP and GradCAM.}
		\label{fig:cles-cs-deepshap-gradcam}
	\end{figure}

	We apply the guidelines proposed in Section \ref{sec:guidelines} to ImageNet for demonstration purposes:
	\begin{enumerate}
		\item \textbf{Baseline selection:} Because we want to assume no prior knowledge about the problem setting, we select a uniform random baseline.
		\item \textbf{Metric selection:}
			\begin{enumerate}
				\item We run all metrics using 64 samples as a small pilot study. The results from this pilot study are used to compute values for Krippendorff $\alpha$. This is shown in Figure \ref{fig:krip-cs}. Metrics with $\alpha < 0.3$ are immediately discarded (shown in red in Figure \ref{fig:krip-cs}).
				\item We compute inter-metric correlations for the remaining metrics (shown in Figure \ref{fig:imr-cs}), and select metrics based on these correlations. The correlations for the final metric selection are shown in Figure \ref{fig:imr-selected-cs}.
					\begin{enumerate}
						\item Different implementations of the same metric have very high correlations. For each metric type, we select the implementation with the highest Krippendorff $\alpha$ value.
						\item There are strong correlations between Deletion and IROF. Because we are working with high-dimensional data, we choose IROF over Deletion.
					\end{enumerate}
			\end{enumerate}
		\item \textbf{Parameter Randomization test:} Next, we perform the Parameter Randomization test from \citep{Adebayo2018a}. We classify any method that obtains a Spearman rank correlation $\rho > 0.05$ as failing the test. All methods that fail the test are discarded.
		\item \textbf{Full benchmark:} Once the metrics and methods are selected, the metric scores are computed on a larger number of images. We use 256 images for the full benchmark.
		\item \textbf{Rough statistical analysis:} We perform paired t-tests on the results, comparing to the uniform random baseline. We observe two groups of methods with competitive but complementary results: the coarse-grained (CAM- and perturbaton-based) methods, and DeepSHAP, DeepLIFT and ExpectedGradients.
		\item \textbf{Detailed comparison:} 
			\begin{enumerate}
				\item From the previous analysis, we select DeepSHAP, DeepLIFT, KernelSHAP and GradCAM for further analysis. DeepSHAP can be viewed as the computationally more expensive method in the fine-grained group, whereas DeepLIFT is computationally much cheaper. An analogous comparison can be made between KernelSHAP and GradCAM for the coarse-grained group.
				\item We first compare DeepSHAP and DeepLIFT using the Probability of Superiority effect size. Results are shown in Figure \ref{fig:cles-cs-deepshap-deeplift}. We see that DeepSHAP significantly outperforms DeepLIFT on most metrics.
				\item Next, we compare GradCAM and KernelSHAP. Results are shown in Figure \ref{fig:cles-cs-kernelshap-gradcam}. Most of the paired t-tests between these two methods seem to be insignificant. Those metrics that do show a significant difference between the two methods disagree on which method is superior. Because no clear conclusion can be drawn in terms of the superiority of one method over the other, we select GradCAM for further analysis as it is computationally much cheaper than KernelSHAP.
				\item Finally, we compare DeepSHAP and GradCAM. Results for this comparison are shown in Figure \ref{fig:cles-cs-deepshap-gradcam}. We see that many of the paired t-test results are significant, but there is a strong complementarity in the results: some metrics favour DeepSHAP, others favour GradCAM. This suggests that both attribution maps might contain valuable complementary information. DeepSHAP is much more computationally expensive that GradCAM however, so depending on the use case, the developer might choose to provide both explanations, or to only use GradCAM explanations.
			\end{enumerate}
	\end{enumerate}
	

	\begin{figure}
		\centering
		\includegraphics[width=\textwidth]{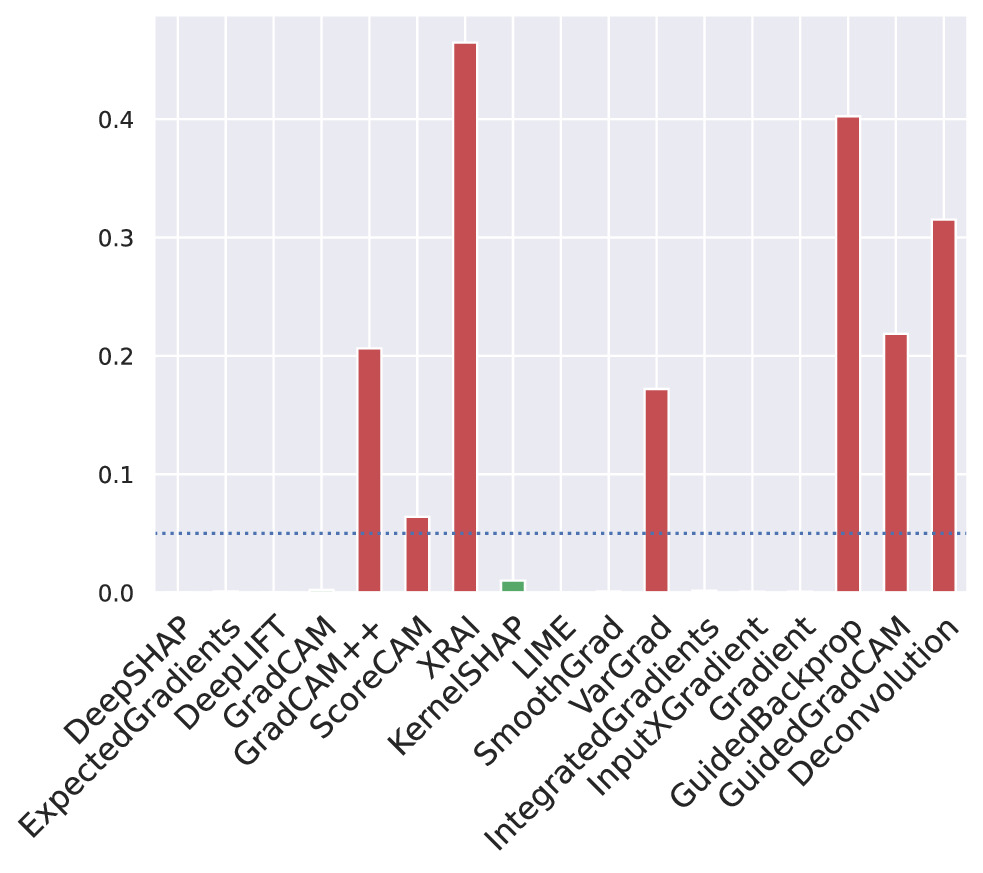}
		\caption{Results of the Parameter Randomization test. All methods with $\rho > 0.05$ are discarded.}
		\label{fig:pr-cs}
	\end{figure}
	
	\begin{figure}
		\centering
		\includegraphics[width=\textwidth]{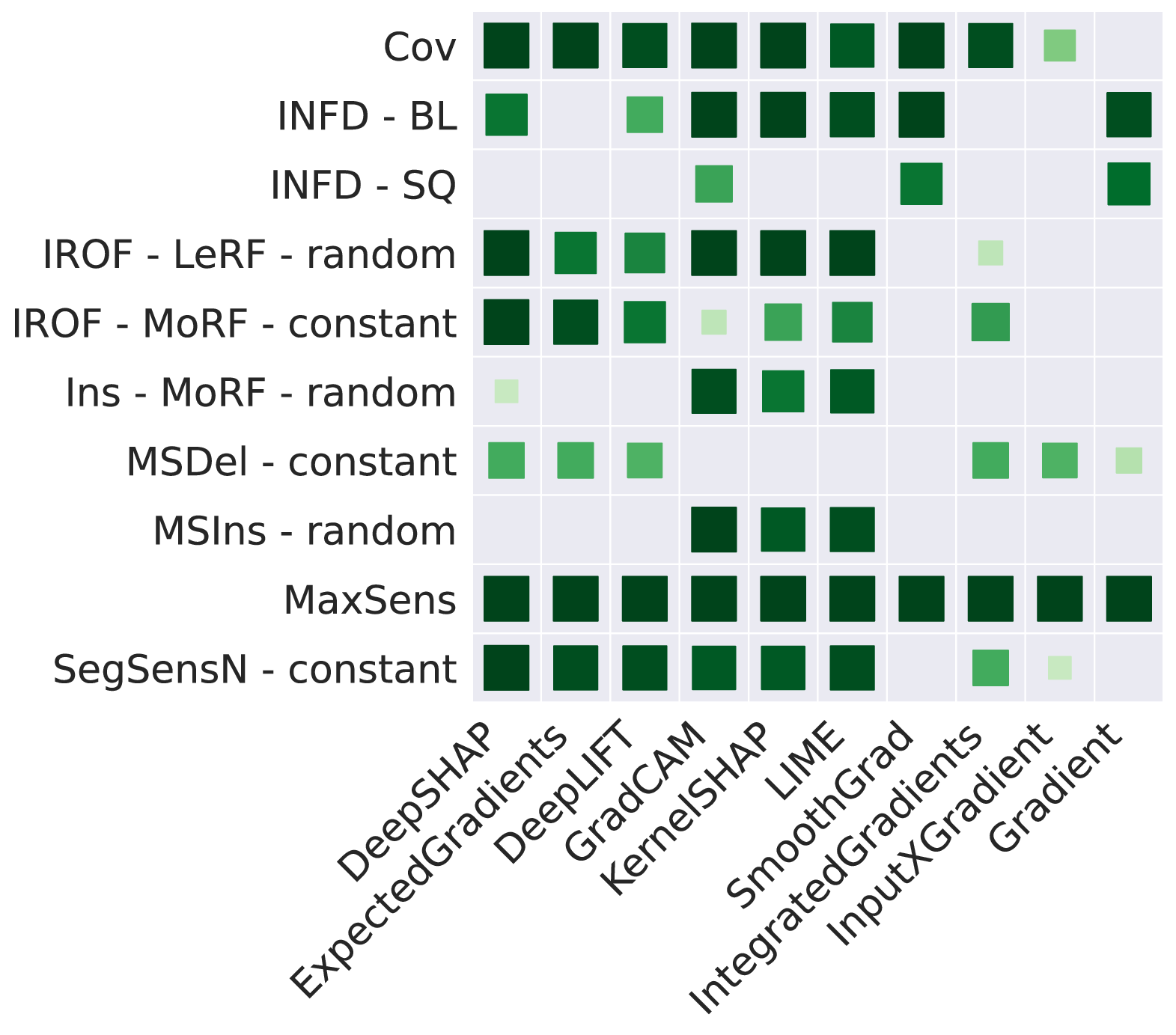}
		\caption{Wilcoxon Signed Rank test results for selection of metrics.}
		\label{fig:wsrt-cs}
	\end{figure}

	\section{Evaluating Attributions on Tabular Datasets}
	\label{app:tabular}
	To demonstrate the general applicability of the proposed methodology, we
	compute and evaluate feature attributions for tabular data. We use the
	Adult \citep{adult_dataset}, DNA \citep{dna_dataset}, Satimage \citep{statlog_landsat} and Spambase \citep{spambase_dataset} datasets from the OpenML repository \citep{OpenML2013}.
	
	On each of these datasets, a fully-connected neural network with two hidden layers
	of 64 neurons each is trained. Because this model has no convolutional layers,
	only the methods that require differentiability of the model or that have
	no model requirements at all were evaluated (see Table \ref{tbl:methods}).
	We evaluated all metrics that are applicable to any type of data (see Table \ref{tbl:metrics}).
	Masking was done by replacing feature values by 0, which after standard scaling
	of the features is equivalent to masking using the feature mean value.
	
	\subsection{Paired t-tests}
\begin{figure}
		\begin{subfigure}[b]{0.49\textwidth}
			\includegraphics[width=\textwidth]{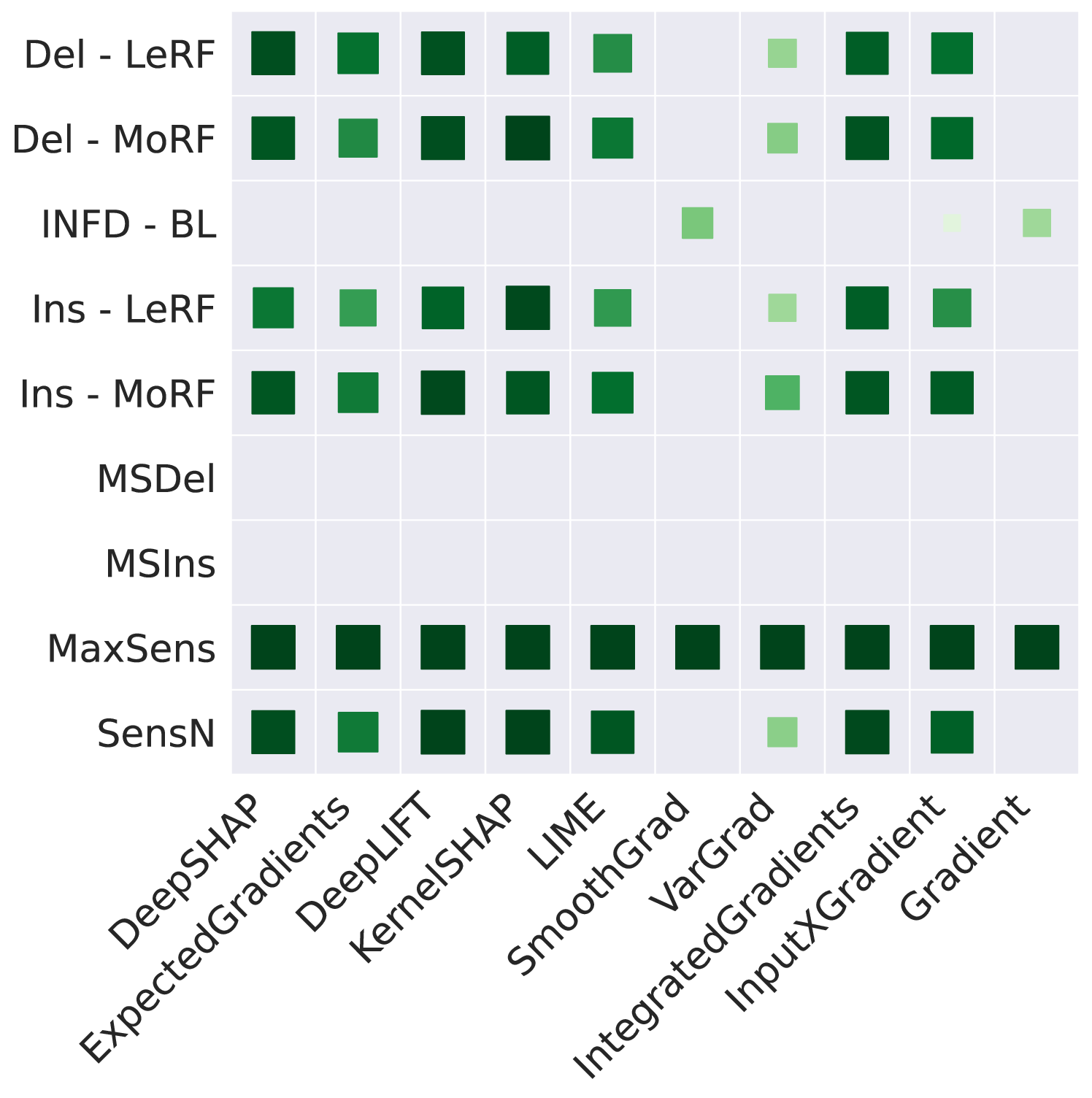}
			\caption{Adult}
			\label{fig:wsrt-adult}
		\end{subfigure}
		\begin{subfigure}[b]{0.49\textwidth}
			\includegraphics[width=\textwidth]{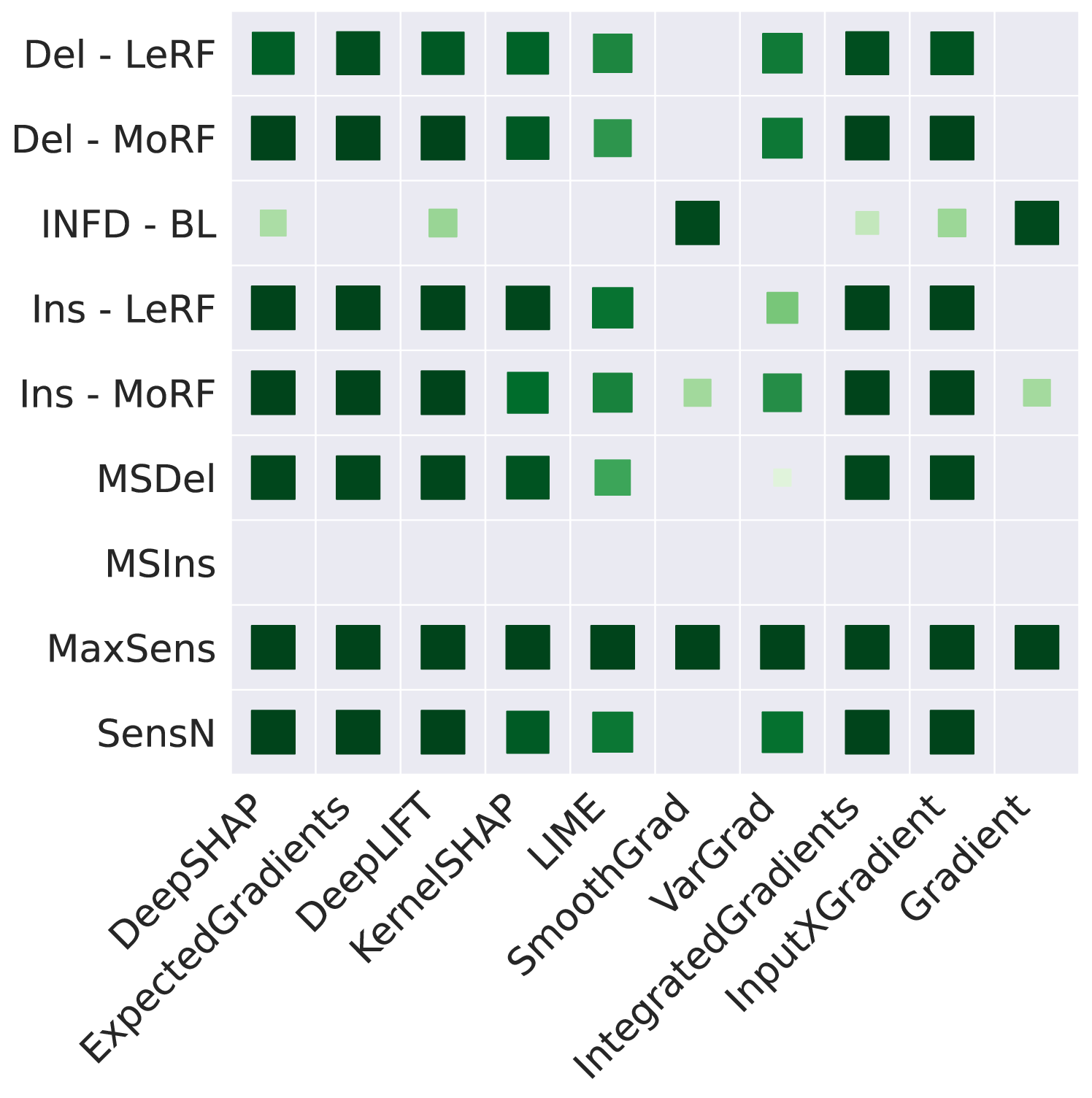}
			\caption{DNA}
			\label{fig:wsrt-dna}
		\end{subfigure}%
		
		\begin{subfigure}[b]{0.49\textwidth}
			\includegraphics[width=\textwidth]{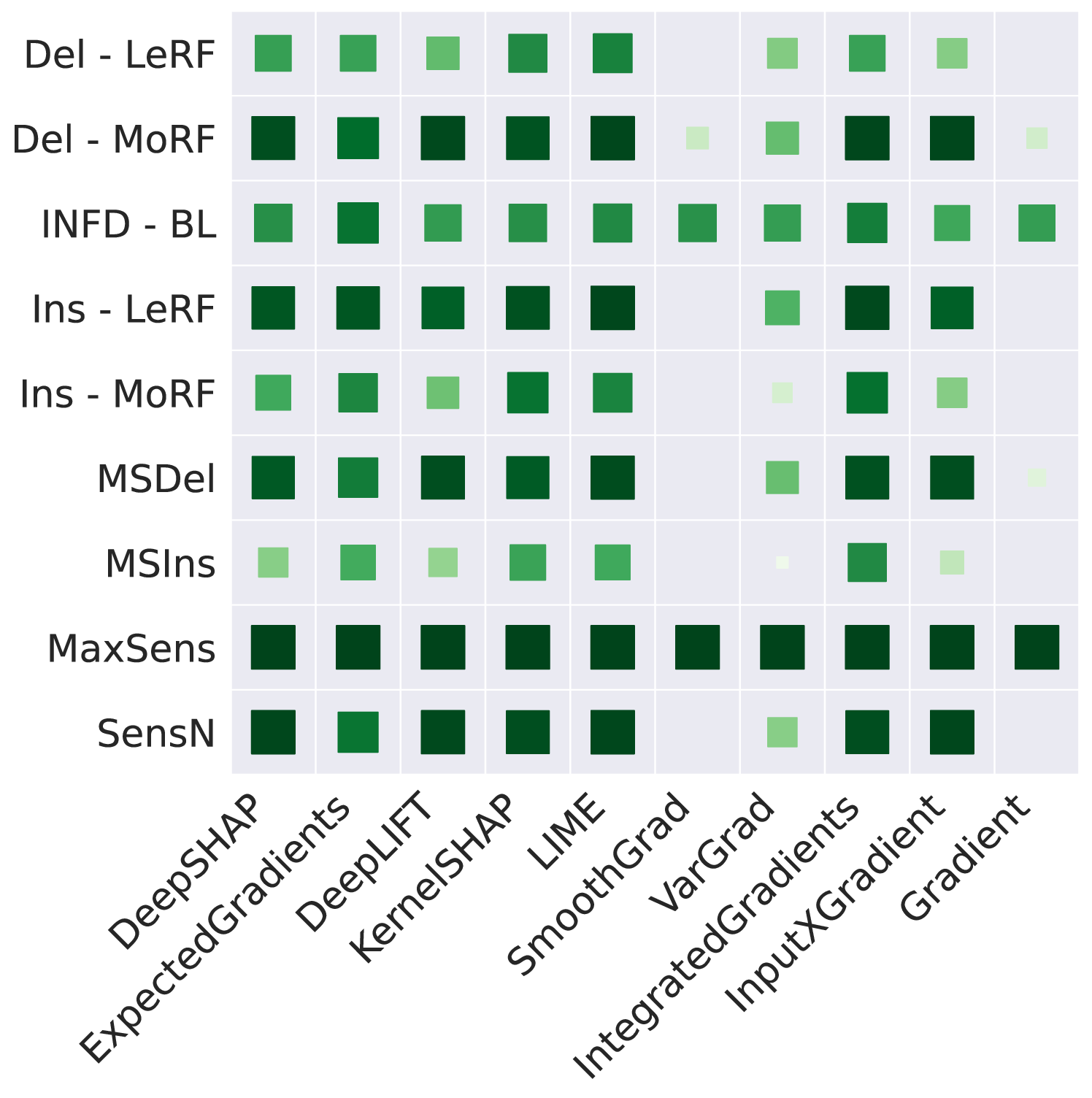}
			\caption{Satimage}
			\label{fig:wsrt-satimage}
		\end{subfigure}
		\begin{subfigure}[b]{0.49\textwidth}
			\includegraphics[width=\textwidth]{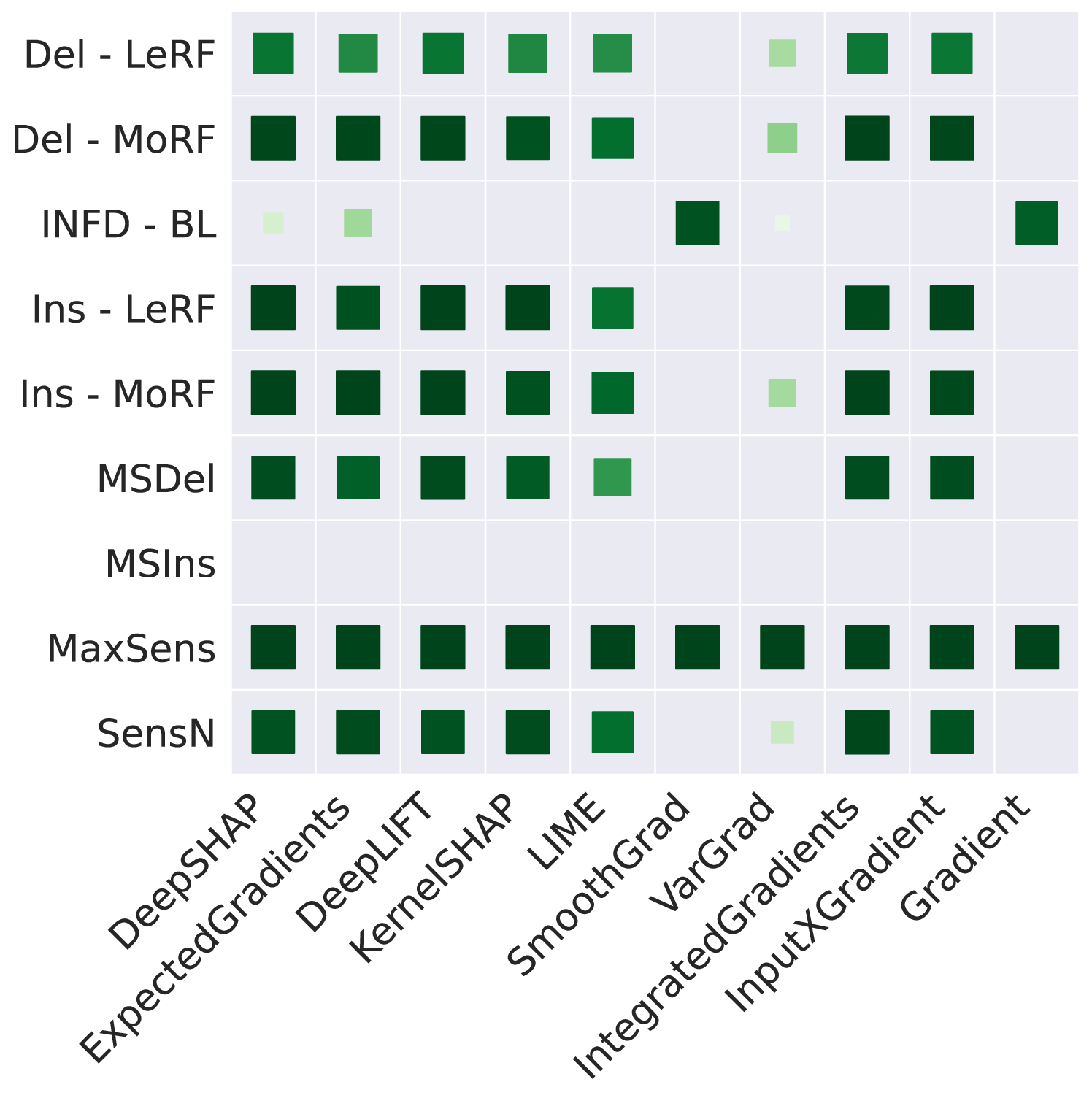}
			\caption{Spambase}
			\label{fig:wsrt-spambase}
		\end{subfigure}
		\caption{Results of paired t-tests on tabular datasets.}
		\label{fig:wsrt-tabular}
	\end{figure}

	Results of the paired t-tests are shown in Figure \ref{fig:wsrt-tabular}.
	We see that the results are very similar across the four datasets, with the
	exception of the minimal subset metrics. This might be linked to the number of
	classes in the dataset, which is 2 for the Adult and Spambase datasets,
	3 for the DNA dataset and 6 for the Satimage dataset.
	
	If the dataset has only 2 classes, a constant 0 vector has a 50\% probability
	of producing the same output as the original sample (assuming no class imbalance).
	Combined with the fact that these datasets have much fewer features than image datasets, there is
	a reasonable probability that masking any number of features for a given sample
	will not flip the result. In this case, both minimal subset insertion and minimal
	subset deletion will have a constant score for all methods. This might explain
	why the minimal subset metrics have much higher p-values for the Adult and
	Spambase datasets than for the DNA and Satimage datasets. More experiments
	can be done to verify this hypothesis.
	
	\subsection{Inter-Metric Correlations}
\begin{figure}
		\centering
		\begin{subfigure}[b]{0.49\textwidth}
			\includegraphics[width=\textwidth]{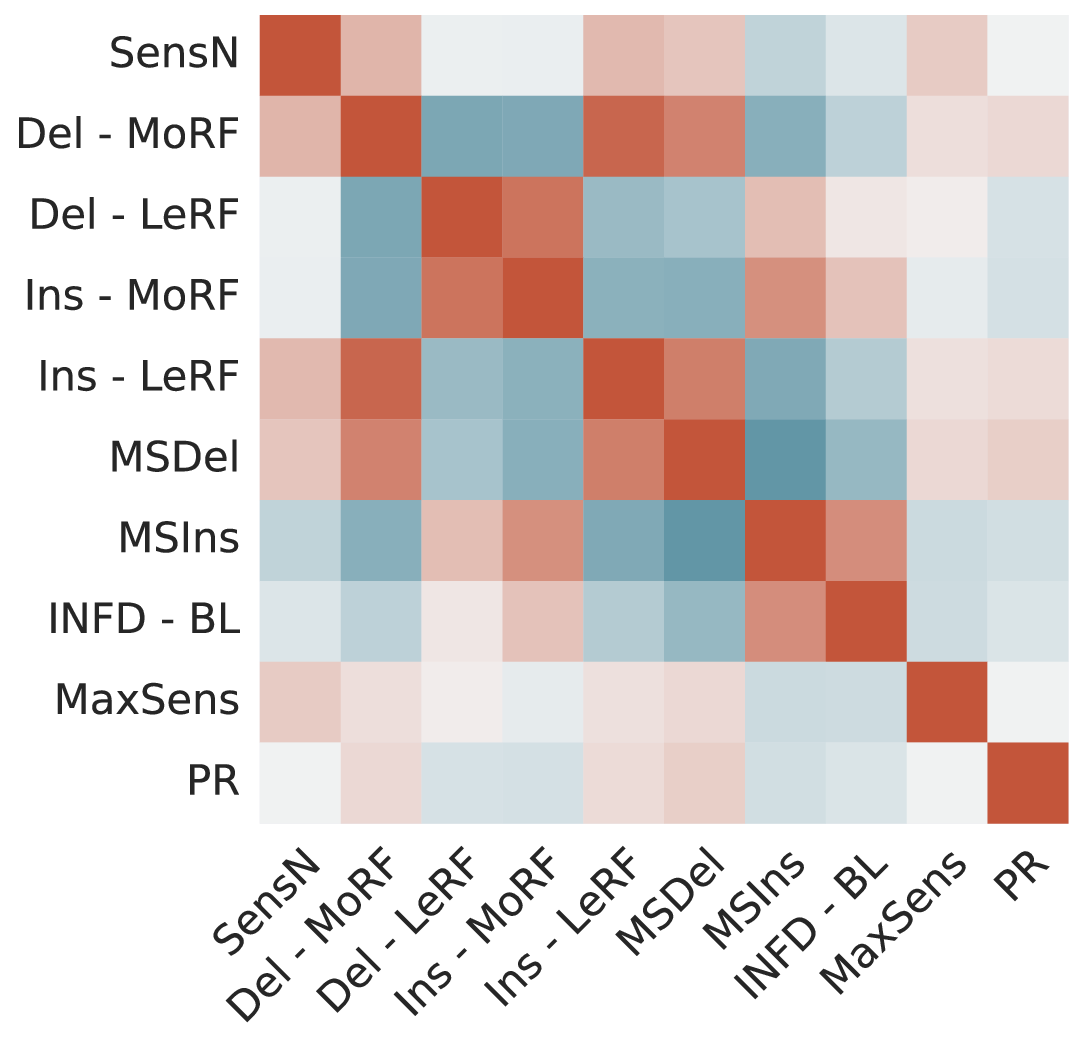}
			\caption{Adult}
			\label{fig:metric-corr-adult}
		\end{subfigure}
		\begin{subfigure}[b]{0.49\textwidth}
			\includegraphics[width=\textwidth]{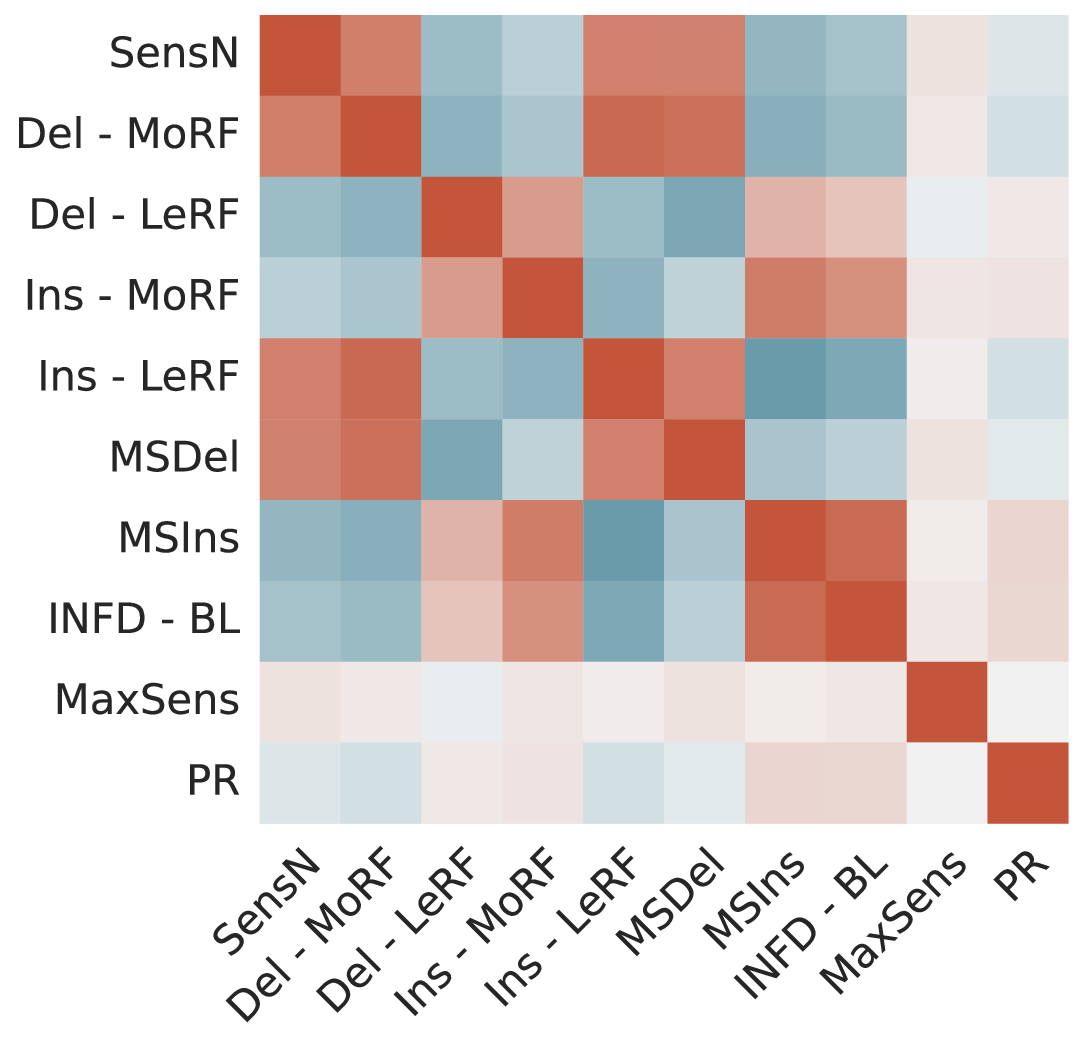}
			\caption{DNA}
			\label{fig:metric-corr-dna}
		\end{subfigure}%
		
		\begin{subfigure}[b]{0.49\textwidth}
			\includegraphics[width=\textwidth]{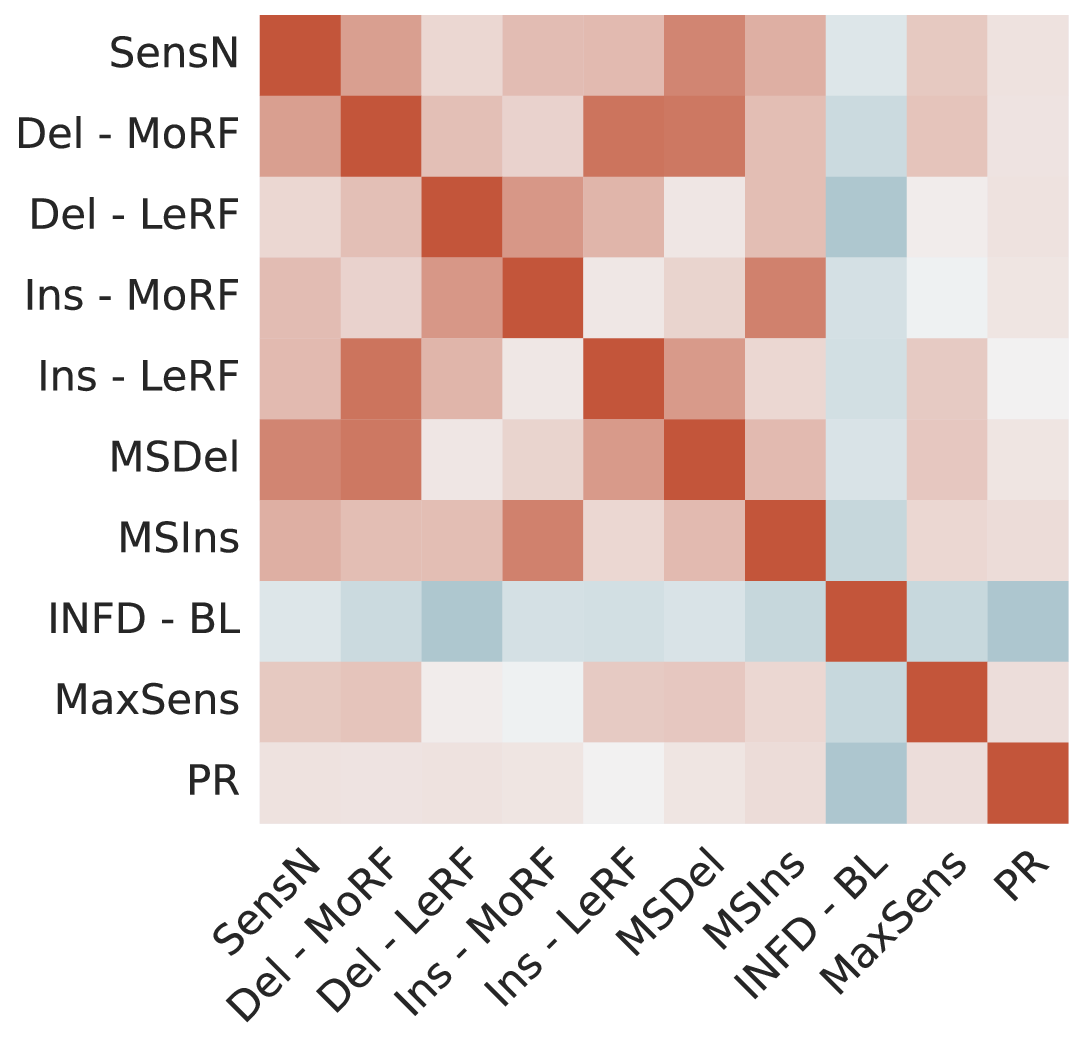}
			\caption{Satimage}
			\label{fig:metric-corr-satimage}
		\end{subfigure}
		\begin{subfigure}[b]{0.49\textwidth}
			\includegraphics[width=\textwidth]{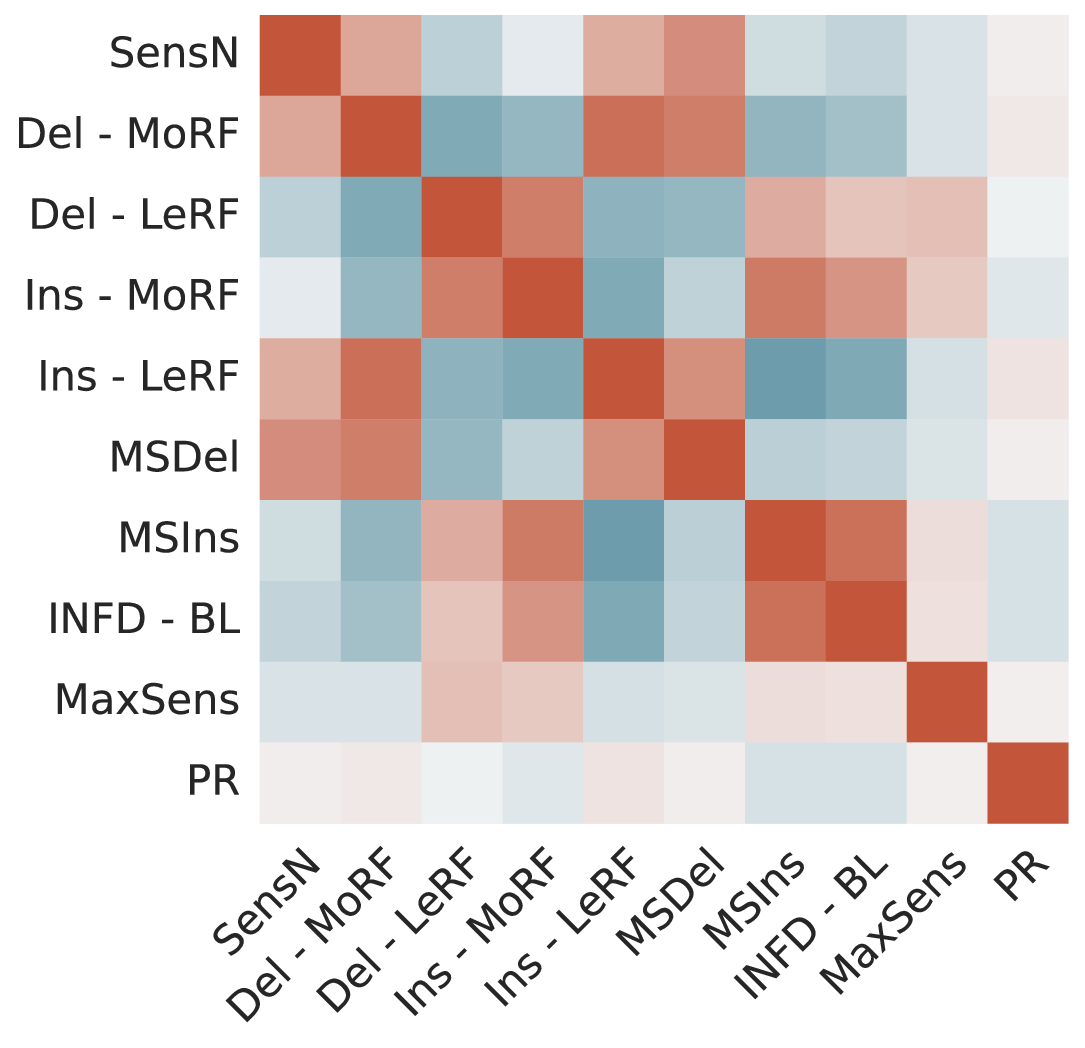}
			\caption{Spambase}
			\label{fig:metric-corr-spambase}
		\end{subfigure}
		\caption{Inter-metric correlations for tabular datasets.}
		\label{fig:metric-corr-tabular}
	\end{figure}

	Inter-metric correlations for the tabular datasets are shown in Figure
	\ref{fig:metric-corr-tabular}. We see that the correlations vary strongly
	across the four datasets. This observation confirms the need for a separate
	benchmarking experiment for each dataset, as proposed in Section \ref{sec:guidelines}.
	
	\subsection{Ranking Consistency}
\begin{figure}
		\centering
		\includegraphics[width=\textwidth]{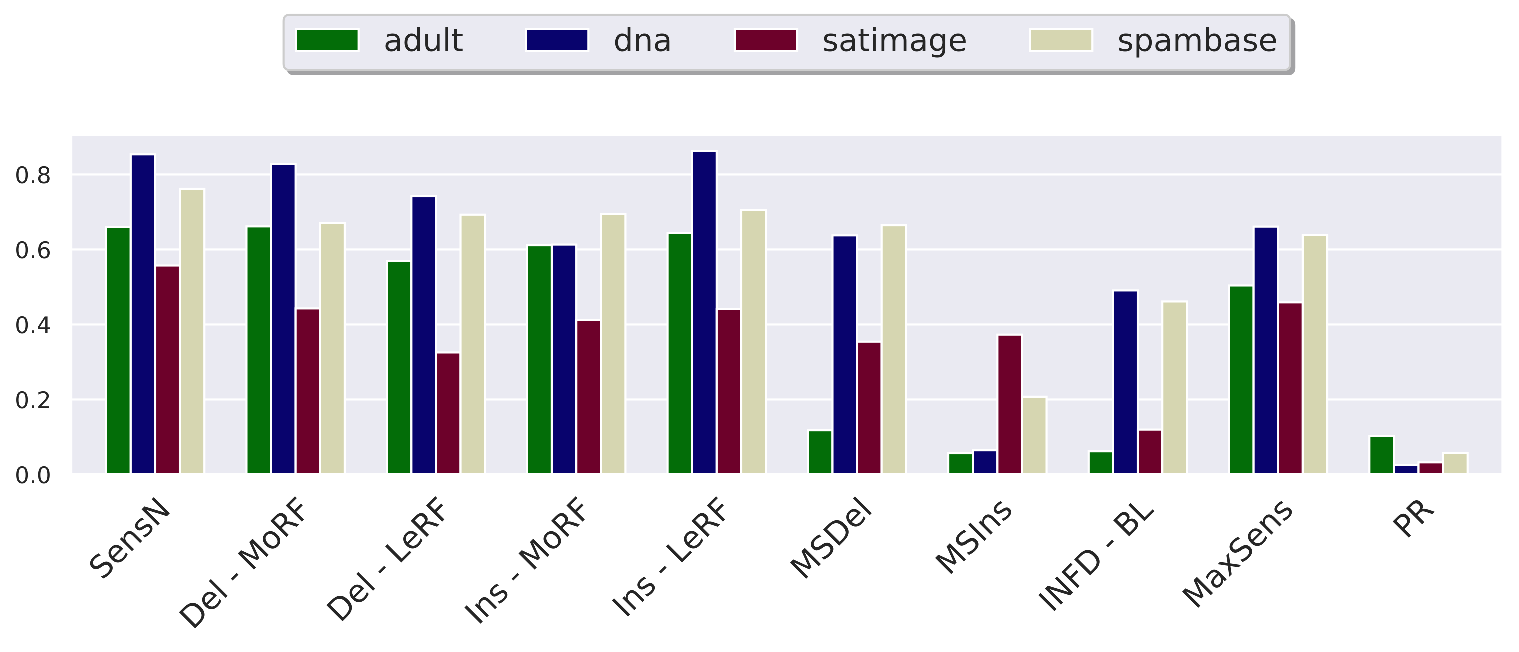}
		\caption{Krippendorff $\alpha$ for tabular datasets.}
		\label{fig:alpha-tabular}
	\end{figure}

	Ranking consistency for the tabular datasets is visualized in Figure \ref{fig:alpha-tabular}.
	We see that the ranking consistency is generally a bit higher than for the image
	datasets, but there is again a large variation across datasets.
	
	\subsection{Parameter Randomization}
\begin{figure}
		\centering
		\includegraphics[width=\textwidth]{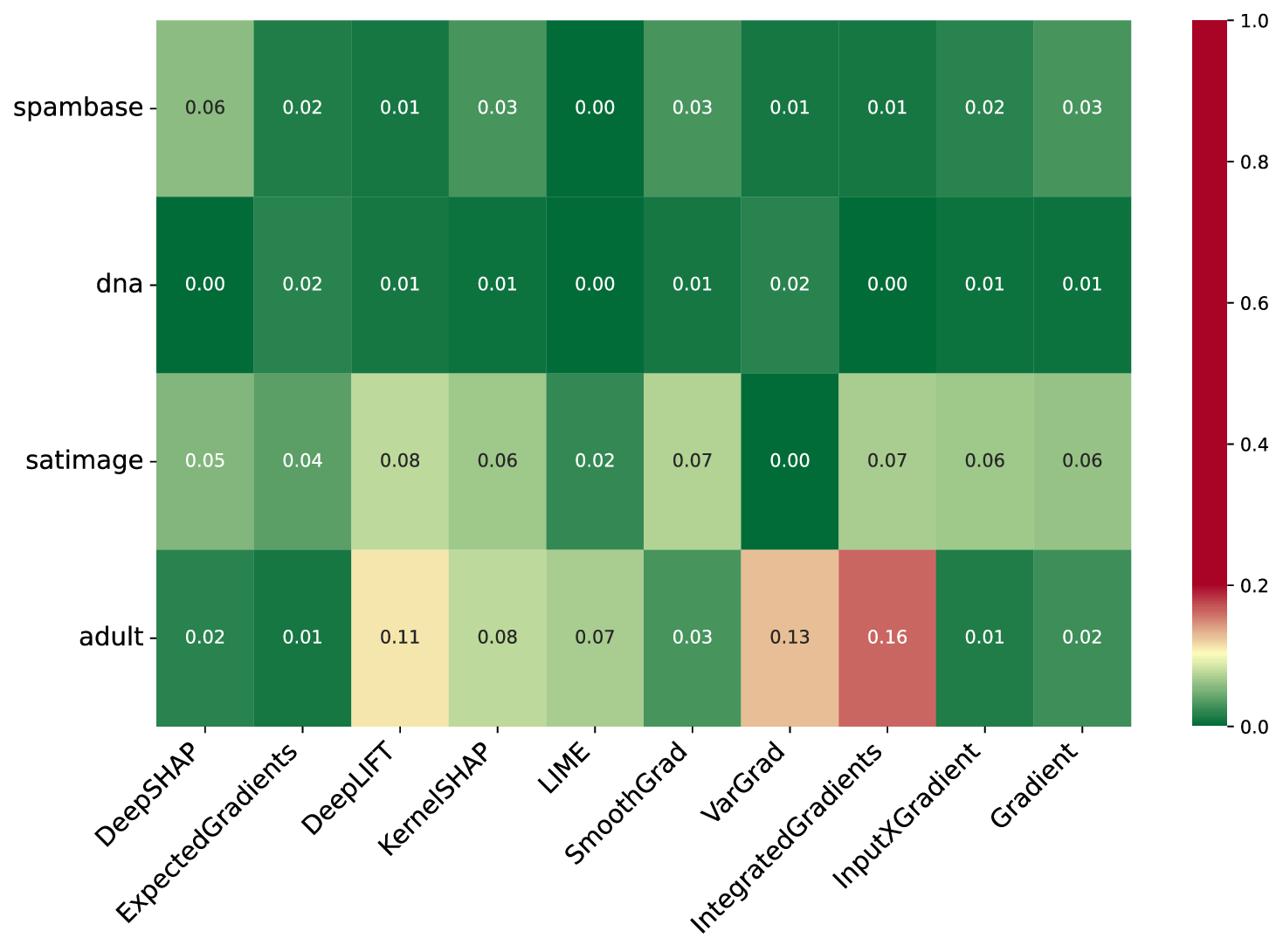}
		\caption{Results of the Parameter Randomization metric on tabular datasets.}
		\label{fig:parameter-randomization-tabular}
	\end{figure}

	Results of the Parameter Randomization test are shown in Figure \ref{fig:parameter-randomization-tabular}. We see that the absolute rank correlation
	is very low in most cases, with the exception of the Adult dataset. Although
	the absolute rank correlation remains below 0.25, it is notably higher than
	for the other datasets. This might be explained by the fact that the Adult
	dataset has the fewest features of all four datasets, which makes it easier
	for a random permutation of the features to have a higher rank correlation
	with the original ordering. Further experiments, where the rank correlation
	is corrected for the total number of features, can be done to verify or
	refute this hypothesis.
	
\end{appendices}

\section*{Declarations}
\subsection*{Funding}
The research leading to these results has received funding from the Flemish Government under the ``Onderzoeksprogramma Artifici\"ele Intelligentie (AI) Vlaanderen'' programme, and from the BOF project 01D13919.

\subsection*{Code availability}
The Python code associated with this work is available at \url{https://github.com/arnegevaert/benchmark-general-imaging}.

\subsection*{Competing interests}
The authors have no competing interests to declare that are relevant to the content of this article.

\subsection*{Ethics approval}
Not applicable.

\subsection*{Consent to participate}
Not applicable.

\subsection*{Consent for publication}
Not applicable.

\subsection*{Availability of data and material}
The necessary code to reproduce the experiments is available at \url{https://github.com/arnegevaert/benchmark-general-imaging}. The implementation of attribution metrics is available at \url{https://github.com/arnegevaert/benchmark}. Datasets, results of the experiments and model parameters are available at \url{https://zenodo.org/record/6221586#.Yos8IHVByV4} and \url{https://zenodo.org/record/6205531#.Yos8H3VByV4}, respectively.

\subsection*{Authors' contributions}
Arne Gevaert and Axel-Jan Rousseau have jointly implemented the different attribution methods and metrics used in the study. Arne Gevaert has then performed the experiments and statistical analysis of the results. Thijs Becker, Dirk Valkenborg, Tijl De Bie and Yvan Saeys have supervised the project and assisted in writing the paper.

\vskip 0.2in
\bibliographystyle{abbrvnat}
\bibliography{bibliography}

\begin{thebibliography}{72}
\providecommand{\natexlab}[1]{#1}
\providecommand{\url}[1]{\texttt{#1}}
\expandafter\ifx\csname urlstyle\endcsname\relax
  \providecommand{\doi}[1]{doi: #1}\else
  \providecommand{\doi}{doi: \begingroup \urlstyle{rm}\Url}\fi

\bibitem[Achanta et~al.(2012)Achanta, Shaji, Smith, Lucchi, Fua, and Suesstrunk]{achanta2010slic}
R.~Achanta, A.~Shaji, K.~Smith, A.~Lucchi, P.~Fua, and S.~Suesstrunk.
\newblock Slic superpixels compared to state-of-the-art superpixel methods.
\newblock \emph{IEEE Transactions on Pattern Analysis and Machine Intelligence}, 34\penalty0 (11):\penalty0 2274--2281, 2012.
\newblock ISSN 0162-8828.
\newblock \doi{10.1109/TPAMI.2012.120}.

\bibitem[Adebayo et~al.(2018)Adebayo, Gilmer, Muelly, Goodfellow, Hardt, and Kim]{Adebayo2018a}
J.~Adebayo, J.~Gilmer, M.~Muelly, I.~Goodfellow, M.~Hardt, and B.~Kim.
\newblock Sanity checks for saliency maps.
\newblock In S.~Bengio, H.~Wallach, H.~Larochelle, K.~Grauman, N.~Cesa-Bianchi, and R.~Garnett, editors, \emph{Advances in Neural Information Processing Systems}, volume~31. Curran Associates, Inc., 2018.

\bibitem[Ancona et~al.(2018)Ancona, Ceolini, Öztireli, and Gross]{Ancona2017}
M.~Ancona, E.~Ceolini, C.~Öztireli, and M.~Gross.
\newblock Towards better understanding of gradient-based attribution methods for deep neural networks.
\newblock In \emph{International Conference on Learning Representations}, 2018.
\newblock URL \url{https://openreview.net/forum?id=Sy21R9JAW}.

\bibitem[Balduzzi et~al.(2017)Balduzzi, Frean, Leary, Lewis, Ma, and McWilliams]{balduzzi2017shattered}
D.~Balduzzi, M.~Frean, L.~Leary, J.~P. Lewis, K.~W.-D. Ma, and B.~McWilliams.
\newblock The shattered gradients problem: If resnets are the answer, then what is the question?
\newblock In D.~Precup and Y.~Teh, editors, \emph{INTERNATIONAL CONFERENCE ON MACHINE LEARNING}, volume~70 of \emph{Proceedings of Machine Learning Research}, 2017.

\bibitem[Becker and Kohavi(1996)]{adult_dataset}
B.~Becker and R.~Kohavi.
\newblock Adult.
\newblock UCI Machine Learning Repository, 1996.

\bibitem[Binder et~al.(2023)Binder, Weber, Lapuschkin, Montavon, M\"uller, and Samek]{binder2023}
A.~Binder, L.~Weber, S.~Lapuschkin, G.~Montavon, K.-R. M\"uller, and W.~Samek.
\newblock Shortcomings of top-down randomization-based sanity checks for evaluations of deep neural network explanations.
\newblock In \emph{Proceedings of the IEEE/CVF Conference on Computer Vision and Pattern Recognition (CVPR)}, pages 16143--16152, June 2023.

\bibitem[Bland and Altman(1995)]{Bland170}
J.~M. Bland and D.~G. Altman.
\newblock Multiple significance tests: the bonferroni method.
\newblock \emph{BMJ}, 310\penalty0 (6973):\penalty0 170, 1995.
\newblock ISSN 0959-8138.
\newblock \doi{10.1136/bmj.310.6973.170}.
\newblock URL \url{https://www.bmj.com/content/310/6973/170}.

\bibitem[Bradski(2000)]{opencv_library}
G.~Bradski.
\newblock {The OpenCV Library}.
\newblock \emph{Dr. Dobb's Journal of Software Tools}, 2000.

\bibitem[Brown et~al.(2020)Brown, Mann, Ryder, Subbiah, Kaplan, Dhariwal, Neelakantan, Shyam, Sastry, Askell, Agarwal, Herbert-Voss, Krueger, Henighan, Child, Ramesh, Ziegler, Wu, Winter, Hesse, Chen, Sigler, Litwin, Gray, Chess, Clark, Berner, McCandlish, Radford, Sutskever, and Amodei]{brown2020}
T.~Brown, B.~Mann, N.~Ryder, M.~Subbiah, J.~D. Kaplan, P.~Dhariwal, A.~Neelakantan, P.~Shyam, G.~Sastry, A.~Askell, S.~Agarwal, A.~Herbert-Voss, G.~Krueger, T.~Henighan, R.~Child, A.~Ramesh, D.~Ziegler, J.~Wu, C.~Winter, C.~Hesse, M.~Chen, E.~Sigler, M.~Litwin, S.~Gray, B.~Chess, J.~Clark, C.~Berner, S.~McCandlish, A.~Radford, I.~Sutskever, and D.~Amodei.
\newblock Language models are few-shot learners.
\newblock In H.~Larochelle, M.~Ranzato, R.~Hadsell, M.~Balcan, and H.~Lin, editors, \emph{Advances in Neural Information Processing Systems}, volume~33, pages 1877--1901. Curran Associates, Inc., 2020.

\bibitem[Chandrashekar and Sahin(2014)]{chandrashekar2014survey}
G.~Chandrashekar and F.~Sahin.
\newblock A survey on feature selection methods.
\newblock \emph{Computers \& Electrical Engineering}, 40\penalty0 (1):\penalty0 16--28, 2014.
\newblock ISSN 0045-7906.
\newblock \doi{https://doi.org/10.1016/j.compeleceng.2013.11.024}.

\bibitem[Chattopadhay et~al.(2018)Chattopadhay, Sarkar, Howlader, and Balasubramanian]{chattopadhay2018gradcampp}
A.~Chattopadhay, A.~Sarkar, P.~Howlader, and V.~N. Balasubramanian.
\newblock Grad-cam++: Generalized gradient-based visual explanations for deep convolutional networks.
\newblock In \emph{2018 IEEE Winter Conference on Applications of Computer Vision (WACV)}, pages 839--847, 2018.
\newblock \doi{10.1109/WACV.2018.00097}.

\bibitem[Chen et~al.(2020)Chen, Janizek, Lundberg, and Lee]{chen2020}
H.~Chen, J.~D. Janizek, S.~Lundberg, and S.-I. Lee.
\newblock True to the model or true to the data?, 2020.

\bibitem[Cohen(1988)]{Cohen1988}
J.~Cohen.
\newblock \emph{Statistical Power Analysis for the Behavioral Sciences (2nd ed.)}.
\newblock Routledge, May 1988.
\newblock ISBN 9781134742707.
\newblock \doi{10.4324/9780203771587}.

\bibitem[Dalal and Triggs(2005)]{dalal2005hogs}
N.~Dalal and B.~Triggs.
\newblock Histograms of oriented gradients for human detection.
\newblock In \emph{2005 IEEE Computer Society Conference on Computer Vision and Pattern Recognition (CVPR'05)}, volume~1, pages 886--893 vol. 1, 2005.
\newblock \doi{10.1109/CVPR.2005.177}.

\bibitem[Dandl et~al.(2020)Dandl, Molnar, Binder, and Bischl]{dandl2020multi}
S.~Dandl, C.~Molnar, M.~Binder, and B.~Bischl.
\newblock Multi-objective counterfactual explanations.
\newblock In T.~B{\"a}ck, M.~Preuss, A.~Deutz, H.~Wang, C.~Doerr, M.~Emmerich, and H.~Trautmann, editors, \emph{Parallel Problem Solving from Nature -- PPSN XVI}, pages 448--469, Cham, 2020. Springer International Publishing.
\newblock ISBN 978-3-030-58112-1.

\bibitem[Deng et~al.(2009)Deng, Dong, Socher, Li, Li, and Fei-Fei]{deng2009}
J.~Deng, W.~Dong, R.~Socher, L.-J. Li, K.~Li, and L.~Fei-Fei.
\newblock Imagenet: A large-scale hierarchical image database.
\newblock In \emph{2009 IEEE Conference on Computer Vision and Pattern Recognition}, pages 248--255, 2009.
\newblock \doi{10.1109/CVPR.2009.5206848}.

\bibitem[Doshi-Velez and Kim(2017)]{Doshi-Velez2017}
F.~Doshi-Velez and B.~Kim.
\newblock Towards a rigorous science of interpretable machine learning, 2017.

\bibitem[Erion et~al.(2021)Erion, Janizek, Sturmfels, Lundberg, and Lee]{erion2020}
G.~Erion, J.~D. Janizek, P.~Sturmfels, S.~M. Lundberg, and S.-I. Lee.
\newblock Improving performance of deep learning models with axiomatic attribution priors and expected gradients.
\newblock \emph{Nature Machine Intelligence}, 3\penalty0 (7):\penalty0 620--631, 2021.
\newblock \doi{10.1038/s42256-021-00343-w}.

\bibitem[Fernandez(2020)]{Fernandez_TorchCAM_class_activation_2021}
F.-G. Fernandez.
\newblock Torchcam: class activation explorer.
\newblock \url{https://github.com/frgfm/torch-cam}, March 2020.

\bibitem[Fong and Vedaldi(2017)]{Fong2017}
R.~C. Fong and A.~Vedaldi.
\newblock Interpretable explanations of black boxes by meaningful perturbation.
\newblock In \emph{2017 IEEE International Conference on Computer Vision (ICCV)}, pages 3449--3457, 2017.
\newblock \doi{10.1109/ICCV.2017.371}.

\bibitem[Ghorbani et~al.(2019)Ghorbani, Abid, and Zou]{Ghorbani2019}
A.~Ghorbani, A.~Abid, and J.~Zou.
\newblock Interpretation of neural networks is fragile.
\newblock \emph{Proceedings of the AAAI Conference on Artificial Intelligence}, 33\penalty0 (01):\penalty0 3681--3688, Jul. 2019.
\newblock \doi{10.1609/aaai.v33i01.33013681}.
\newblock URL \url{https://ojs.aaai.org/index.php/AAAI/article/view/4252}.

\bibitem[Griffin et~al.(2022)Griffin, Holub, and Perona]{griffin2022}
G.~Griffin, A.~Holub, and P.~Perona.
\newblock Caltech 256, Apr 2022.

\bibitem[He et~al.(2016)He, Zhang, Ren, and Sun]{he2015}
K.~He, X.~Zhang, S.~Ren, and J.~Sun.
\newblock Deep residual learning for image recognition.
\newblock In \emph{2016 IEEE Conference on Computer Vision and Pattern Recognition (CVPR)}, pages 770--778, 2016.
\newblock \doi{10.1109/CVPR.2016.90}.

\bibitem[Hedstr{\"o}m et~al.(2023)Hedstr{\"o}m, Weber, Lapuschkin, and H{\"o}hne]{hedstrom2023a}
A.~Hedstr{\"o}m, L.~Weber, S.~Lapuschkin, and M.~H{\"o}hne.
\newblock Sanity checks revisited: An exploration to repair the model parameter randomisation test.
\newblock In \emph{XAI in Action: Past, Present, and Future Applications}, 2023.
\newblock URL \url{https://openreview.net/forum?id=vVpefYmnsG}.

\bibitem[Hooker et~al.(2019)Hooker, Erhan, Kindermans, and Kim]{Hooker2018}
S.~Hooker, D.~Erhan, P.-J. Kindermans, and B.~Kim.
\newblock A benchmark for interpretability methods in deep neural networks.
\newblock In H.~Wallach, H.~Larochelle, A.~Beygelzimer, F.~d\textquotesingle Alch\'{e}-Buc, E.~Fox, and R.~Garnett, editors, \emph{Advances in Neural Information Processing Systems}, volume~32. Curran Associates, Inc., 2019.

\bibitem[Hopkins et~al.(1999)Hopkins, Reeber, Forman, and Suermondt]{spambase_dataset}
M.~Hopkins, E.~Reeber, G.~Forman, and J.~Suermondt.
\newblock {Spambase}.
\newblock UCI Machine Learning Repository, 1999.
\newblock {DOI}: https://doi.org/10.24432/C53G6X.

\bibitem[Kapishnikov et~al.(2019)Kapishnikov, Bolukbasi, Viegas, and Terry]{Kapishnikov_2019_ICCV}
A.~Kapishnikov, T.~Bolukbasi, F.~Viegas, and M.~Terry.
\newblock Xrai: Better attributions through regions.
\newblock In \emph{Proceedings of the IEEE/CVF International Conference on Computer Vision (ICCV)}, October 2019.

\bibitem[King(1992)]{dna_dataset}
R.~King.
\newblock Molecular biology (splice-junction gene sequences).
\newblock UCI Machine Learning Repository, 1992.
\newblock {DOI}: https://doi.org/10.24432/C5M888.

\bibitem[Kokhlikyan et~al.(2020)Kokhlikyan, Miglani, Martin, Wang, Alsallakh, Reynolds, Melnikov, Kliushkina, Araya, Yan, and Reblitz-Richardson]{kokhlikyan2020captum}
N.~Kokhlikyan, V.~Miglani, M.~Martin, E.~Wang, B.~Alsallakh, J.~Reynolds, A.~Melnikov, N.~Kliushkina, C.~Araya, S.~Yan, and O.~Reblitz-Richardson.
\newblock Captum: A unified and generic model interpretability library for pytorch, 2020.

\bibitem[Krippendorff(2011)]{krippendorff2011computing}
K.~Krippendorff.
\newblock Computing krippendorff’s alpha-reliability, 2011.

\bibitem[Krippendorff(2019)]{krippendorff2018content}
K.~Krippendorff.
\newblock \emph{Content Analysis: An Introduction to Its Methodology}.
\newblock SAGE Publications, Inc., 2019.
\newblock ISBN 9781071878781.
\newblock \doi{10.4135/9781071878781}.
\newblock URL \url{http://dx.doi.org/10.4135/9781071878781}.

\bibitem[Krizhevsky(2009)]{krizhevsky2009}
A.~Krizhevsky.
\newblock Learning multiple layers of features from tiny images.
\newblock Technical report, 2009.

\bibitem[Krizhevsky et~al.(2012)Krizhevsky, Sutskever, and Hinton]{krizhevsky2012}
A.~Krizhevsky, I.~Sutskever, and G.~E. Hinton.
\newblock Imagenet classification with deep convolutional neural networks.
\newblock In F.~Pereira, C.~Burges, L.~Bottou, and K.~Weinberger, editors, \emph{Advances in Neural Information Processing Systems}, volume~25. Curran Associates, Inc., 2012.

\bibitem[Lecun et~al.(1998)Lecun, Bottou, Bengio, and Haffner]{lecun1998}
Y.~Lecun, L.~Bottou, Y.~Bengio, and P.~Haffner.
\newblock Gradient-based learning applied to document recognition.
\newblock \emph{Proceedings of the IEEE}, 86\penalty0 (11):\penalty0 2278–2324, 1998.
\newblock ISSN 0018-9219.
\newblock \doi{10.1109/5.726791}.
\newblock URL \url{http://dx.doi.org/10.1109/5.726791}.

\bibitem[Lillicrap et~al.(2019)Lillicrap, Hunt, Pritzel, Heess, Erez, Tassa, Silver, and Wierstra]{lillicrap2019}
T.~P. Lillicrap, J.~J. Hunt, A.~Pritzel, N.~Heess, T.~Erez, Y.~Tassa, D.~Silver, and D.~Wierstra.
\newblock Continuous control with deep reinforcement learning, 2019.

\bibitem[Lin et~al.(2014)Lin, Maire, Belongie, Hays, Perona, Ramanan, Doll{\'a}r, and Zitnick]{lin2015}
T.-Y. Lin, M.~Maire, S.~Belongie, J.~Hays, P.~Perona, D.~Ramanan, P.~Doll{\'a}r, and C.~L. Zitnick.
\newblock Microsoft coco: Common objects in context.
\newblock In D.~Fleet, T.~Pajdla, B.~Schiele, and T.~Tuytelaars, editors, \emph{Computer Vision -- ECCV 2014}, pages 740--755. Springer International Publishing, 2014.
\newblock ISBN 978-3-319-10602-1.

\bibitem[Lin et~al.(2019)Lin, Shafiee, Bochkarev, Jules, Wang, and Wong]{QiuLin}
Z.~Q. Lin, M.~J. Shafiee, S.~Bochkarev, M.~S. Jules, X.~Y. Wang, and A.~Wong.
\newblock Do explanations reflect decisions? a machine-centric strategy to quantify the performance of explainability algorithms, 2019.

\bibitem[Liu et~al.(2018)Liu, Wang, and Matwin]{liu2018}
X.~Liu, X.~Wang, and S.~Matwin.
\newblock Improving the interpretability of deep neural networks with knowledge distillation.
\newblock In \emph{2018 IEEE International Conference on Data Mining Workshops (ICDMW)}, pages 905--912, 2018.
\newblock \doi{10.1109/ICDMW.2018.00132}.

\bibitem[Lundberg and Lee(2017)]{Lundberg2017}
S.~M. Lundberg and S.-I. Lee.
\newblock A unified approach to interpreting model predictions.
\newblock In I.~Guyon, U.~V. Luxburg, S.~Bengio, H.~Wallach, R.~Fergus, S.~Vishwanathan, and R.~Garnett, editors, \emph{Advances in Neural Information Processing Systems}, volume~30. Curran Associates, Inc., 2017.

\bibitem[Mnih et~al.(2015)Mnih, Kavukcuoglu, Silver, Rusu, Veness, Bellemare, Graves, Riedmiller, Fidjeland, Ostrovski, Petersen, Beattie, Sadik, Antonoglou, King, Kumaran, Wierstra, Legg, and Hassabis]{mnih2015}
V.~Mnih, K.~Kavukcuoglu, D.~Silver, A.~A. Rusu, J.~Veness, M.~G. Bellemare, A.~Graves, M.~Riedmiller, A.~K. Fidjeland, G.~Ostrovski, S.~Petersen, C.~Beattie, A.~Sadik, I.~Antonoglou, H.~King, D.~Kumaran, D.~Wierstra, S.~Legg, and D.~Hassabis.
\newblock Human-level control through deep reinforcement learning.
\newblock \emph{Nature}, 518\penalty0 (7540):\penalty0 529--533, 2015.
\newblock ISSN 0028-0836.
\newblock \doi{10.1038/nature14236}.

\bibitem[Netzer et~al.(2011)Netzer, Wang, Coates, Bissacco, Wu, and Ng]{netzer2011}
Y.~Netzer, T.~Wang, A.~Coates, A.~Bissacco, B.~Wu, and A.~Y. Ng.
\newblock Reading digits in natural images with unsupervised feature learning.
\newblock In \emph{NIPS workshop on deep learning and unsupervised feature learning}, volume 2011, page~4, 2011.

\bibitem[Pearl(2009)]{pearl2009causality}
J.~Pearl.
\newblock \emph{Causality}.
\newblock Cambridge University Press, 2 edition, 2009.

\bibitem[Petsiuk et~al.(2018)Petsiuk, Das, and Saenko]{Petsiuk}
V.~Petsiuk, A.~Das, and K.~Saenko.
\newblock Rise: Randomized input sampling for explanation of black-box models.
\newblock In \emph{British Machine Vision Conference (BMVC)}, 2018.
\newblock URL \url{http://bmvc2018.org/contents/papers/1064.pdf}.

\bibitem[Ribeiro et~al.(2016)Ribeiro, Singh, and Guestrin]{Ribeiro2016}
M.~T. Ribeiro, S.~Singh, and C.~Guestrin.
\newblock "why should i trust you?": Explaining the predictions of any classifier.
\newblock In \emph{Proceedings of the 22nd ACM SIGKDD International Conference on Knowledge Discovery and Data Mining}, KDD '16, page 1135–1144, New York, NY, USA, 2016. Association for Computing Machinery.
\newblock ISBN 9781450342322.
\newblock \doi{10.1145/2939672.2939778}.

\bibitem[Ribeiro et~al.(2018)Ribeiro, Singh, and Guestrin]{ribeiro2018anchors}
M.~T. Ribeiro, S.~Singh, and C.~Guestrin.
\newblock Anchors: High-precision model-agnostic explanations.
\newblock \emph{Proceedings of the AAAI Conference on Artificial Intelligence}, 32\penalty0 (1), Apr. 2018.
\newblock \doi{10.1609/aaai.v32i1.11491}.

\bibitem[Rieger and Hansen(2020)]{Rieger}
L.~Rieger and L.~Hansen.
\newblock Irof: a low resource evaluation metric for explanation methods.
\newblock In \emph{Proceedings of the Workshop AI for Affordable Healthcare at ICLR 2020}, 2020.

\bibitem[Samek et~al.(2017)Samek, Binder, Montavon, Lapuschkin, and Müller]{Samek2015}
W.~Samek, A.~Binder, G.~Montavon, S.~Lapuschkin, and K.-R. Müller.
\newblock Evaluating the visualization of what a deep neural network has learned.
\newblock \emph{IEEE Transactions on Neural Networks and Learning Systems}, 28\penalty0 (11):\penalty0 2660--2673, 2017.
\newblock \doi{10.1109/TNNLS.2016.2599820}.

\bibitem[Schmidt and Biessmann(2019)]{schmidt2019quantifying}
P.~Schmidt and F.~Biessmann.
\newblock Quantifying interpretability and trust in machine learning systems.
\newblock In \emph{AAAI 2019 Workshop on Network Interpretability for Deep Learning}, 2019.

\bibitem[Selvaraju et~al.(2017)Selvaraju, Cogswell, Das, Vedantam, Parikh, and Batra]{Selvaraju2017}
R.~R. Selvaraju, M.~Cogswell, A.~Das, R.~Vedantam, D.~Parikh, and D.~Batra.
\newblock Grad-cam: Visual explanations from deep networks via gradient-based localization.
\newblock In \emph{Proceedings of the IEEE International Conference on Computer Vision (ICCV)}, 2017.

\bibitem[Shrikumar et~al.(2017)Shrikumar, Greenside, and Kundaje]{Shrikumar2017}
A.~Shrikumar, P.~Greenside, and A.~Kundaje.
\newblock Learning important features through propagating activation differences.
\newblock In D.~Precup and Y.~W. Teh, editors, \emph{Proceedings of the 34th International Conference on Machine Learning}, volume~70 of \emph{Proceedings of Machine Learning Research}, pages 3145--3153. PMLR, 06--11 Aug 2017.

\bibitem[Silver et~al.(2017)Silver, Schrittwieser, Simonyan, Antonoglou, Huang, Guez, Hubert, Baker, Lai, Bolton, Chen, Lillicrap, Hui, Sifre, van~den Driessche, Graepel, and Hassabis]{silver2017}
D.~Silver, J.~Schrittwieser, K.~Simonyan, I.~Antonoglou, A.~Huang, A.~Guez, T.~Hubert, L.~Baker, M.~Lai, A.~Bolton, Y.~Chen, T.~Lillicrap, F.~Hui, L.~Sifre, G.~van~den Driessche, T.~Graepel, and D.~Hassabis.
\newblock Mastering the game of go without human knowledge.
\newblock \emph{Nature}, 550\penalty0 (7676):\penalty0 354–359, 2017.
\newblock ISSN 1476-4687.
\newblock \doi{10.1038/nature24270}.

\bibitem[Simonyan and Zisserman(2015)]{simonyan2015}
K.~Simonyan and A.~Zisserman.
\newblock Very deep convolutional networks for large-scale image recognition.
\newblock In Y.~Bengio and Y.~LeCun, editors, \emph{3rd International Conference on Learning Representations, {ICLR} 2015, San Diego, CA, USA, May 7-9, 2015, Conference Track Proceedings}, 2015.

\bibitem[Simonyan et~al.(2014)Simonyan, Vedaldi, and Zisserman]{Simonyan2014}
K.~Simonyan, A.~Vedaldi, and A.~Zisserman.
\newblock Deep inside convolutional networks: Visualising image classification models and saliency maps.
\newblock In Y.~Bengio and Y.~LeCun, editors, \emph{2nd International Conference on Learning Representations, {ICLR} 2014, Banff, AB, Canada, April 14-16, 2014, Workshop Track Proceedings}, 2014.

\bibitem[Smilkov et~al.(2017)Smilkov, Thorat, Kim, Viégas, and Wattenberg]{Smilkov2017}
D.~Smilkov, N.~Thorat, B.~Kim, F.~Viégas, and M.~Wattenberg.
\newblock Smoothgrad: removing noise by adding noise, 2017.

\bibitem[Springenberg et~al.(2014)Springenberg, Dosovitskiy, Brox, and Riedmiller]{Springenberg2014}
J.~T. Springenberg, A.~Dosovitskiy, T.~Brox, and M.~Riedmiller.
\newblock Striving for {{Simplicity}}: {{The All Convolutional Net}}.
\newblock \emph{International Conference on Learning Representations - Workshop Track Proceedings}, 3, Dec. 2014.

\bibitem[Srinivasan(1993)]{statlog_landsat}
A.~Srinivasan.
\newblock {Statlog (Landsat Satellite)}.
\newblock UCI Machine Learning Repository, 1993.

\bibitem[Sturmfels et~al.(2020)Sturmfels, Lundberg, and Lee]{Sturmfels2020}
P.~Sturmfels, S.~Lundberg, and S.-I. Lee.
\newblock Visualizing the impact of feature attribution baselines.
\newblock \emph{Distill}, 2020.
\newblock \doi{10.23915/distill.00022}.
\newblock https://distill.pub/2020/attribution-baselines.

\bibitem[Sundararajan and Najmi(2020)]{Sundararajan2019}
M.~Sundararajan and A.~Najmi.
\newblock The many shapley values for model explanation.
\newblock In H.~D. III and A.~Singh, editors, \emph{Proceedings of the 37th International Conference on Machine Learning}, volume 119 of \emph{Proceedings of Machine Learning Research}, pages 9269--9278. PMLR, 13--18 Jul 2020.

\bibitem[Sundararajan et~al.(2017)Sundararajan, Taly, and Yan]{Sundararajan2017}
M.~Sundararajan, A.~Taly, and Q.~Yan.
\newblock Axiomatic attribution for deep networks.
\newblock In D.~Precup and Y.~W. Teh, editors, \emph{Proceedings of the 34th International Conference on Machine Learning}, volume~70 of \emph{Proceedings of Machine Learning Research}, pages 3319--3328. PMLR, 06--11 Aug 2017.

\bibitem[Tomsett et~al.(2020)Tomsett, Harborne, Chakraborty, Gurram, and Preece]{Tomsett}
R.~Tomsett, D.~Harborne, S.~Chakraborty, P.~Gurram, and A.~Preece.
\newblock Sanity checks for saliency metrics.
\newblock \emph{Proceedings of the AAAI Conference on Artificial Intelligence}, 34\penalty0 (04):\penalty0 6021–6029, 2020.
\newblock ISSN 2159-5399.
\newblock \doi{10.1609/aaai.v34i04.6064}.

\bibitem[van~der Velden et~al.(2022)van~der Velden, Kuijf, Gilhuijs, and Viergever]{vandervelden2022}
B.~H. van~der Velden, H.~J. Kuijf, K.~G. Gilhuijs, and M.~A. Viergever.
\newblock Explainable artificial intelligence (xai) in deep learning-based medical image analysis.
\newblock \emph{Medical Image Analysis}, 79:\penalty0 102470, July 2022.
\newblock ISSN 1361-8415.
\newblock \doi{10.1016/j.media.2022.102470}.

\bibitem[Vanschoren et~al.(2013)Vanschoren, van Rijn, Bischl, and Torgo]{OpenML2013}
J.~Vanschoren, J.~N. van Rijn, B.~Bischl, and L.~Torgo.
\newblock Openml: networked science in machine learning.
\newblock \emph{SIGKDD Explorations}, 15\penalty0 (2):\penalty0 49--60, 2013.
\newblock \doi{10.1145/2641190.2641198}.

\bibitem[Vaswani et~al.(2017)Vaswani, Shazeer, Parmar, Uszkoreit, Jones, Gomez, Kaiser, and Polosukhin]{vaswani2017}
A.~Vaswani, N.~Shazeer, N.~Parmar, J.~Uszkoreit, L.~Jones, A.~N. Gomez, L.~u. Kaiser, and I.~Polosukhin.
\newblock Attention is all you need.
\newblock In I.~Guyon, U.~V. Luxburg, S.~Bengio, H.~Wallach, R.~Fergus, S.~Vishwanathan, and R.~Garnett, editors, \emph{Advances in Neural Information Processing Systems}, volume~30. Curran Associates, Inc., 2017.

\bibitem[Wachter et~al.(2017)Wachter, Mittelstadt, and Russell]{wachter2017counterfactual}
S.~Wachter, B.~Mittelstadt, and C.~Russell.
\newblock Counterfactual explanations without opening the black box: Automated decisions and the gdpr.
\newblock \emph{Harvard Journal of Law \& Technology (Harvard JOLT)}, 31:\penalty0 841, 2017.

\bibitem[Wang et~al.(2020)Wang, Wang, Du, Yang, Zhang, Ding, Mardziel, and Hu]{Wang2020ScoreCAM}
H.~Wang, Z.~Wang, M.~Du, F.~Yang, Z.~Zhang, S.~Ding, P.~Mardziel, and X.~Hu.
\newblock Score-cam: Score-weighted visual explanations for convolutional neural networks.
\newblock In \emph{Proceedings of the IEEE/CVF Conference on Computer Vision and Pattern Recognition (CVPR) Workshops}, June 2020.

\bibitem[{Wang} et~al.(2004){Wang}, {Bovik}, {Sheikh}, and {Simoncelli}]{wang2004ssim}
Z.~{Wang}, A.~C. {Bovik}, H.~R. {Sheikh}, and E.~P. {Simoncelli}.
\newblock {Image Quality Assessment: From Error Visibility to Structural Similarity}.
\newblock \emph{IEEE Transactions on Image Processing}, 13\penalty0 (4):\penalty0 600--612, Apr. 2004.
\newblock \doi{10.1109/TIP.2003.819861}.

\bibitem[Xiao et~al.(2017)Xiao, Rasul, and Vollgraf]{xiao2017}
H.~Xiao, K.~Rasul, and R.~Vollgraf.
\newblock Fashion-mnist: a novel image dataset for benchmarking machine learning algorithms, 2017.

\bibitem[Yang and Kim(2019)]{Yang2019}
M.~Yang and B.~Kim.
\newblock Benchmarking attribution methods with relative feature importance, 2019.

\bibitem[Yeh et~al.(2019)Yeh, Hsieh, Suggala, Inouye, and Ravikumar]{Yeh2019}
C.-K. Yeh, C.-Y. Hsieh, A.~Suggala, D.~I. Inouye, and P.~K. Ravikumar.
\newblock On the (in)fidelity and sensitivity of explanations.
\newblock In H.~Wallach, H.~Larochelle, A.~Beygelzimer, F.~d\textquotesingle Alch\'{e}-Buc, E.~Fox, and R.~Garnett, editors, \emph{Advances in Neural Information Processing Systems}, volume~32. Curran Associates, Inc., 2019.

\bibitem[Yona and Greenfeld(2021)]{yona2021}
G.~Yona and D.~Greenfeld.
\newblock Revisiting {{Sanity Checks}} for {{Saliency Maps}}.
\newblock In \emph{{{eXplainable AI}} Approaches for Debugging and Diagnosis}, Oct. 2021.

\bibitem[Zeiler and Fergus(2014)]{Zeiler2014}
M.~D. Zeiler and R.~Fergus.
\newblock Visualizing and understanding convolutional networks.
\newblock In D.~Fleet, T.~Pajdla, B.~Schiele, and T.~Tuytelaars, editors, \emph{Computer Vision -- ECCV 2014}, pages 818--833, Cham, 2014. Springer International Publishing.
\newblock ISBN 978-3-319-10590-1.

\bibitem[Zhou et~al.(2017)Zhou, Lapedriza, Khosla, Oliva, and Torralba]{zhou2017places}
B.~Zhou, A.~Lapedriza, A.~Khosla, A.~Oliva, and A.~Torralba.
\newblock Places: {{A}} 10 million image database for scene recognition.
\newblock \emph{IEEE Transactions on Pattern Analysis and Machine Intelligence}, 2017.

\end{thebibliography}

\end{document}